\newif\ifpaper
\papertrue

\documentclass[acmtog]{acmart}

\usepackage[ruled, vlined]{algorithm2e}
\usepackage{amsthm}
\usepackage{makecell}
\usepackage{multirow}
\usepackage{tabularx}
\usepackage{rotating}
\usepackage{siunitx}
\usepackage{xcolor}

\usepackage{hyperref}

\usepackage{colortbl}

\usepackage{enumitem,kantlipsum}

\newcolumntype{P}[1]{>{\centering\arraybackslash}p{#1}}
\newcolumntype{Y}{>{\centering\arraybackslash}X}

\DeclareMathOperator*{\argmin}{argmin}

\definecolor{color_1}{RGB}{255,0,128}
\definecolor{color_2}{RGB}{128,128,0}
\definecolor{color_3}{RGB}{0,128,0}
\definecolor{color_4}{RGB}{0,0,0}
\definecolor{Gray}{gray}{0.90}  

\AtBeginDocument{%
  }

\copyrightyear{2024}
\acmYear{2024}
\setcopyright{rightsretained}
\acmConference[SA Conference Papers '24]{SIGGRAPH Asia 2024 Conference Papers}{December 3--6, 2024}{Tokyo, Japan}
\acmBooktitle{SIGGRAPH Asia 2024 Conference Papers (SA Conference Papers '24), December 3--6, 2024, Tokyo, Japan}\acmDOI{10.1145/3680528.3687581} \acmISBN{979-8-4007-1131-2/24/12}

\begin{document}

\title{Occupancy-Based Dual Contouring}


\author{Jisung Hwang}
\affiliation{%
  \institution{KAIST}
  \city{Daejeon}
  \country{South Korea}}
\email{4011hjs@kaist.ac.kr}

\author{Minhyuk Sung}
\affiliation{%
  \institution{KAIST}
  \city{Daejeon}
  \country{South Korea}}
\email{mhsung@kaist.ac.kr}


\begin{abstract}
We introduce a dual contouring method that provides state-of-the-art performance for occupancy functions while achieving computation times of a few seconds. Our method is learning-free and carefully designed to maximize the use of GPU parallelization. The recent surge of implicit neural representations has led to significant attention to occupancy fields, resulting in a wide range of 3D reconstruction and generation methods based on them. However, the outputs of such methods have been underestimated due to the bottleneck in converting the resulting occupancy function to a mesh. Marching Cubes tends to produce staircase-like artifacts, and most subsequent works focusing on exploiting signed distance functions as input also yield suboptimal results for occupancy functions. Based on Manifold Dual Contouring (MDC), we propose Occupancy-Based Dual Contouring (ODC), which mainly modifies the computation of grid edge points (1D points) and grid cell points (3D points) to not use any distance information. We introduce auxiliary 2D points that are used to compute local surface normals along with the 1D points, helping identify 3D points via the quadric error function. To search the 1D, 2D, and 3D points, we develop fast algorithms that are parallelizable across all grid edges, faces, and cells. Our experiments with several 3D neural generative models and a 3D mesh dataset demonstrate that our method achieves the best fidelity compared to prior works.
\ifpaper The code is available at \href{https://github.com/KAIST-Visual-AI-Group/ODC}{https://github.com/KAIST-Visual-AI-Group/ODC}.
\else
\fi

\end{abstract}



\begin{teaserfigure}
  \includegraphics[width=\textwidth]{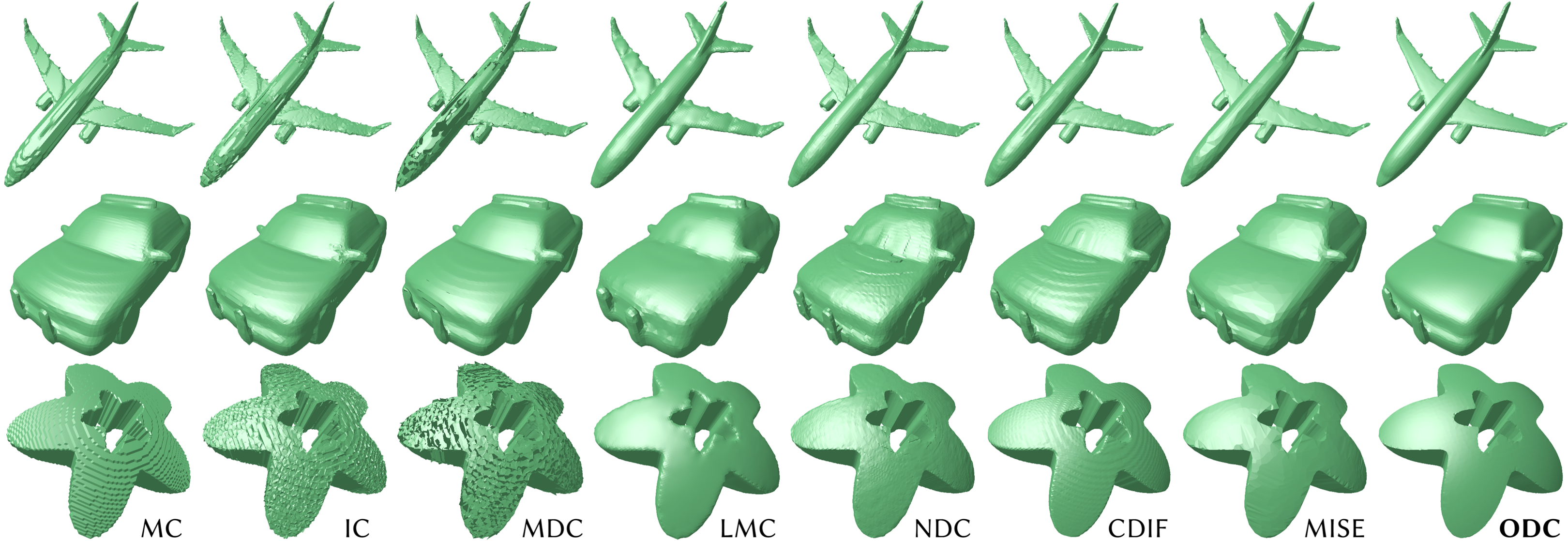}
  \caption{Comparisons among our \textbf{Occupancy-based Dual Contouring (ODC)} and various prior mesh extraction methods, including Marching Cubes (MC)~\cite{lewiner2003efficient}, Intersection-free Contouring (IC)~\cite{ju2006intersection}, Manifold Dual Contouring (MDC)~\cite{manifolddualcontouring}, Lempitsky's smoothing method (LMC)~\cite{lempitsky2010surface}, Neural Dual Contouring (NDC)~\cite{neuraldualcontouring}, Contouring DIF (CDIF)~\cite{manson2011contouring}, and Multiresolution IsoSurface Extraction (MISE)~\cite{occupancenetworks}. The meshes are extracted from the output of SALAD~\cite{koo2023salad}, 3DShape2VecSet~\cite{zhang20233dshape2vecset}, and Fast winding number~\cite{barill2018fast} on Myles et al.'s dataset~\cite{myles2014robust} for each row, respectively. Notably, as shown in the first row, we have perceived the output of SALAD as an aliased shape (leftmost), as provided by the authors. However, its genuine shape, as revealed by our method, ODC, looks like the rightmost one.}
  \label{fig:teaser}
\end{teaserfigure}

\maketitle

\section{Introduction}
\label{sec:introduction}



Meshes are undoubtedly the most fundamental and common form of 3D content representation in computer graphics. Amid the recent surge of neural techniques for 3D reconstruction and generation using implicit representations, converting an implicit function to a 3D mesh has become increasingly important, enabling the use of outputs in a wide range of applications: editing and manipulation, texturing, applying material properties, physical simulation, 3D printing, and more.

In the realm of neural implicit representations, \emph{occupancy} functions have particularly risen in prominence, as evidenced by the vast array of previous works~\cite{imnet, occupancenetworks, jia2020learning, peng2020convolutional, poursaeed2020coupling, genova2020local, ibing20213d, lionar2021dynamic, tang2021sa, lamb2022deepmend, zhang20223dilg, boulch2022poco, hertz2022spaghetti, lin2022neuform, feng2022fof, shue20233d, zhang20233dshape2vecset, liu2023exim, erkocc2023hyperdiffusion, koo2023salad, tian2024occ3d, zhao2024michelangelo, wang2024sparse}. The attention to occupancy functions has also simultaneously highlighted the challenges associated with converting these functions into a mesh. For example, when applying Marching Cubes~\cite{marchingcubes}, the most fundamental yet standard technique for implicit-to-mesh conversion, to an occupancy function that outputs binary inside/outside labels, it inevitably produces staircase-like artifacts, as shown in the first column of Figure~\ref{fig:teaser}, since the mesh vertex on the surface-crossing grid edge is always chosen as the midpoint.
This is also generally observed even when the occupancy is represented as a continuous value as the quantity still does not represent distance.

The lack of fidelity in the outputs from Marching Cubes is a recognized issue not only for occupancy functions but also for signed distance functions. This has prompted extensive research over decades, although most studies have focused on the case when the given function is a \emph{signed distance function}. Notably, Fuhrmann et al.~\shortcite{fuhrmann2015accurate} proposed to replace the linear interpolation when finding the surface-intersecting point on a grid edge, referred to as \emph{1D points}, with a cubic interpolation using the distance information. Extended Marching Cubes~\cite{kobbelt2001feature}, Dual Contouring~\cite{dualcontouring}, and Manifold Dual Contouring~\cite{manifolddualcontouring} are also milestone works proposing to leverage the \emph{gradient} of the signed distance function as a surface normal and identifying sharp corner points within a grid cell, termed \emph{3D point}, through the Quadric Error Function (QEF) with an assumption of local flatness. Manson and Schaefer~\shortcite{manson2010isosurfaces} and Bhattacharya and Wenger~\shortcite{bhattacharya2013constructing} have further extended this approach by incorporating distance values into the QEF to improve vertex placement. While these advancements have been successful in replacing the original Marching Cubes and improving fidelity, they produce suboptimal results with occupancy functions even when they output continuous values in the $[0, 1]$ range since the gradient\textcolor{color_4}{, which is orthogonal to an infinitesimal region, may not properly represent the local surface area.}

To generate high-fidelity 3D meshes from occupancy functions, we propose a novel method called \textbf{Occupancy-Based Dual Contouring (ODC)}.
We particularly focus on (but are not limited to) cases where the occupancy function is provided as a \emph{neural network} and aim to reveal the \emph{true} shape encoded in the network-based function. Even when a neural occupancy function encodes the fine details of the geometry in continuous 3D space, previous mesh conversion methods have failed to precisely render these details into a mesh, resulting in the user perceiving the artifact-laden mesh as the final output (as shown in 1st-7th columns in Figure~\ref{fig:teaser}). Our novel technique reveals the true outputs of such methods, as shown in the last column of Figure~\ref{fig:teaser}, demonstrating its significant impact on research in 3D neural implicit representations.


Building upon the Manifold Dual Contouring~\cite{manifolddualcontouring}, which captures sharp features using the QEF and surface normal information (without directly using distance), our main goal is to replace the following components with iterative searches: 1) linear interpolation in the \emph{1D point} search and 2) the use of the gradient of the signed distance as a surface normal in \emph{3D point} identification. Specifically, to compute the surface normal without using the gradient of the signed distance, we first define \emph{2D points} on grid faces that are assumed to be on the same local flat surface as neighboring 1D points,
\textcolor{color_4}{and then compute the surface normal at a 1D point using two adjacent 2D points which are assumed to be on the same plane.}
Both the 1D and 2D points are identified with binary or a combination of linear and binary searches, thus fully exploiting the occupancy information in continuous space, not only at the grid points. 

Our technical contributions can be summarized as follows:
\begin{enumerate}[leftmargin=*,noitemsep,topsep=0em]
\item 
We propose a novel 2D point search algorithm (Section~\ref{sec:method_2d_search}) that guarantees finding points on the same local plane as the neighboring 1D points under the local flatness assumption.
\item 
We designed the 1D, 2D, and 3D point calculations (Section~\ref{sec:method}) to fully exploit GPU parallelism, enabling parallel processing for all grid edges, faces, and cells simultaneously. This allows computation times to remain within five seconds for resolutions as high as $128^3$.
\item 
We additionally modify the final polygonization step (Section~\ref{sec:method_poly}) to significantly reduce self-intersecting mesh faces in the outputs by incorporating the quad splitting technique of Intersection-free Contouring~\cite{ju2006intersection}.
\end{enumerate}

\textcolor{color_4}{While most prior methods can only use samples of an implicit function available on a grid~\cite{marchingcubes, dualmarchingcubes, lewiner2003efficient, lempitsky2010surface, manson2011contouring, neuralmarchingcubes, neuraldualcontouring} or grid samples augmented with additional data~\cite{kobbelt2001feature, dualcontouring, ju2006intersection, manifolddualcontouring}, our method can fully exploit the underlying implicit function defined in \emph{continuous} 3D space, resulting in high-fidelity meshes.}
\textcolor{color_4}{Although}
Multiresolution Isosurface Extraction (MISE)~\cite{occupancenetworks} is a notable exception, its gradient-descent-based vertex pulling approach often fails to faithfully capture the entire surface, frequently breaking manifoldness and resulting in numerous mesh face intersections.
Our ODC produces significantly higher-fidelity outputs while guaranteeing manifoldness and greatly reducing intersecting faces.

In our experiments, we present results with neural implicit representations generated from SALAD~\cite{koo2023salad}, 3DShape2VecSet \cite{zhang20233dshape2vecset}, Michelangelo~\cite{zhao2024michelangelo}, and IM-NET~\cite{imnet}. To assess the quality of the output meshes, we additionally use 3D meshes from Myles' dataset, determining the occupancy using winding number computation~\cite{barill2018fast}.
We compare our method with occupancy function polygonization methods \cite{lempitsky2010surface, manson2011contouring, occupancenetworks}, recent learning-based methods \cite{neuraldualcontouring, maruani2024ponq}, as well as the base methods of ours \cite{lewiner2003efficient, ju2006intersection, manifolddualcontouring}, and demonstrate the superior performance of our method.

\section{Related Work}
\label{sec:related_work}

Mesh construction from implicit functions has been researched for over three decades. De Arjuo et al.~\shortcite{survey} have summarized previous work in their survey, categorizing the methodologies into three classes: spatial decomposition, surface tracking~\cite{hilton1997marching,marchingtriangles,akkouche2001adaptive,karkanis2001curvature,mccormick2002edge}, and inflation and shrinkwrap~\cite{kobbelt1999shrink,van2004shrinkwrap,stander1995interactive,bottino1996shrinkwrap,hormann2002hierarchical,sellan2023reach}. Among these, the spatial decomposition approach, represented by Marching Cubes~\cite{marchingcubes}, has been predominant due to its simplicity, robustness, parameter-free nature, and compatibility with parallel GPU computation. Since its introduction, numerous improvements have addressed the limitations of Marching Cubes, such as resolving topology ambiguities~\cite{nielson1991asymptotic,natarajan1994generating,chernyaev1995marching,lopes2003improving,marchingcubes2003,lewiner2003efficient,custodio2013practical}, enabling adaptive resolution~\cite{bloomenthal1988polygonization, muller1993adaptive,shekhar1996octree,westermann1999real,hormann2002hierarchical}, and supporting the use of tetrahedral instead of cubic grids~\cite{marchingtetrahedra,zhou1997multiresolution,muller1997visualization,nielson2008dual}.

A key challenge of Marching Cubes has been its limited fidelity, which has spurred extensive research. Fuhrmann et al.~\shortcite{fuhrmann2015accurate} suggested replacing linear interpolation for surface point detection with cubic interpolation. Extended Marching Cubes~\cite{kobbelt2001feature} and Dual Contouring~\cite{dualcontouring} were pivotal works that first proposed placing mesh vertices on sharp corners using a quadric error function, leveraging the gradient of the signed distance function as the surface normal.
Cubical Marching Squares~\cite{ho2005cubical} resolved topology ambiguity problem with sharp features. 
Manifold Dual Contouring~\cite{manifolddualcontouring} further refined Dual Contouring, ensuring manifoldness while placing multiple mesh vertices in each cell based on the idea of Dual Marching Cubes~\cite{dualmarchingcubes}. Intersection-free Contouring~\cite{ju2006intersection} is another direction that ensures the output mesh is free of intersections by modifying the final polygonization step. Manson and Schaefer~\shortcite{manson2010isosurfaces} and Bhattacharya and Wenger~\shortcite{bhattacharya2013constructing} went even further by adapting the quadric error function to include distance information. The aforementioned methods assume that the implicit function is provided as a signed distance function, making their application to occupancy functions either infeasible or non-trivial. Building on Manifold Dual Contouring, which uses the gradient as the surface normal without directly employing distance information, we propose a novel method that produces high-fidelity meshes with sharp features, eliminating the need for gradients.

Recent work has also explored mesh construction with sharp features from point clouds~\cite{salman2010feature,dey2012feature,wang2013feature} or unsigned distance functions as inputs~\cite{guillard2022meshudf,zhou2022learning,hou2023robust,zhang2023surface}. However, our objective differs as we focus on using an occupancy function, a prevalent format in neural implicit representations~\cite{imnet, occupancenetworks, jia2020learning, peng2020convolutional, poursaeed2020coupling, genova2020local, ibing20213d, lionar2021dynamic, tang2021sa, lamb2022deepmend, zhang20223dilg, boulch2022poco, hertz2022spaghetti, lin2022neuform, feng2022fof, shue20233d, zhang20233dshape2vecset, liu2023exim, erkocc2023hyperdiffusion, koo2023salad, tian2024occ3d, zhao2024michelangelo, wang2024sparse}.

Among the vast array of literature, in the following, we specifically focus on surveying prior works that take an occupancy function as an input for mesh construction. Additionally, we review recent advancements in neural-network-based approaches.

\begin{figure}[!t]
  \centering
  \includegraphics[width=\linewidth]{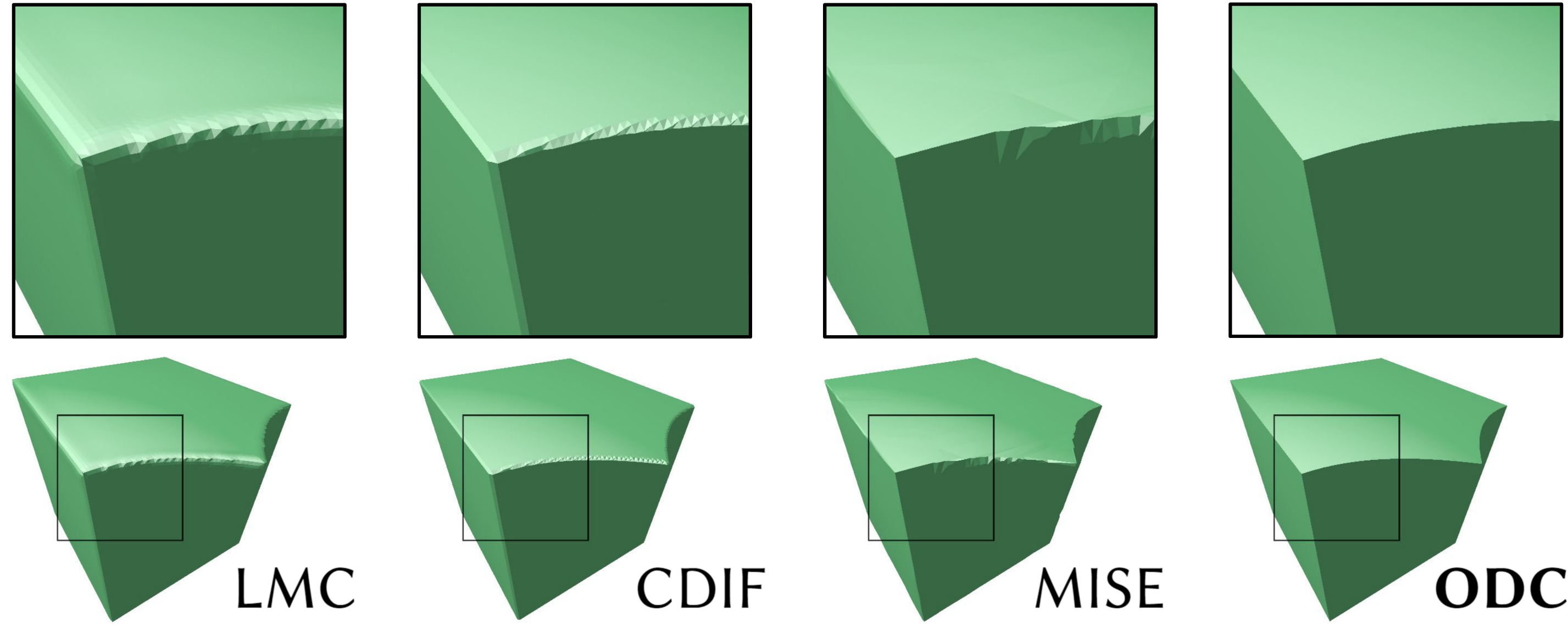}
  \vspace{-0.7cm}
  \caption{
  Sharp feature preservation of occupancy-based models (from left to right): Smoothing method (LMC)~\cite{lempitsky2010surface}, Contouring DIF (CDIF)~\cite{manson2011contouring}, MISE~\cite{occupancenetworks}, and ODC (ours). ODC effectively captures sharp features.
  }
  \label{fig:Occ-based method}
  \vspace{-0.3cm}
\end{figure}

\subsection{Occupancy Function Polygonization}
While it is feasible to apply Marching Cubes to an occupancy function, despite the lack of linearity due to discretization, the typical limitation is the production of staircase-like outputs, as illustrated in Figure~\ref{fig:Myles_occ_result}. A straightforward solution to these visual artifacts is to smooth the occupancy values~\cite{taubin1995signal, desbrun1999implicit,ohtake2001mesh,gibson1998constrained,chica2008pressing,lempitsky2010surface}, converting them into continuous quantities. However, this approach results in deviating from the original function and thus losing fidelity. Discrete Indicator Function (DIF) and Contouring DIF~\cite{manson2011contouring,evrard2018surface,incahuanaco2023surface} represent another attempt to relax the discretized occupancy by depicting it as the fraction of the cell contained by the object. However, these quantities are not distance values and thus cannot properly provide surface-intersecting points and surface normals through linear interpolation and gradients. Furthermore, the methods are limited to using the quantities in discrete samples in the 3D space (the grid points). Multiresolution Isosurface Extraction (MISE)~\cite{occupancenetworks} is a notable technique that fully utilizes the given occupancy function throughout the continuous space while refining initial mesh vertices toward the surface via gradient descent. However, it often fails to capture all sharp features, as shown in Figure~\ref{fig:Occ-based method}. We introduce a novel method that also fully exploits the given occupancy information in the continuous space but more effectively captures sharp features through a spatial decomposition approach.

\subsection{Neural Mesh Construction}
Recent research has explored mesh construction using neural approaches. MeshSDF~\cite{meshsdf}, DMTet~\cite{dmtet}, and FlexiCubes~\cite{flexicubes} introduce \emph{differentiable} methods that optimize an implicit function to achieve the best mesh with respect to a given objective function. However, without optimization, their mesh construction methods are identical to existing techniques: Marching Cubes~\cite{marchingcubes}, Marching Tetrahedra~\cite{marchingtetrahedra}, and Dual Marching Cubes~\cite{dualmarchingcubes}. Our goal is to improve the mesh construction itself given an occupancy function.
VoroMesh \cite{voromesh} and PoNQ \cite{maruani2024ponq} also propose novel methods leveraging inside/outside regions for mesh construction, either by optimizing auxiliary points or by learning them using a neural network. However, these methods are based on signed distance functions and perform poorly when applied to an occupancy function, as illustrated in Figure~\ref{fig:RFS_PoNQ_ODC}. Neural Marching Cubes~\cite{neuralmarchingcubes} and Neural Dual Contouring~\cite{neuraldualcontouring} are notable for applying Marching Cubes~\cite{marchingcubes} and Dual Contouring~\cite{dualcontouring} with additional information predicted by a neural network, although they also demonstrate limited generalizability to unseen data. We propose a neural-network-free method for polygonizing occupancy functions that can be directly applied to any occupancy function without the need for training.

\section{Problem Definition and Notations}
\label{sec:problem_definition}

Given an implicit \emph{occupancy} function $\phi:\mathbb{R}^3 \rightarrow \{0, 1\}$ representing a shape $\mathcal{S}$ as the boundary of the two discrete labels (0 for outside and 1 for inside), we aim to extract a mesh describing the surface using a 3D grid uniformly dividing a 3D space $\mathcal{P} \subset \mathbb{R}^3$. 
Let $\mathcal{C}$ denote the set of grid cells and $\mathcal{V}$ and $\mathcal{E}$ indicate the sets of grid vertices and edges, respectively. For each cell $c \in \mathcal{C}$, $\mathcal{V}_c \subset \mathcal{V}$ and $\mathcal{E}_c \subset \mathcal{E}$ denote the sets of associated 8 grid vertices and 12 grid edges, respectively.
$\mathbf{p}_v \in \mathcal{P}$ denotes the \textcolor{color_4}{3D coordinates} of a grid vertex $v \in \mathcal{V}$.

The 3D space $\mathcal{P}$ is partitioned into two regions: inside $\mathcal{P}^{\text{in}}$ and outside $\mathcal{P}^{\text{out}}$.
$\mathcal{P}^{\text{in}}$ includes points $\mathbf{p}$ where $\phi(\mathbf{p}) = 1$, and $\mathcal{P}^{\text{out}}$ consists of points $\mathbf{p}$ where $\phi(\mathbf{p}) = 0$.
The sets of grid edges and cells passing through the surface $\mathcal{S}$ are denoted as $\mathcal{E}^\mathcal{S} = \{e = (v^\text{in}v^\text{out}) \in \mathcal{E}:\mathbf{p}_{v^\text{in}} \in \mathcal{P}^\text{in},\ \mathbf{p}_{v^\text{out}} \in \mathcal{P}^\text{out} \}$ and $\mathcal{C}^\mathcal{S} = \{c \in \mathcal{C}:\ \mathcal{E}_c \cap \mathcal{E}^\mathcal{S} \neq \emptyset\}$, respectively.



\section{Background: Manifold Dual Contouring}
\label{sec:mdc}
Manifold Dual Contouring (MDC)~\cite{manifolddualcontouring} is a seminal advancement over Marching Cubes~\cite{marchingcubes}, specifically designed to capture sharp features while ensuring manifoldness. However, MDC is not suitable for binary occupancy functions, as it requires the implicit function to be a \emph{signed distance function} and utilizes its \emph{gradient} as surface normal. Our goal is to develop a novel method that adapts Manifold Dual Contouring to work with occupancy function inputs, while fully leveraging the advantages of a neural implicit function that enables the parallel evaluation of thousands of points in several milliseconds.

To provide background, we briefly describe the three main steps of MDC below: 1) Sampling surface points along grid edges, referred to as \emph{1D points} in this paper; 2) Identifying sharp corner points within each cell, termed \emph{3D points}; and 3) Connecting these sharp corner points to form the final mesh. Here, we assume that the input implicit function, $\psi:\mathbb{R}^3 \rightarrow \mathbb{R}$, is a signed distance function, with $\psi(\mathbf{p}) < 0$ indicating inside and $\psi(\mathbf{p}) > 0$ denoting outside. The surface $\mathcal{S}$ is thus represented as the zero-level set.

\paragraph{1D Point \textcolor{color_4}{Identification}}
\label{sec:mdc_1d_search}
A 1D point $\mathbf{p}_e$, representing a point on the surface $\mathcal{S}$ along a zero-crossing grid edge $e = (v^\text{in}v^\text{out}) \in \mathcal{E}^\mathcal{S}$, is determined exactly as in the original Marching Cubes~\cite{marchingcubes}, while assuming the \emph{linearity} of the given signed distance function $\psi$ along the edge $e$:
\begin{align}
    \mathbf{p}_e = \frac{\psi(\mathbf{p}_{v^\text{in}})\ \mathbf{p}_{v^\text{out}} - \psi(\mathbf{p}_{v^\text{out}})\ \mathbf{p}_{v^\text{in}}}{\psi(\mathbf{p}_{v^\text{in}})-\psi(\mathbf{p}_{v^\text{out}})}.
    \label{eq:1d_iso-level}
\end{align}

\paragraph{3D Point Identification}
\label{sec:mdc_3d_search}
Unlike Marching Cubes, Manifold Dual Contouring uses 1D points not as vertices of the final mesh, but as \emph{auxiliary} points to identify \emph{sharp corners} within cells, referred to as 3D points. These 3D points then serve as the vertices of the final mesh.

To detect the 3D points $\{\mathbf{p}_{c,i}\}_i$ within a zero-crossing cell $c$, the 1D points belonging to the cell are first partitioned based on their adjacency, following a grouping pattern used in Dual Marching Cubes~\cite{dualmarchingcubes} to form faces
\ifpaper (see Figure~\ref{fig:22_config} in the appendix).
\fi
A 3D point is then defined for each partition using the Quadric Error Function (QEF).
Let $\mathcal{E}^\mathcal{S}_{c,i}$ denote the zero-crossing edges corresponding to the 1D points in the $i$-th partition of the grid cell $c$. Assuming \emph{local flatness} of the surface $\mathcal{S}$, the 3D point for the $i$-th partition, representing a sharp corner point, is determined as the intersection of flat planes defined at each 1D point in the partition:
\begin{equation}
    \mathbf{p}_{c,i} = \underset{\mathbf{p}}{\argmin}\sum_{e \in \mathcal{E}_{c,i}^{S}} [\nabla \psi(\mathbf{p}_e) \cdot (\mathbf{p} - \mathbf{p}_e)]^2.
    \label{eq:mdc_qef}
\end{equation}
Note that the gradient of the signed distance function, $\nabla\psi(\mathbf{p}_e)$, is used as the surface normal.


\paragraph{Polygonization}
\label{sec:mdc_poly}
The 3D points are finally connected to each other to form the output mesh. Consider a zero-crossing grid edge $e \in \mathcal{E}^\mathcal{S}$ and its four adjacent grid cells $\{c_k\}_{k=1,...,4}$.
When identifying the 3D point $\mathbf{p}_{c_k, i_k}$ for each cell $c_k$ such that $e \in \mathcal{E}_{c_k, i_k}^\mathcal{S}$, the output mesh is formed by splitting a quadrangle, defined by the set of $\{\mathbf{p}_{c_k, i_k}\}_k$, into two triangles.

\begin{figure}[t]
    \centering
    \includegraphics[width = \linewidth]{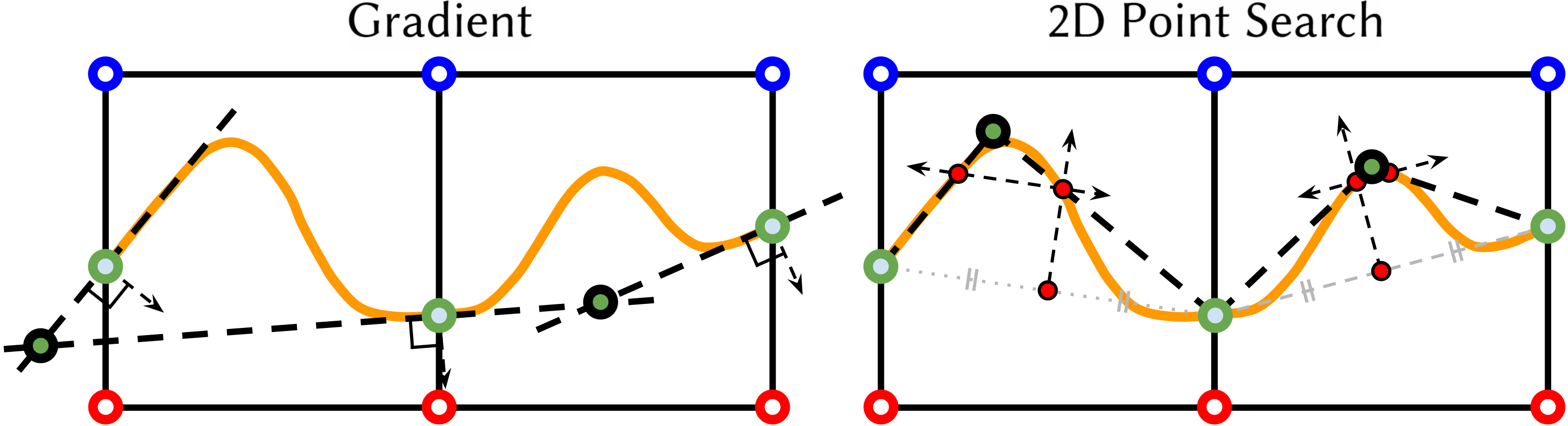}
    \vspace{-0.7cm}
    \caption{
    \textcolor{color_4}{2D illustration comparing local planes derived from gradients and 2D points. The QEF identifies a point where local flat planes (thick dashed lines) passing through the 1D points (gray dots with green boundaries) meet. On the left, gradients are used as normals for the local flat planes. Although these planes fit the surface of infinitesimal regions, they do not represent larger surface areas within voxels, causing their intersections to fall outside the actual surface. In contrast, the right side shows local planes formed using 2D points (green dots with black boundaries). Since 2D points are determined by searching within a grid face, the resulting local planes align more closely with the actual surface inside the voxel, leading to improved QEF results. Refer to Section~\ref{sec:method_2d_search} for more details on the 2D point search.}
    }
    \label{fig:benefit_2d_search_illustration}
    \vspace{-0.4cm}
\end{figure}

\paragraph{Challenges of MDC on Occupancy Fields}
\label{sec:mdc_challenge}
Manifold Dual Contouring (MDC) is not applicable to an occupancy field due to its assumptions of linearity and the use of gradients as surface normals, which are incompatible with an occupancy function that outputs discrete values. Even when the occupancy field is modeled as a \emph{continuous} function outputting values in the range of $[0, 1]$, typical for neural implicit functions, MDC still produces suboptimal results as illustrated on the first and second rows in Figure~\ref{fig:teaser}. \textcolor{color_4}{This is because the linearity of the occupancy function cannot be assumed, and the QEF output from the planes, which are represented by the gradient of the function, can be misaligned with the surface, as shown in Figure~\ref{fig:benefit_2d_search_illustration}.}

\begin{figure*}[!t]
  \centering
  \includegraphics[width=\linewidth]{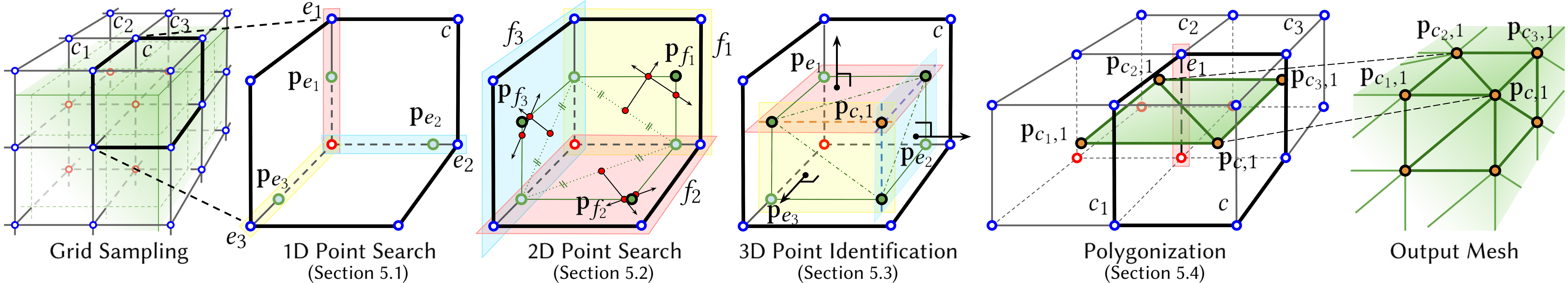}
  \vspace{-0.7cm}
  \caption{
  Overview of Occupancy-Based Dual Contouring (ODC). Dots with blue, red, and green boundaries represent grid vertices that are outside, inside, and 1D points, respectively. Dots with black boundaries, colored green and orange, represent 2D points and 3D points, respectively. Refer to Section~\ref{sec:method} for details.
  }
  \vspace{0cm}
  \label{fig:odc_overview}
\end{figure*}

\section{Occupancy-Based Dual Contouring}
\label{sec:method}

Based on the Manifold Dual Contouring framework outlined above, we introduce our novel method for extracting a boundary surface $\mathcal{S}$ from an occupancy function $\phi:\mathbb{R}^3 \rightarrow \{0, 1\}$. The core idea of our method is to locate 1D and 3D points without assuming linearity in the function or relying on gradients as surface normals. Instead, we fully leverage the capabilities of neural implicit functions, enabling GPU parallelization to evaluate the function at thousands of query points within several milliseconds.

Below, we first describe how we modify the 1D point search (Section~\ref{sec:method_1d_search}) using a binary search. For the 3D point identification, which involves defining a local flat plane at each 1D point, we identify additional auxiliary points, termed \emph{2D points}, assumed to be on the same local plane. We begin by introducing our novel approach to the 2D point search (Section~\ref{sec:method_2d_search}), followed by an explanation of how the 3D points are determined based on these 2D points (Section~\ref{sec:method_3d_search}). Additionally, we explain how we improve the final polygonization to ensure the resulting mesh is intersection-free in practice (Section~\ref{sec:method_poly}).



\subsection{1D Point Search}
\label{sec:method_1d_search}
Without the linearity assumption, we locate 1D points using a \emph{binary search}. This root-finding approach, previously used by Manson and Schaefer~\shortcite{manson2010isosurfaces} to improve triangulation in their method, is adapted here to identify 1D points.
For a grid edge $e = (v^\text{in}v^\text{out})$ intersecting the surface $\mathcal{S}$, with  $\mathbf{p}_{v^\text{in}} \in \mathcal{P}^\text{in}$ and $\mathbf{p}_{v^\text{out}} \in \mathcal{P}^\text{out}$, starting with the entire edge as the search space, we iteratively halve the search space, ensuring at each step that endpoints retain distinct inside/outside labels.
This process continues until the search space is reduced below a set threshold. In our experiments at a resolution of $128^3$, the number of iterations is set to 15.


Note that with the aid of a GPU, the binary search can be performed simultaneously across \emph{all} surface-intersecting edges at the same time. For example, in our experiments with a resolution of $128^3$ (Section~\ref{sec:unconditional_results}), the number of surface-intersecting edges is approximately $16k$, which can be processed in parallel using a 12GB VRAM GPU. A single parallel evaluation of the occupancy function takes just 0.013 seconds. Therefore, the 1D point search, conducted over 15 iterations, takes roughly 0.2 seconds to complete.

\subsection{2D Point Search}
\label{sec:method_2d_search}


\textcolor{color_4}{The Quadric Error Function (Equation~\ref{eq:mdc_qef}) assumes local flatness, meaning it presumes that local flat planes continue within a grid cell, thereby determining the 3D point at the intersection of these planes. For this to work correctly, the normals of the planes should properly approximate the local geometry within the grid cell. However, Manifold Dual Contouring (MDC) uses the gradient of the implicit function as the normal for the flat plane, which fits the infinitesimal region but may not represent the surface within the grid cell.
This may result in incorrect identification of 3D points, as illustrated in Figure~\ref{fig:benefit_2d_search_illustration}. Therefore, we search auxiliary points, termed 2D points, to better approximate the local surface normal by capturing the actual local geometry within each grid face, thereby providing suitable local flat planes that approximate the local geometries of different grid cells.}

\begin{figure}[!t]
  \centering
  \includegraphics[width=\linewidth]{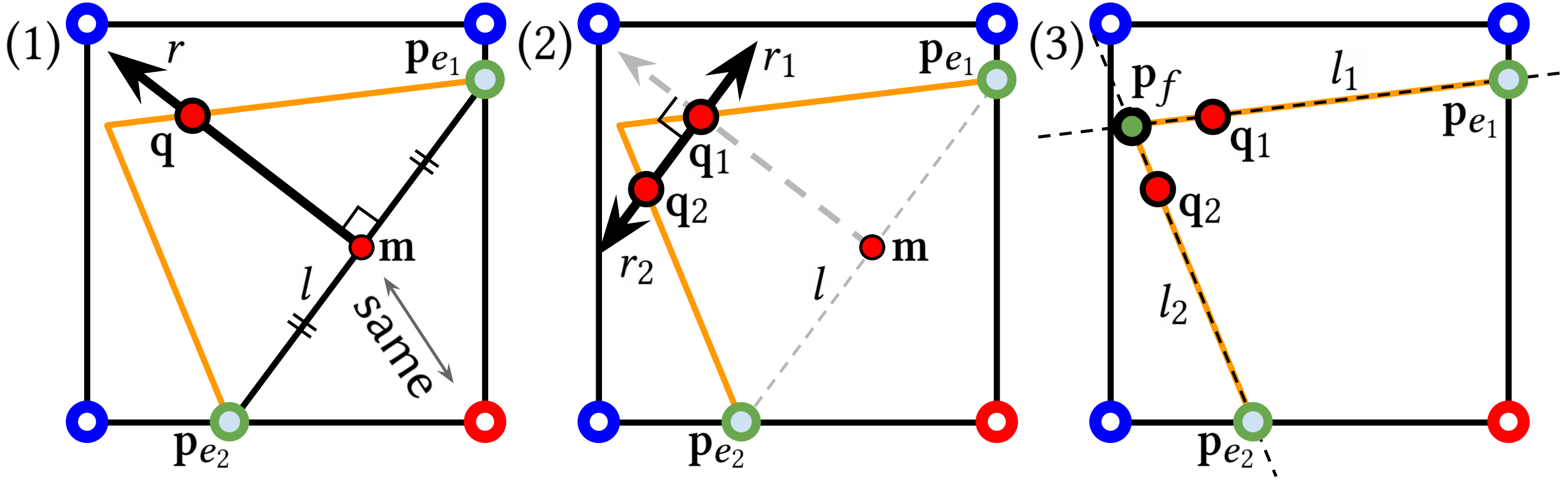}
  \vspace{-0.7cm}
  \caption{
  Illustration of the 2D point search. The orange line denotes the ground truth isocurve in 2D, and the dots with blue, red and green boundary denotes grid vertices outside, inside and 1D points, respectively. Please refer to the explanation on Section~\ref{sec:method_2d_search}.
  }
  \label{fig:2d_search}
  \vspace{-\baselineskip}
\end{figure}

Consider a pair of 1D points $\mathbf{p}_{e_1}$ and $\mathbf{p}_{e_2}$ on the same grid face $f$. Assuming that the surface $\mathcal{S}$ is locally flat at these points, the 2D point $\mathbf{p}_f$ is defined at the intersection of the plane extending the grid face $f$ with the flat planes at $\mathbf{p}_{e_1}$ and $\mathbf{p}_{e_2}$. We identify the intersection point as follows (see Figure~\ref{fig:2d_search} for an illustration):

\begin{enumerate}[leftmargin=*,noitemsep,topsep=0em]
\item Let $l=\overleftrightarrow{\mathbf{p}_{e_1}\mathbf{p}_{e_2}}$ and $\mathbf{m}$ denote the line connecting the two 1D points and their middle point, respectively. Define $r$ as a ray originating from $\mathbf{m}$ with the direction orthogonal to $l$; the direction is chosen towards the side where a grid vertex has a different inside/outside label compared to $\mathbf{m}$'s label. Detect the closest surface-crossing point $\mathbf{q}$ along $r$. If $\mathbf{q}$ coincides with $\mathbf{m}$, then $\mathbf{p}_f$ is set to be $\mathbf{m}$.

\item Starting at point $\mathbf{q}$, shoot two rays, $r_1$ and $r_2$, in opposite directions and parallel to line $l$. Ray $r_1$ is directed towards $\mathbf{p}_{e_1}$ from $\mathbf{m}$, and ray $r_2$ is directed towards $\mathbf{p}_{e_2}$ from $\mathbf{m}$. Find the closest intersection point of each ray with the surface $\mathcal{S}$, denoting these as $\mathbf{q}_1$ and $\mathbf{q}_2$, respectively. (Note that, under the flat plane assumption, either $\mathbf{q}_1$ or $\mathbf{q}_2$ coincides with $\mathbf{q}$.)

\item Determine $\mathbf{p}_f$ as the intersection of two lines: $l_1=\overleftrightarrow{\mathbf{p}_{e_1}\mathbf{q}_1}$ and $l_2=\overleftrightarrow{\mathbf{p}_{e_2}\mathbf{q}_2}$.
\end{enumerate}

The procedure produces the desired 2D point $\mathbf{p}_f$ under the local flatness assumption, as outlined in the following lemma:

\begin{lemma}
    \label{lem:2d_search}
    Let $l_1$ and $l_2$ be lines that pass distinct 1D points $\mathbf{p}_{e_1}$ and $\mathbf{p}_{e_2}$, respectively. $l_1$ and $l_2$ are identical or meet at a unique point $\mathbf{u}$ which is neither $\mathbf{p}_{e_1}$ nor $\mathbf{p}_{e_2}$. Then, \textcolor{color_4}{the 2D point $\mathbf{p}_f$ is chosen as either} the middle point $\mathbf{m}$ on the former case or the intersection $\mathbf{u}$ on the latter case.
\end{lemma}

\begin{proof}
\ifpaper See the appendix (Section~\ref{supp_sec:proof_lemma_1}).
\fi
\end{proof}

\begin{figure}[!t]
    \vspace{0.2cm}
    \centering
    \includegraphics[width=0.85\linewidth]{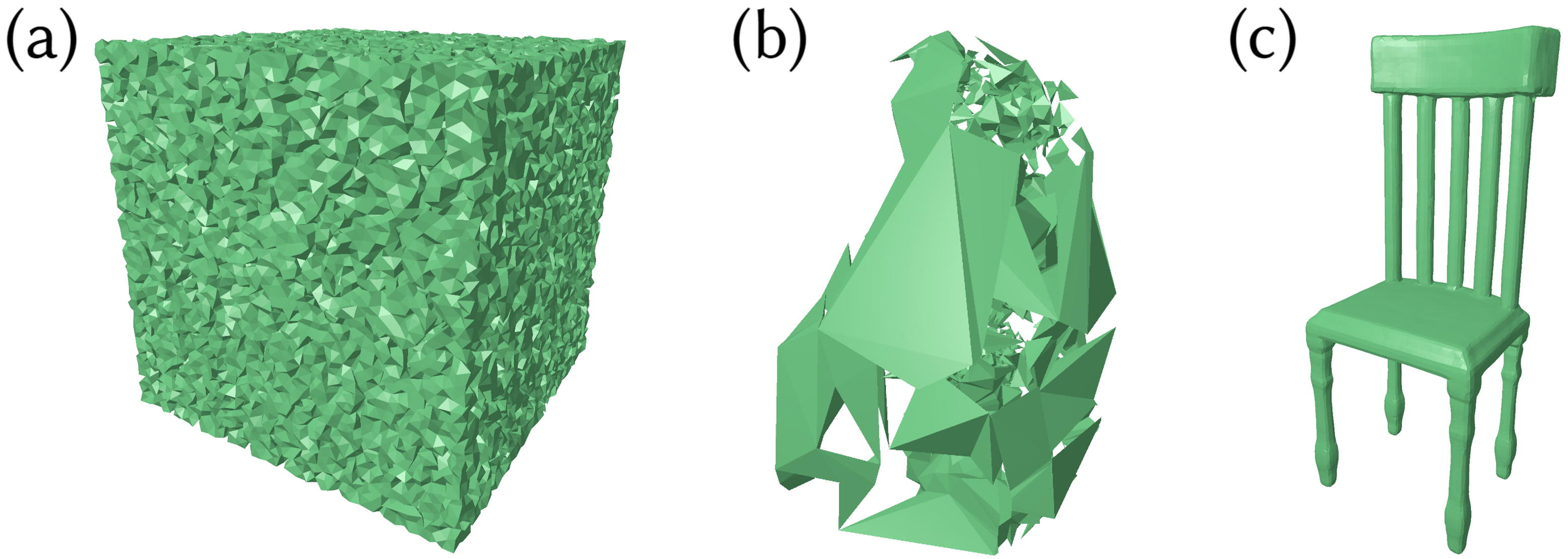}
    \vspace{-0.2cm}
    \caption{Qualitative results with SALAD. (a) Reach for the Spheres~\cite{sellan2023reach}: inflation and shrinkwrap method, (b) PoNQ~\cite{maruani2024ponq}: learning-based method, and (c) ODC (ours). Recent SDF-based methods fail to construct meshes with occupancy functions.}
    \label{fig:RFS_PoNQ_ODC}
    \vspace{-\baselineskip}
\end{figure}

Note that steps 1 and 2 involve detecting the closest intersection point with the surface $\mathbf{S}$ along a ray. This task cannot be achieved with a binary search as done in 1D point search, since we need to find the closest point from the ray's origin. Thus, we incorporate a combination of line search and binary search, termed \emph{line-binary search}. This method first samples points along the ray at uniform intervals to locate the initial interval where the endpoints have different inside/outside labels. Once such an interval is identified, a binary search is conducted within it to precisely refine the position of the intersection point.
\ifpaper More details are provided in the appendix (Section~\ref{supp_sec:line_binary_search}).
\fi

Like the 1D point search, the 2D point search can also be executed simultaneously for all surface-crossing faces using GPU parallelization. For example, in our experiments with a resolution of $128^3$ (Section~\ref{sec:unconditional_results}), the number of surface-intersecting faces is roughly $33k$, all of which can be processed in parallel using a 12GB VRAM GPU.
The 2D point search algorithm includes three line-binary searches for rays $r$, $r_1$, and $r_2$. We take 4 samples for the line search and perform 11 iterations of the binary search, totaling 
$(4+11)\times3=45$ evaluations of the occupancy function. Since a single evaluation takes about 0.03 seconds, the 2D point search for the entire grid can be completed within 1.3 seconds.



\subsection{3D Point Identification}
\label{sec:method_3d_search}
Each 1D point along a grid edge derives two 2D points from the adjacent grid faces. Since the two 2D points lie on the same flat plane as the 1D point, which locally represents the surface $\mathcal{S}$, the plane's normal can now be calculated from these three points. This replaces the gradient of the input implicit function in Equation~\ref{eq:mdc_qef}. The 3D points within each cell are then determined as done in Manifold Dual Contouring (Section~\ref{sec:mdc_3d_search}).



\subsection{Polygonization}
\label{sec:method_poly}
While our main goal is to adapt Manifold Dual Contouring (MDC) to perform with an occupancy function by replacing the 1D point search and 3D point identification, we additionally improve the MDC algorithm to overcome its limitation, often resulting in self-intersections in the output mesh. We incorporate the quad splitting technique from Intersection-Free Contouring (IC)~\cite{ju2006intersection} to resolve this issue. IC was originally designed to handle a unique 3D point within each cell; we extend this idea to accommodate multiple 3D points per cell.
\ifpaper We describe the details in the appendix (Section~\ref{supp_sec:polygonization}) due to space constraints.
\fi

\begin{figure*}[!]
    \centering
    \setlength{\tabcolsep}{0em}
    \def\arraystretch{0.0}

    \begin{tabular}{P{0.125\textwidth}P{0.125\textwidth}P{0.125\textwidth}P{0.125\textwidth}P{0.125\textwidth}P{0.125\textwidth}P{0.125\textwidth}P{0.125\textwidth}}
        \textbf{MC}~\shortcite{lewiner2003efficient} & \textbf{IC}~\shortcite{ju2006intersection} & \textbf{MDC}~\shortcite{manifolddualcontouring} & \textbf{LMC}~\shortcite{lempitsky2010surface} & \textbf{NDC}~\shortcite{neuraldualcontouring} & \textbf{CDIF}~\shortcite{manson2011contouring} & \textbf{MISE}~\shortcite{occupancenetworks} & \textbf{ODC} (ours) \\
        \midrule
        \multicolumn{8}{c}{\emph{Qualitative results with SALAD.}} \\
        \multicolumn{8}{c}{\includegraphics[width=\textwidth]{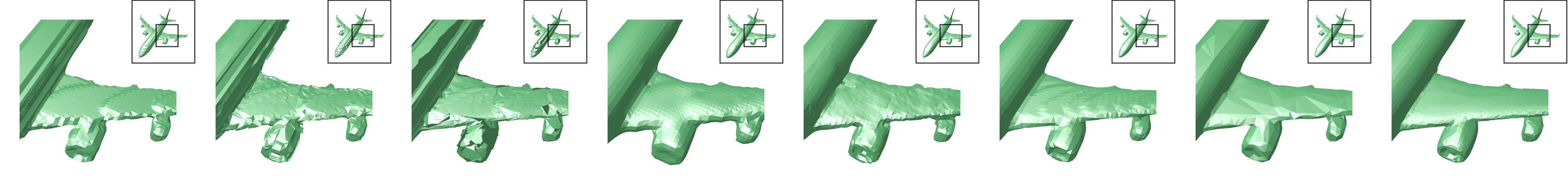}} \\
        \multicolumn{8}{c}{\includegraphics[width=\textwidth]{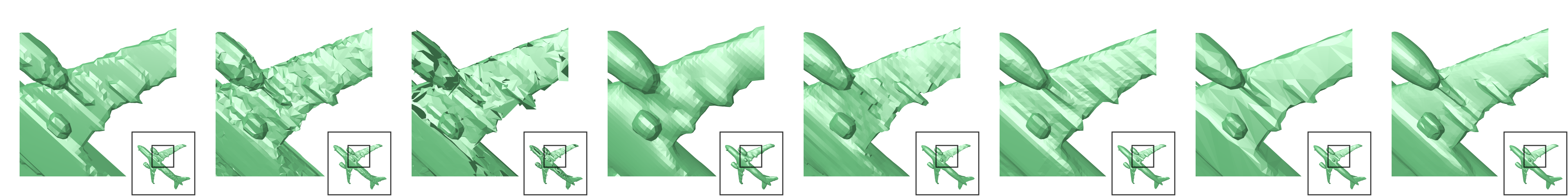}} \\
        \multicolumn{8}{c}{\includegraphics[width=\textwidth]{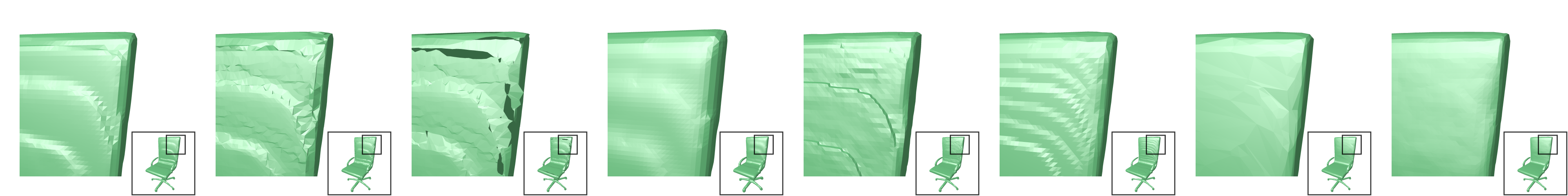}} \\
        \multicolumn{8}{c}{\includegraphics[width=\textwidth]{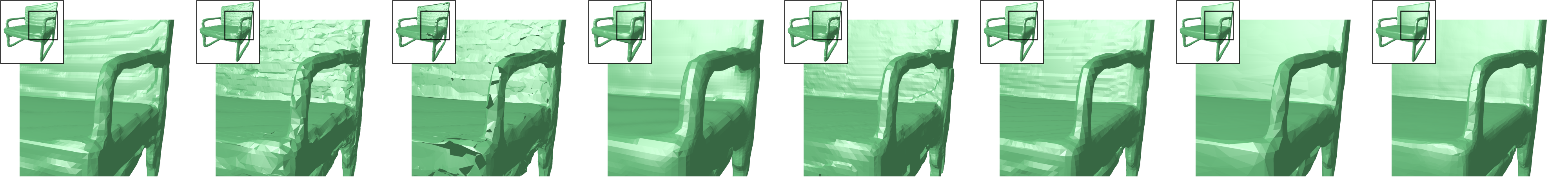}} \\
        \multicolumn{8}{c}{\includegraphics[width=\textwidth]{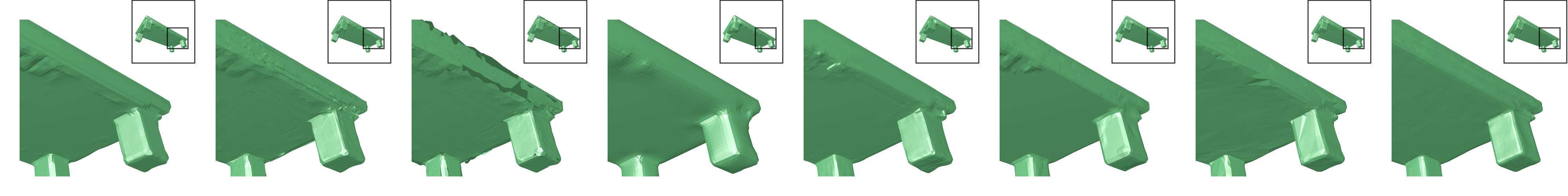}} \\
        \midrule
        \multicolumn{8}{c}{\emph{Qualitative results with 3DShape2VecSet.}} \\
        \multicolumn{8}{c}{\includegraphics[width=\textwidth]{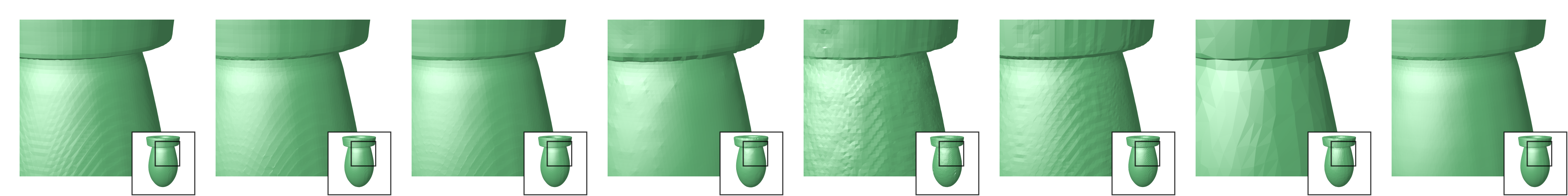}} \\
        \multicolumn{8}{c}{\includegraphics[width=\textwidth]{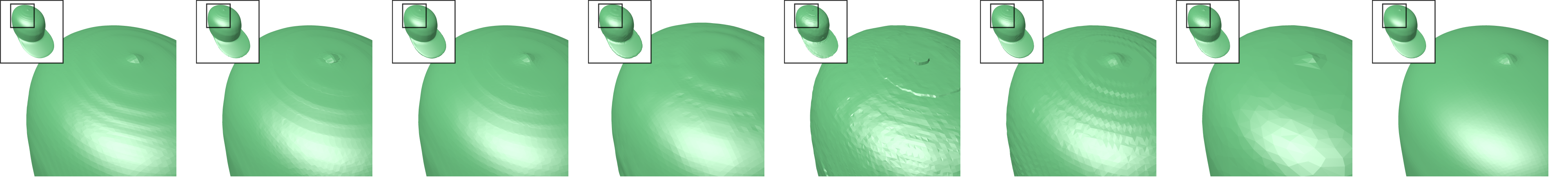}} \\
        \multicolumn{8}{c}{\includegraphics[width=\textwidth]{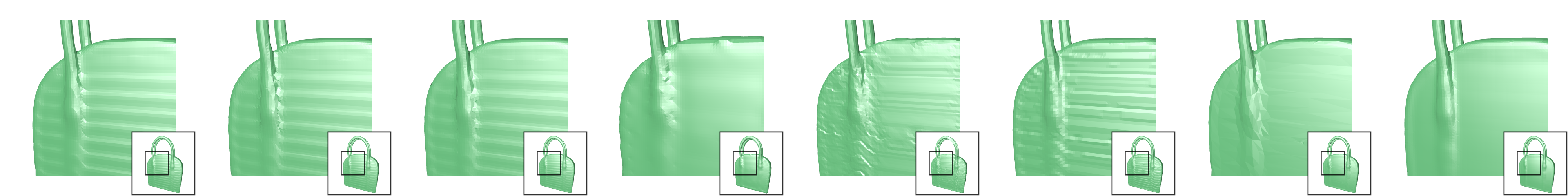}} \\
        \multicolumn{8}{c}{\includegraphics[width=\textwidth]{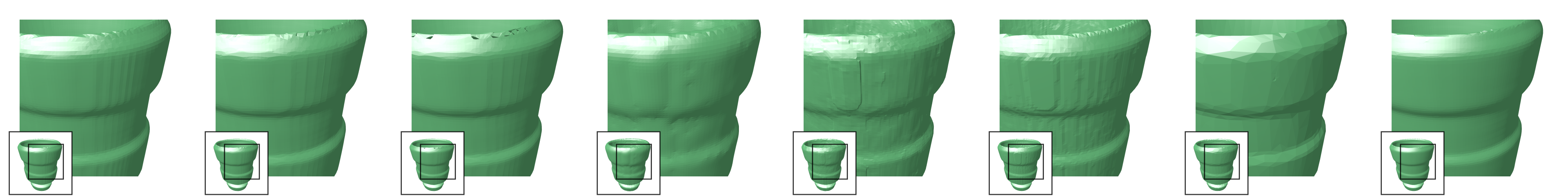}}
    \end{tabular}
    \vspace{-\baselineskip}
    \caption{Qualitative results with SALAD and 3DShape2VecSet. ODC effectively captures the fine details, avoiding the aliasing effects seen in existing methods. Notably, since ODC focuses on precisely converting the functions rather than visual plausibility, it shows the true shapes that the generative models yield.}
    \label{fig:SALAD_3DShape2VecSet_result}
\end{figure*}

\section{Results}
\label{sec:results}




In this section, we present experimental results for our method, ODC. For all the following experiments, we fixed the resolution at $128^3$.
\ifpaper Refer to the appendix (Section~\ref{supp_sec:salad_scalability}) for results with other resolutions.
\fi
We compare our results with several baselines that take \emph{occupancy functions} as input: Lempitsky's binary volume smoothing method (LMC) \cite{lempitsky2010surface}, Neural Dual Contouring with binary voxel input (NDC, specifically, NDCx) \cite{neuraldualcontouring}, Contouring DIF (CDIF) \cite{manson2011contouring}, and Multiresolution IsoSurface Extraction (MISE) \cite{occupancenetworks}. We also compare our method with the base methods of our framework: Marching Cubes (MC, specifically, the topology-preserved version \cite{lewiner2003efficient}), Intersection-free Contouring (IC) \cite{ju2006intersection}, and Manifold Dual Contouring (MDC) \cite{manifolddualcontouring}. For Neural Dual Contouring (NDC), we use the official pretrained models provided by the authors, trained on the ABC dataset \cite{koch2019abc}. For Neural Dual Contouring, the improved version with the Neural Marching Cubes \cite{neuralmarchingcubes} backbone is used. Note that MISE produces much fewer triangles than the other methods due to the mesh simplification step involved. For results of MISE without the mesh simplification step,
\ifpaper refer to Section~\ref{supp_sec:MISE_mesh_simplification} in the appendix.
\fi

We additionally tested Reach for the Spheres~\cite{sellan2023reach}, the most recent inflation and shrinkwrap method, and PoNQ~\cite{maruani2024ponq}, a learning-based method whose model is trained with signed distance values. However, these methods were unstable for occupancy functions and produced poor results, as shown in Figure~\ref{fig:RFS_PoNQ_ODC}. Therefore, we omitted them from our comparison.



\subsection{Neural Implicit Functions and Datasets}

We conduct our experiments with occupancy functions generated by the following 3D generative models and a 3D mesh dataset.

\begin{enumerate}[leftmargin=*,noitemsep,topsep=0em]

\item SALAD~\cite{koo2023salad}: A 3D generative model was trained with each of three classes of ShapeNet: Airplane, Chair, and Table. We use 500 generated MLP-based occupancy functions per class, for a total of 1,500.

\item 3DShape2VecSet~\cite{zhang20233dshape2vecset}: Another 3D generative model was trained with 55 categories of ShapeNet. We use a total of 1,650 generated occupancy functions, comprising 30 shapes for each category.

\item Michelangelo~\cite{zhao2024michelangelo}: A conditional 3D generative model that accepts either text, images, or both as inputs. We use the official pretrained model along with the provided 32 images and 29 texts as inputs.



\item Myles' dataset~\cite{myles2014robust}: A 3D mesh dataset including 79 meshes, excluding some skinny shapes (excluded by previous work~\cite{neuralmarchingcubes, neuraldualcontouring}). Occupancy functions are derived by computing winding numbers~\cite{barill2018fast} from the meshes.
This dataset is specifically used to measure similarity not only to the input occupancy function but also to the original 3D mesh.
\end{enumerate}

While MLP-based occupancy functions output the probability of occupancy within the range $[0, 1]$, other models (3DShape2VecSet and Michelangelo) do not constrain their output ranges, although they still encode a shape at the isolevel of 0.5. In our experiments, occupancy-based methods (LMC, NDC, CDIF, and ours) take binary occupancy values, either 0 or 1. In contrast, models designed for SDF (MC, IC, and MDC) or continuous functions (MISE) use the raw values directly from the network. To accommodate all cases, we define $\phi$ using the raw values of the MLP in this section. 

\ifpaper In the appendix, we present additional results with Michelangelo~\cite{zhao2024michelangelo}, IM-NET~\cite{imnet}, and DeepSDF
\cite{deepsdf} in Section~\ref{supp_sec:michelangelo_results},~\ref{supp_sec:imnet_results}, and~\ref{supp_sec:deepsdf_results}, respectively.
More quantitative and qualitative results are also included in Section~\ref{supp_sec:more_results}.

\fi

\begin{table}[!t]
    \setlength{\tabcolsep}{2.0pt}
    \small
    \centering
    \caption{Quantitative results with unconditional generation of SALAD~\cite{koo2023salad} and 3DShape2VecSet~\cite{zhang20233dshape2vecset}. The values are the average of all categories.}
    \label{tab:salad_3dshape2vecset}
    \vspace{-\baselineskip}
    \begin{tabular}{ccccc|cccc}
        \toprule
        & \multicolumn{4}{c}{\textbf{SALAD}} & \multicolumn{4}{c}{\textbf{3DShape2VecSet}} \\
        \textbf{Uncond.} & \multirow{2}{*}{$|\phi - 0.5|$ $\downarrow$} & SI $\downarrow$ & Man. $\uparrow$ & Time & \multirow{2}{*}{$|\phi - 0.5|$ $\downarrow$}  & SI $\downarrow$ & Man. $\uparrow$ & Time \\
        \textbf{Gen.} & & (\%) & (\%) & (s) & & (\%) & (\%) & (s) \\
        \midrule
        MC & 0.324 & 0.00 & 100. & 1.9 & 2.330 & 0.00 & 100. & 0.4 \\
        IC & 0.329 & 0.00 & 21.8 & 1.9 & 2.292 & 0.00 & 53.2 & 0.5 \\
        MDC & 0.305 & 100. & 100. & 2.0 & 2.245 & 100. & 100. & 0.5 \\
        LMC & 0.484 & 0.00 & 56.8 & 4.4 & 8.824 & 0.00 & 72.8 & 9.8 \\
        NDC & 0.324 & 33.8 & 37.1 & 0.6 & 3.552 & 21.2 & 53.2 & 1.0 \\
        CDIF & 0.234 & 0.00 & 100. & 4.4 & 1.961 & 0.00 & 100. & 1.4 \\
        MISE & 0.237 & 23.1 & 74.2 & 1.2 & 1.733 & 22.3 & 82.0 & 1.8 \\
        \rowcolor{Gray}
        ODC & \textbf{0.037} & 0.00 & 100. & 3.5 & \textbf{0.112} & 0.06 & 100. & 1.7 \\
        \bottomrule
    \end{tabular}
    \vspace{-0.55cm}
\end{table}


\subsection{Evaluation Metrics}
\label{sec:evaluation_metrics}
When extracting a mesh from a neural occupancy function without the ground truth 3D mesh, we measure the following metrics:
\begin{itemize}[leftmargin=*,noitemsep,topsep=0em]
\item $|\phi - 0.5|$, measuring the alignment with the surface of the given occupancy function.
\item Self-Intersection (SI)\textcolor{color_4}{, the ratio of self-intersecting meshes.}
\item Manifoldness (Man.)\textcolor{color_4}{, the ratio of manifold meshes.}
\end{itemize}

In the experiments with Myles' dataset, where we obtain the occupancy function from a ground truth 3D mesh, we additionally measure the following metrics to compare the output mesh with the ground truth mesh:
\begin{itemize}[leftmargin=*,noitemsep,topsep=0em]
\item Mesh Distance (MD2), the squared mesh distance from a point cloud sampled from another mesh in both directions.
\item Normal InConsistency (NIC), the radian angle difference of normals between the the ground truth and the output.
\item HausDorff Distance (HDD) between the ground truth and the output.
\end{itemize}

In all experiments, we sample $2M$ points per shape for metrics that require point samples. Note that SI and Man. are not per-shape quantities but rather ratios of output meshes. As for the other metrics, we report the average of per-shape quantities. We also report the number of vertices (\#V) and triangles (\#T) of the output mesh along with the runtime. All the runtimes \textcolor{color_4}{including the occupancy function inference time} were measured with a single NVIDIA A6000 GPU, 48-core Intel CPUs, and 504GB RAM.

\subsection{Results with SALAD and 3DShape2VecSet}
\label{sec:unconditional_results}
The qualitative results of the experiment on unconditional generation using SALAD~\cite{koo2023salad} and 3DShape2VecSet~\cite{zhang20233dshape2vecset} are presented in Figure~\ref{fig:SALAD_3DShape2VecSet_result}. The output from MC~\cite{lewiner2003efficient}, IC~\cite{ju2006intersection} and CDIF~\cite{manson2011contouring} produces artifacts with staircase-like local patterns, while MDC~\cite{manifolddualcontouring} produces self-intersections of mesh faces, indicated by darker triangles in the figure, especially in the fourth and ninth rows. LMC~\cite{lempitsky2010surface} generates overly smoothed shapes, as seen in the second row. Additionally, NDC~\cite{neuraldualcontouring}, trained on occupancy functions from the ABC dataset~\cite{koch2019abc}, demonstrates limited generalizability to ShapeNet-based models generated by SALAD and 3DShape2VecSet, evident from the angular patterns on slightly inclined geometry in the third and eighth rows. MISE~\cite{occupancenetworks} displays cracked surfaces and non-crooked edges in the first and fifth row. In contrast, our ODC consistently produces meshes with much more plausible shapes.


The quantitative results are presented in Table~\ref{tab:salad_3dshape2vecset}. Regarding the fidelity of the output mesh to the given implicit function, our ODC achieves more than 6 times and 15 times smaller fitting errors $|\phi - 0.5|$ on average for the same resolution ($128^3$). Additionally, all 1,500 ODC output meshes in the SALAD experiment are free of self-intersections and have manifold properties. In the 3DShape2VecSet experiment, all 1,650 ODC output meshes are manifold, with only one exhibiting self-intersection, a significant improvement over our backbone MDC, which has 100\% of meshes with self-intersections. The runtime of ODC is roughly 1.8 times slower than MC in the SALAD experiment, and in the 3DShape2VecSet experiment, ODC is four times slower than MC but still under 2 seconds.

\begin{table}[!t]
    \setlength{\tabcolsep}{5.0pt}
    \small
    \centering
    \caption{Quantitative results with Myles' dataset~\cite{myles2014robust}. The input is given by occupancy. $^\dagger$Baselines that require gradient of SDF for running.}
    \label{tab:myles_occ}
    \vspace{-\baselineskip}
    \begin{tabular}{cccccccc}
        \toprule
        \textbf{Myles'} & MD2 $\downarrow$ & \multirow{2}{*}{NIC $\downarrow$} & HDD $\downarrow$ & SI $\downarrow$ & Man. $\uparrow$ & \multirow{2}{*}{\#V}& \multirow{2}{*}{\#T} \\
        \textbf{Occ.} & ($\times 10^6$) &  & ($\times 10^2$) & (\%) & (\%) &  & \\
        \midrule
        MC & 2.261 & 0.366 & 0.894 & 0.00 & 100. & 38963 & 77944 \\
        $^\dagger$IC & 2.808 & 0.583 & 0.959 & 0.00 & 11.4 & 40095 & 80267 \\
        $^\dagger$MDC & 1.857 & 1.042 & 2.309 & 100. & 100. & 38961 & 77926 \\
        LMC & 32.06 & 0.146 & 2.091 & 0.00 & 59.5 & 34213 & 68429 \\
        NDC & 1.493 & 0.163 & 1.521 & 36.7 & 13.9 & 38555 & 77156 \\
        CDIF & 0.562 & 0.121 & 1.043 & 0.00 & 100. & 38423 & 76855 \\
        $^\dagger$MISE & 2.124 & 0.168 & 1.212 & 10.1 & 35.4 & 2496 & 5000 \\
        \rowcolor{Gray}
        ODC & \textbf{0.113} & \textbf{0.072} & \textbf{0.632} & 0.00 & 100. & 39070 & 78142 \\
        \bottomrule
    \end{tabular}
    \vspace{-0.5cm}
\end{table}

\begin{figure*}[!]
    \centering
    \setlength{\tabcolsep}{0em}
    \def\arraystretch{0.0}

    \begin{tabular}{P{0.111\textwidth}P{0.111\textwidth}P{0.111\textwidth}P{0.111\textwidth}P{0.111\textwidth}P{0.111\textwidth}P{0.111\textwidth}P{0.111\textwidth}P{0.112\textwidth}}
        \textbf{MC}~\shortcite{lewiner2003efficient} & \textbf{$^\dagger$IC}~\shortcite{ju2006intersection} & \textbf{$^\dagger$MDC}~\shortcite{manifolddualcontouring} & \textbf{LMC}~\shortcite{lempitsky2010surface} & \textbf{NDC}~\shortcite{neuraldualcontouring} & \textbf{CDIF}~\shortcite{manson2011contouring} & \textbf{$^\dagger$MISE}~\shortcite{occupancenetworks} & \textbf{ODC} (ours)  & \textbf{GT}\\
        \midrule
        \multicolumn{9}{c}{\includegraphics[width=\textwidth]{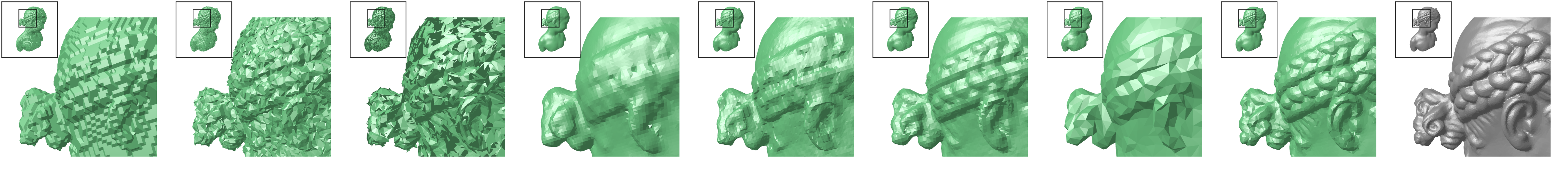}} \\
        \multicolumn{9}{c}{\includegraphics[width=\textwidth]{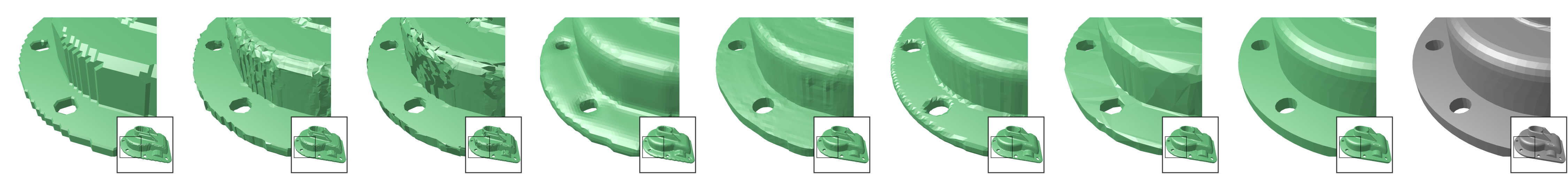}} \\
        \multicolumn{9}{c}{\includegraphics[width=\textwidth]{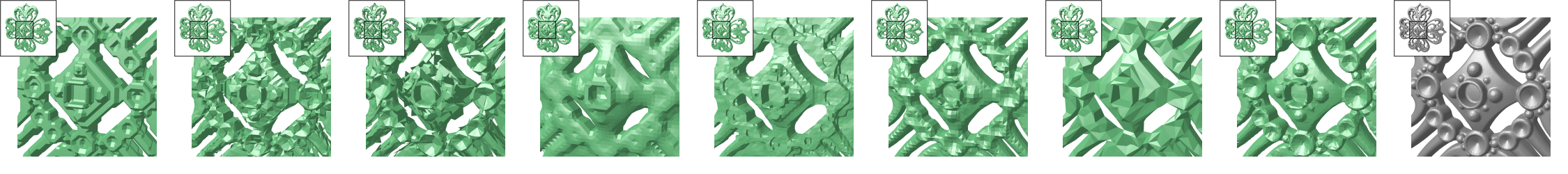}} \\
        \multicolumn{9}{c}{\includegraphics[width=\textwidth]{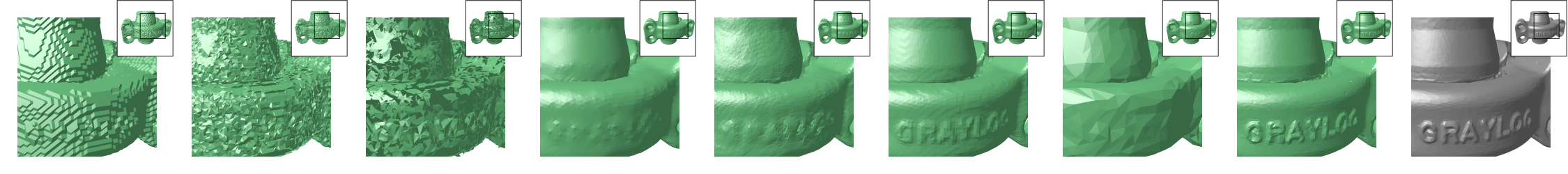}}
    \end{tabular}
    \vspace{-\baselineskip}
    \caption{Qualitative results with Myles'~\cite{myles2014robust} dataset. Input is an occupancy function derived from the winding number computation. $^\dagger$The baselines also require the gradient of the SDF due to their specifications. ODC outperforms other methods in capturing fine details, including sharp features.}
    \label{fig:Myles_occ_result}
    \vspace{-0.2cm}
\end{figure*}



\begin{figure}[!]
    \centering
    \setlength{\tabcolsep}{0em}
    \def\arraystretch{0.0}

    \begin{tabular}{P{0.25\linewidth}P{0.25\linewidth}P{0.25\linewidth}P{0.25\linewidth}}
        \textbf{MDC}~\shortcite{manifolddualcontouring} & \textbf{+ 1D Search} & \textbf{+ 2D Search} & \textbf{+IC (=ODC)} \\
        \midrule
        \multicolumn{4}{c}{\includegraphics[width=\linewidth]{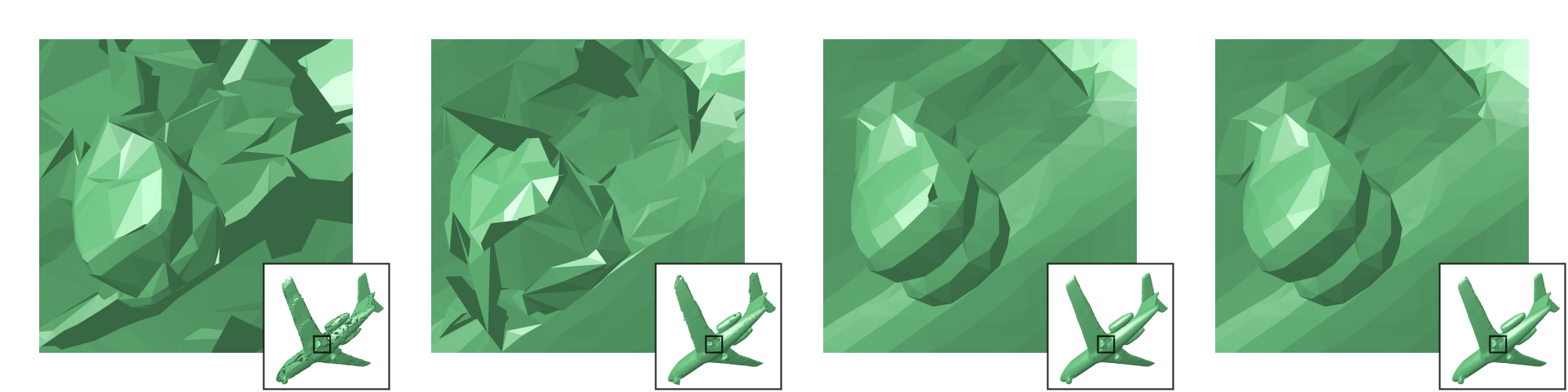}}
    \end{tabular}
    \vspace{-0.3cm}
    \caption{\textcolor{color_4}{Ablation study with SALAD~\cite{koo2023salad}.
    Linearity denotes that 1D points are calculated by Equation~\ref{eq:1d_iso-level}. IC indicates the quad splitting technique of IC~\cite{ju2006intersection}. Darker triangles represent flipped triangles identicating self-intersection.
    }}
    \label{fig:benefit_2d_search}
    \vspace{-0.4cm}
\end{figure}


\subsection{Results with Myles' Dataset}
\label{sec:imnet_results}
Figure~\ref{fig:Myles_occ_result} presents qualitative comparisons with the 79 meshes collected by Myles et al.~\cite{myles2014robust}, testing the methods with pure occupancy functions. MC~\cite{lewiner2003efficient} tends to produce excessive staircase-like artifacts. IC~\cite{ju2006intersection} and MDC~\cite{manifolddualcontouring} present broken surfaces and the other methods smooth out fine details. In contrast, our ODC shows a significant advantage in accurately capturing fine details, including sharp features, as shown in the second row.

The quantitative comparisons shown in Table~\ref{tab:myles_occ} also demonstrate the superior performance of our method, providing the best results in all evaluation metrics, with comparable number of vertices and triangles with the other methods.

\subsection{Ablation Study with SALAD}
\label{sec:salad_ablation}
\textcolor{color_4}{Figure~\ref{fig:benefit_2d_search} illustrates how the mesh extraction of a wheel of an airplane improves with each step of our method. When the linearity assumption is applied to 1D point identification, it often leads to inaccurate 1D points, resulting in a cracked surface (1st column). Although the 1D point search mitigates this issue to some extent, the use of gradients still results in a cracked surface (2nd column). In contrast, the 2D point search captures finer details and produces a more accurate output (3rd column). Self-intersections, indicated by the darker triangles, are further mitigated by the IC-based polygonization, as demonstrated in our full method (4th column).}

\textcolor{color_4}{The quantitative results of our ablation study with SALAD~\cite{koo2023salad} are presented in Table~\ref{tab:benefit_2d_search}, demonstrating the effectiveness of our 1D and 2D point search and polygonization methods. When the 1D point search is applied, the fitting error $|\phi - 0.5|$ is reduced by a factor of three. Employing the 2D point search further reduces the error by an additional factor of approximately three. This indicates that while accurately locating 1D points enhances extraction fidelity, the local surface normal calculated from 2D points also contributes to better mesh extraction compared to using gradients as local surface normals. Finally, our full method, which incorporates the quad splitting technique, completely eliminates self-intersections.}

\begin{table}[!t]
    \setlength{\tabcolsep}{3.1pt}
    \small
    \centering
    \vspace{0.0cm}
    \caption{\textcolor{color_4}{Ablation study with SALAD~\cite{koo2023salad}. IC indicates quad splitting technique of IC~\cite{ju2006intersection}.}}
    \label{tab:benefit_2d_search}
    \vspace{-\baselineskip}
    \begin{tabular}{ccccccc}
        \toprule
        \textbf{Ablation} & \multirow{2}{*}{$|\phi - 0.5|$ $\downarrow$} & \multirow{2}{*}{SI (\%) $\downarrow$} & \multirow{2}{*}{Man. (\%) $\uparrow$} & \multirow{2}{*}{\#V} & \multirow{2}{*}{\#T} & \multirow{2}{*}{Time (s)} \\
        \textbf{SALAD} & & & & & & \\
        \midrule
        MDC & 0.305 & 100. & 100. & 16286 & 32570 & 2.0 \\
        + 1D Search & 0.109 & 100. & 100. & 16286 & 32570 & 2.2 \\
        + 2D Search & 0.038 & 100. & 100. & 16286 & 32570 & 3.5 \\
        \rowcolor{Gray}
        + IC (=ODC) & \textbf{0.037} & \textbf{0.00} & 100. & 16310 & 32619 & 3.5 \\
        \bottomrule
    \end{tabular}
    \ifpaper \vspace{-0.5cm}
    \fi
\end{table}

\section{Conclusion}
\label{sec:conclusion}

We have introduced Occupancy-Based Dual Contouring (ODC), a dual contouring method specialized for neural occupancy functions. Our core idea was the design of 1D and 2D point search algorithms, all of which fully leverage GPU acceleration while parallelizing computation across all grid edges, faces, and cells. 3D points were particularly identified by calculating surface normals in the quadric error function with newly introduced auxiliary 2D points defined on the grid faces. The 2D point search algorithm was developed to guarantee finding the points on the same plane as the 1D point under the local flatness assumption, simply using three ray searches.
Experimental results with several 3D generative models producing 3D shapes as occupancy functions, as well as with a 3D mesh dataset converted to occupancy functions, demonstrated the state-of-the-art fidelity of our method with a few-second computation time. For future work, we plan to further improve our method to apply to adaptive resolutions and resolve topology ambiguity.

\section*{Acknowledgements}
This work was supported by the NRF grant (RS-2023-00209723), IITP grants (RS-2022-II220594, RS-2023-00227592, RS-2024-00399817), and KEIT grant (RS-2024-00423625), all funded by the Korean government (MSIT and MOTIE), as well as grants from the DRB-KAIST SketchTheFuture Research Center, NAVER-Intel, Hyundai NGV, KT, and Samsung Electronics.

\bibliographystyle{ACM-Reference-Format}
\bibliography{bibfile}


\begin{thebibliography}{90}


\ifx \showCODEN    \undefined \def \showCODEN     #1{\unskip}     \fi
\ifx \showDOI      \undefined \def \showDOI       #1{#1}\fi
\ifx \showISBNx    \undefined \def \showISBNx     #1{\unskip}     \fi
\ifx \showISBNxiii \undefined \def \showISBNxiii  #1{\unskip}     \fi
\ifx \showISSN     \undefined \def \showISSN      #1{\unskip}     \fi
\ifx \showLCCN     \undefined \def \showLCCN      #1{\unskip}     \fi
\ifx \shownote     \undefined \def \shownote      #1{#1}          \fi
\ifx \showarticletitle \undefined \def \showarticletitle #1{#1}   \fi
\ifx \showURL      \undefined \def \showURL       {\relax}        \fi
\providecommand\bibfield[2]{#2}
\providecommand\bibinfo[2]{#2}
\providecommand\natexlab[1]{#1}
\providecommand\showeprint[2][]{arXiv:#2}

\bibitem[Akkouche and Galin(2001)]%
        {akkouche2001adaptive}
\bibfield{author}{\bibinfo{person}{Samir Akkouche} {and} \bibinfo{person}{Eric Galin}.} \bibinfo{year}{2001}\natexlab{}.
\newblock \showarticletitle{Adaptive implicit surface polygonization using marching triangles}. In \bibinfo{booktitle}{\emph{Computer Graphics Forum}}.
\newblock


\bibitem[Barill et~al\mbox{.}(2018)]%
        {barill2018fast}
\bibfield{author}{\bibinfo{person}{Gavin Barill}, \bibinfo{person}{Neil~G Dickson}, \bibinfo{person}{Ryan Schmidt}, \bibinfo{person}{David~IW Levin}, {and} \bibinfo{person}{Alec Jacobson}.} \bibinfo{year}{2018}\natexlab{}.
\newblock \showarticletitle{Fast winding numbers for soups and clouds}.
\newblock \bibinfo{journal}{\emph{ACM Transactions on Graphics (TOG)}} \bibinfo{volume}{37}, \bibinfo{number}{4} (\bibinfo{year}{2018}), \bibinfo{pages}{1--12}.
\newblock


\bibitem[Bhattacharya and Wenger(2013)]%
        {bhattacharya2013constructing}
\bibfield{author}{\bibinfo{person}{Arindam Bhattacharya} {and} \bibinfo{person}{Rephael Wenger}.} \bibinfo{year}{2013}\natexlab{}.
\newblock \showarticletitle{Constructing isosurfaces with sharp edges and corners using cube merging}. In \bibinfo{booktitle}{\emph{Computer Graphics Forum}}, Vol.~\bibinfo{volume}{32}. Wiley Online Library, \bibinfo{pages}{11--20}.
\newblock


\bibitem[Bloomenthal(1988)]%
        {bloomenthal1988polygonization}
\bibfield{author}{\bibinfo{person}{Jules Bloomenthal}.} \bibinfo{year}{1988}\natexlab{}.
\newblock \showarticletitle{Polygonization of implicit surfaces}.
\newblock \bibinfo{journal}{\emph{Computer Aided Geometric Design}} \bibinfo{volume}{5}, \bibinfo{number}{4} (\bibinfo{year}{1988}), \bibinfo{pages}{341--355}.
\newblock


\bibitem[Bottino et~al\mbox{.}(1996)]%
        {bottino1996shrinkwrap}
\bibfield{author}{\bibinfo{person}{Andrea Bottino}, \bibinfo{person}{Wim Nuij}, {and} \bibinfo{person}{Kees Van~Overveld}.} \bibinfo{year}{1996}\natexlab{}.
\newblock \showarticletitle{How to shrinkwrap through a critical point: an algorithm for the adaptive triangulation of iso-surfaces with arbitrary topology}. In \bibinfo{booktitle}{\emph{Proc. Implicit Surfaces}}, Vol.~\bibinfo{volume}{96}. \bibinfo{pages}{53--72}.
\newblock


\bibitem[Boulch and Marlet(2022)]%
        {boulch2022poco}
\bibfield{author}{\bibinfo{person}{Alexandre Boulch} {and} \bibinfo{person}{Renaud Marlet}.} \bibinfo{year}{2022}\natexlab{}.
\newblock \showarticletitle{Poco: Point convolution for surface reconstruction}. In \bibinfo{booktitle}{\emph{Proceedings of the IEEE/CVF Conference on Computer Vision and Pattern Recognition}}. \bibinfo{pages}{6302--6314}.
\newblock


\bibitem[Chang et~al\mbox{.}(2015)]%
        {chang2015shapenet}
\bibfield{author}{\bibinfo{person}{Angel~X Chang}, \bibinfo{person}{Thomas Funkhouser}, \bibinfo{person}{Leonidas Guibas}, \bibinfo{person}{Pat Hanrahan}, \bibinfo{person}{Qixing Huang}, \bibinfo{person}{Zimo Li}, \bibinfo{person}{Silvio Savarese}, \bibinfo{person}{Manolis Savva}, \bibinfo{person}{Shuran Song}, \bibinfo{person}{Hao Su}, {et~al\mbox{.}}} \bibinfo{year}{2015}\natexlab{}.
\newblock \showarticletitle{Shapenet: {An} information-rich 3d model repository}.
\newblock \bibinfo{journal}{\emph{arXiv preprint arXiv:1512.03012}} (\bibinfo{year}{2015}).
\newblock


\bibitem[Chen et~al\mbox{.}(2022)]%
        {neuraldualcontouring}
\bibfield{author}{\bibinfo{person}{Zhiqin Chen}, \bibinfo{person}{Andrea Tagliasacchi}, \bibinfo{person}{Thomas Funkhouser}, {and} \bibinfo{person}{Hao Zhang}.} \bibinfo{year}{2022}\natexlab{}.
\newblock \showarticletitle{Neural dual contouring}.
\newblock \bibinfo{journal}{\emph{ACM Transactions on Graphics}} (\bibinfo{year}{2022}).
\newblock


\bibitem[Chen and Zhang(2019)]%
        {imnet}
\bibfield{author}{\bibinfo{person}{Zhiqin Chen} {and} \bibinfo{person}{Hao Zhang}.} \bibinfo{year}{2019}\natexlab{}.
\newblock \showarticletitle{{Learning Implicit Fields for Generative Shape Modeling}}. In \bibinfo{booktitle}{\emph{CVPR}}.
\newblock


\bibitem[Chen and Zhang(2021)]%
        {neuralmarchingcubes}
\bibfield{author}{\bibinfo{person}{Zhiqin Chen} {and} \bibinfo{person}{Hao Zhang}.} \bibinfo{year}{2021}\natexlab{}.
\newblock \showarticletitle{Neural marching cubes}.
\newblock \bibinfo{journal}{\emph{ACM Transactions on Graphics}} (\bibinfo{year}{2021}).
\newblock


\bibitem[Chernyaev(1995)]%
        {chernyaev1995marching}
\bibfield{author}{\bibinfo{person}{Evgeni Chernyaev}.} \bibinfo{year}{1995}\natexlab{}.
\newblock \bibinfo{booktitle}{\emph{Marching cubes 33: Construction of topologically correct isosurfaces}}.
\newblock \bibinfo{type}{{T}echnical {R}eport}.
\newblock


\bibitem[Chica et~al\mbox{.}(2008)]%
        {chica2008pressing}
\bibfield{author}{\bibinfo{person}{Antoni Chica}, \bibinfo{person}{Jason Williams}, \bibinfo{person}{Carlos And{\'u}jar}, \bibinfo{person}{Pere Brunet}, \bibinfo{person}{Isabel Navazo}, \bibinfo{person}{Jarek Rossignac}, {and} \bibinfo{person}{{\`A}lvar Vinacua}.} \bibinfo{year}{2008}\natexlab{}.
\newblock \showarticletitle{Pressing: Smooth isosurfaces with flats from binary grids}. In \bibinfo{booktitle}{\emph{Computer Graphics Forum}}, Vol.~\bibinfo{volume}{27}. Wiley Online Library, \bibinfo{pages}{36--46}.
\newblock


\bibitem[Custodio et~al\mbox{.}(2013)]%
        {custodio2013practical}
\bibfield{author}{\bibinfo{person}{Lis Custodio}, \bibinfo{person}{Tiago Etiene}, \bibinfo{person}{Sinesio Pesco}, {and} \bibinfo{person}{Claudio Silva}.} \bibinfo{year}{2013}\natexlab{}.
\newblock \showarticletitle{Practical considerations on marching cubes 33 topological correctness}.
\newblock \bibinfo{journal}{\emph{Computers \& graphics}} (\bibinfo{year}{2013}).
\newblock


\bibitem[De~Ara{\'u}jo et~al\mbox{.}(2015)]%
        {survey}
\bibfield{author}{\bibinfo{person}{Bruno~Rodrigues De~Ara{\'u}jo}, \bibinfo{person}{Daniel~S Lopes}, \bibinfo{person}{Pauline Jepp}, \bibinfo{person}{Joaquim~A Jorge}, {and} \bibinfo{person}{Brian Wyvill}.} \bibinfo{year}{2015}\natexlab{}.
\newblock \showarticletitle{A survey on implicit surface polygonization}.
\newblock \bibinfo{journal}{\emph{Comput. Surveys}} (\bibinfo{year}{2015}).
\newblock


\bibitem[Desbrun et~al\mbox{.}(1999)]%
        {desbrun1999implicit}
\bibfield{author}{\bibinfo{person}{Mathieu Desbrun}, \bibinfo{person}{Mark Meyer}, \bibinfo{person}{Peter Schr{\"o}der}, {and} \bibinfo{person}{Alan~H Barr}.} \bibinfo{year}{1999}\natexlab{}.
\newblock \showarticletitle{Implicit fairing of irregular meshes using diffusion and curvature flow}. In \bibinfo{booktitle}{\emph{Proceedings of the 26th annual conference on Computer graphics and interactive techniques}}. \bibinfo{pages}{317--324}.
\newblock


\bibitem[Dey et~al\mbox{.}(2012)]%
        {dey2012feature}
\bibfield{author}{\bibinfo{person}{Tamal~Krishna Dey}, \bibinfo{person}{Xiaoyin Ge}, \bibinfo{person}{Qichao Que}, \bibinfo{person}{Issam Safa}, \bibinfo{person}{Lei Wang}, {and} \bibinfo{person}{Yusu Wang}.} \bibinfo{year}{2012}\natexlab{}.
\newblock \showarticletitle{Feature-preserving reconstruction of singular surfaces}. In \bibinfo{booktitle}{\emph{Computer Graphics Forum}}, Vol.~\bibinfo{volume}{31}. Wiley Online Library, \bibinfo{pages}{1787--1796}.
\newblock


\bibitem[Doi and Koide(1991)]%
        {marchingtetrahedra}
\bibfield{author}{\bibinfo{person}{Akio Doi} {and} \bibinfo{person}{Akio Koide}.} \bibinfo{year}{1991}\natexlab{}.
\newblock \showarticletitle{An efficient method of triangulating equi-valued surfaces by using tetrahedral cells}.
\newblock \bibinfo{journal}{\emph{IEICE Transactions on Information and Systems}} (\bibinfo{year}{1991}).
\newblock


\bibitem[Erko{\c{c}} et~al\mbox{.}(2023)]%
        {erkocc2023hyperdiffusion}
\bibfield{author}{\bibinfo{person}{Ziya Erko{\c{c}}}, \bibinfo{person}{Fangchang Ma}, \bibinfo{person}{Qi Shan}, \bibinfo{person}{Matthias Nie{\ss}ner}, {and} \bibinfo{person}{Angela Dai}.} \bibinfo{year}{2023}\natexlab{}.
\newblock \showarticletitle{Hyperdiffusion: Generating implicit neural fields with weight-space diffusion}. In \bibinfo{booktitle}{\emph{Proceedings of the IEEE/CVF International Conference on Computer Vision}}. \bibinfo{pages}{14300--14310}.
\newblock


\bibitem[Evrard et~al\mbox{.}(2018)]%
        {evrard2018surface}
\bibfield{author}{\bibinfo{person}{Fabien Evrard}, \bibinfo{person}{Fabian Denner}, {and} \bibinfo{person}{Berend Van~Wachem}.} \bibinfo{year}{2018}\natexlab{}.
\newblock \showarticletitle{Surface reconstruction from discrete indicator functions}.
\newblock \bibinfo{journal}{\emph{IEEE Transactions on Visualization and Computer Graphics}} \bibinfo{volume}{25}, \bibinfo{number}{3} (\bibinfo{year}{2018}), \bibinfo{pages}{1629--1635}.
\newblock


\bibitem[Feng et~al\mbox{.}(2022)]%
        {feng2022fof}
\bibfield{author}{\bibinfo{person}{Qiao Feng}, \bibinfo{person}{Yebin Liu}, \bibinfo{person}{Yu-Kun Lai}, \bibinfo{person}{Jingyu Yang}, {and} \bibinfo{person}{Kun Li}.} \bibinfo{year}{2022}\natexlab{}.
\newblock \showarticletitle{Fof: Learning fourier occupancy field for monocular real-time human reconstruction}.
\newblock \bibinfo{journal}{\emph{Advances in Neural Information Processing Systems}}  \bibinfo{volume}{35} (\bibinfo{year}{2022}), \bibinfo{pages}{7397--7409}.
\newblock


\bibitem[Fuhrmann et~al\mbox{.}(2015)]%
        {fuhrmann2015accurate}
\bibfield{author}{\bibinfo{person}{Simon Fuhrmann}, \bibinfo{person}{Michael Kazhdan}, {and} \bibinfo{person}{Michael Goesele}.} \bibinfo{year}{2015}\natexlab{}.
\newblock \showarticletitle{Accurate isosurface interpolation with hermite data}. In \bibinfo{booktitle}{\emph{2015 International Conference on 3D Vision}}. IEEE, \bibinfo{pages}{256--263}.
\newblock


\bibitem[Genova et~al\mbox{.}(2020)]%
        {genova2020local}
\bibfield{author}{\bibinfo{person}{Kyle Genova}, \bibinfo{person}{Forrester Cole}, \bibinfo{person}{Avneesh Sud}, \bibinfo{person}{Aaron Sarna}, {and} \bibinfo{person}{Thomas Funkhouser}.} \bibinfo{year}{2020}\natexlab{}.
\newblock \showarticletitle{Local deep implicit functions for 3d shape}. In \bibinfo{booktitle}{\emph{Proceedings of the IEEE/CVF conference on computer vision and pattern recognition}}. \bibinfo{pages}{4857--4866}.
\newblock


\bibitem[Gibson(1998)]%
        {gibson1998constrained}
\bibfield{author}{\bibinfo{person}{Sarah~FF Gibson}.} \bibinfo{year}{1998}\natexlab{}.
\newblock \showarticletitle{Constrained elastic surface nets: Generating smooth surfaces from binary segmented data}. In \bibinfo{booktitle}{\emph{International Conference on Medical Image Computing and Computer-Assisted Intervention}}. Springer, \bibinfo{pages}{888--898}.
\newblock


\bibitem[Guillard et~al\mbox{.}(2022)]%
        {guillard2022meshudf}
\bibfield{author}{\bibinfo{person}{Benoit Guillard}, \bibinfo{person}{Federico Stella}, {and} \bibinfo{person}{Pascal Fua}.} \bibinfo{year}{2022}\natexlab{}.
\newblock \showarticletitle{Meshudf: Fast and differentiable meshing of unsigned distance field networks}. In \bibinfo{booktitle}{\emph{European conference on computer vision}}. Springer, \bibinfo{pages}{576--592}.
\newblock


\bibitem[Hertz et~al\mbox{.}(2022)]%
        {hertz2022spaghetti}
\bibfield{author}{\bibinfo{person}{Amir Hertz}, \bibinfo{person}{Or Perel}, \bibinfo{person}{Raja Giryes}, \bibinfo{person}{Olga Sorkine-Hornung}, {and} \bibinfo{person}{Daniel Cohen-Or}.} \bibinfo{year}{2022}\natexlab{}.
\newblock \showarticletitle{Spaghetti: Editing implicit shapes through part aware generation}.
\newblock \bibinfo{journal}{\emph{ACM Transactions on Graphics (TOG)}} \bibinfo{volume}{41}, \bibinfo{number}{4} (\bibinfo{year}{2022}), \bibinfo{pages}{1--20}.
\newblock


\bibitem[Hilton et~al\mbox{.}(1997)]%
        {hilton1997marching}
\bibfield{author}{\bibinfo{person}{Adrian Hilton}, \bibinfo{person}{John Illingworth}, {et~al\mbox{.}}} \bibinfo{year}{1997}\natexlab{}.
\newblock \showarticletitle{Marching triangles: Delaunay implicit surface triangulation}.
\newblock \bibinfo{journal}{\emph{University of Surrey}} (\bibinfo{year}{1997}).
\newblock


\bibitem[Hilton et~al\mbox{.}(1996)]%
        {marchingtriangles}
\bibfield{author}{\bibinfo{person}{Adrian Hilton}, \bibinfo{person}{Andrew~J Stoddart}, \bibinfo{person}{John Illingworth}, {and} \bibinfo{person}{Terry Windeatt}.} \bibinfo{year}{1996}\natexlab{}.
\newblock \showarticletitle{Marching triangles: range image fusion for complex object modelling}. In \bibinfo{booktitle}{\emph{IEEE International Conference on Image Processing}}.
\newblock


\bibitem[Ho et~al\mbox{.}(2005)]%
        {ho2005cubical}
\bibfield{author}{\bibinfo{person}{Chien-Chang Ho}, \bibinfo{person}{Fu-Che Wu}, \bibinfo{person}{Bing-Yu Chen}, \bibinfo{person}{Yung-Yu Chuang}, \bibinfo{person}{Ming Ouhyoung}, {et~al\mbox{.}}} \bibinfo{year}{2005}\natexlab{}.
\newblock \showarticletitle{Cubical marching squares: Adaptive feature preserving surface extraction from volume data}. In \bibinfo{booktitle}{\emph{Computer graphics forum}}, Vol.~\bibinfo{volume}{24}. Amsterdam: North Holland, 1982-, \bibinfo{pages}{537--546}.
\newblock


\bibitem[Hormann et~al\mbox{.}(2002)]%
        {hormann2002hierarchical}
\bibfield{author}{\bibinfo{person}{Kai Hormann}, \bibinfo{person}{Ulf Labsik}, \bibinfo{person}{Martin Meister}, {and} \bibinfo{person}{Gunther Greiner}.} \bibinfo{year}{2002}\natexlab{}.
\newblock \showarticletitle{Hierarchical extraction of iso-surfaces with semi-regular meshes}. In \bibinfo{booktitle}{\emph{Proceedings of the seventh ACM symposium on Solid modeling and applications}}. \bibinfo{pages}{53--58}.
\newblock


\bibitem[Hou et~al\mbox{.}(2023)]%
        {hou2023robust}
\bibfield{author}{\bibinfo{person}{Fei Hou}, \bibinfo{person}{Xuhui Chen}, \bibinfo{person}{Wencheng Wang}, \bibinfo{person}{Hong Qin}, {and} \bibinfo{person}{Ying He}.} \bibinfo{year}{2023}\natexlab{}.
\newblock \showarticletitle{Robust Zero Level-Set Extraction from Unsigned Distance Fields Based on Double Covering}.
\newblock \bibinfo{journal}{\emph{ACM Transactions on Graphics (TOG)}} \bibinfo{volume}{42}, \bibinfo{number}{6} (\bibinfo{year}{2023}), \bibinfo{pages}{1--15}.
\newblock


\bibitem[Ibing et~al\mbox{.}(2021)]%
        {ibing20213d}
\bibfield{author}{\bibinfo{person}{Moritz Ibing}, \bibinfo{person}{Isaak Lim}, {and} \bibinfo{person}{Leif Kobbelt}.} \bibinfo{year}{2021}\natexlab{}.
\newblock \showarticletitle{3d shape generation with grid-based implicit functions}. In \bibinfo{booktitle}{\emph{Proceedings of the IEEE/CVF Conference on Computer Vision and Pattern Recognition}}. \bibinfo{pages}{13559--13568}.
\newblock


\bibitem[Incahuanaco and Paiva(2023)]%
        {incahuanaco2023surface}
\bibfield{author}{\bibinfo{person}{Filomen Incahuanaco} {and} \bibinfo{person}{Afonso Paiva}.} \bibinfo{year}{2023}\natexlab{}.
\newblock \showarticletitle{Surface reconstruction method for particle-based fluids using discrete indicator functions}.
\newblock \bibinfo{journal}{\emph{Computers \& Graphics}}  \bibinfo{volume}{114} (\bibinfo{year}{2023}), \bibinfo{pages}{26--35}.
\newblock


\bibitem[Jia and Kyan(2020)]%
        {jia2020learning}
\bibfield{author}{\bibinfo{person}{Meng Jia} {and} \bibinfo{person}{Matthew Kyan}.} \bibinfo{year}{2020}\natexlab{}.
\newblock \showarticletitle{Learning occupancy function from point clouds for surface reconstruction}.
\newblock \bibinfo{journal}{\emph{arXiv preprint arXiv:2010.11378}} (\bibinfo{year}{2020}).
\newblock


\bibitem[Ju et~al\mbox{.}(2002)]%
        {dualcontouring}
\bibfield{author}{\bibinfo{person}{Tao Ju}, \bibinfo{person}{Frank Losasso}, \bibinfo{person}{Scott Schaefer}, {and} \bibinfo{person}{Joe Warren}.} \bibinfo{year}{2002}\natexlab{}.
\newblock \showarticletitle{Dual contouring of hermite data}. In \bibinfo{booktitle}{\emph{SIGGRAPH}}.
\newblock


\bibitem[Ju and Udeshi(2006)]%
        {ju2006intersection}
\bibfield{author}{\bibinfo{person}{Tao Ju} {and} \bibinfo{person}{Tushar Udeshi}.} \bibinfo{year}{2006}\natexlab{}.
\newblock \showarticletitle{Intersection-free contouring on an octree grid}. In \bibinfo{booktitle}{\emph{Proceedings of Pacific graphics}}, Vol.~\bibinfo{volume}{2006}. Citeseer.
\newblock


\bibitem[Karkanis and Stewart(2001)]%
        {karkanis2001curvature}
\bibfield{author}{\bibinfo{person}{Tasso Karkanis} {and} \bibinfo{person}{A~James Stewart}.} \bibinfo{year}{2001}\natexlab{}.
\newblock \showarticletitle{Curvature-dependent triangulation of implicit surfaces}.
\newblock \bibinfo{journal}{\emph{IEEE Computer Graphics and Applications}} (\bibinfo{year}{2001}).
\newblock


\bibitem[Kobbelt et~al\mbox{.}(2001)]%
        {kobbelt2001feature}
\bibfield{author}{\bibinfo{person}{Leif~P Kobbelt}, \bibinfo{person}{Mario Botsch}, \bibinfo{person}{Ulrich Schwanecke}, {and} \bibinfo{person}{Hans-Peter Seidel}.} \bibinfo{year}{2001}\natexlab{}.
\newblock \showarticletitle{Feature sensitive surface extraction from volume data}. In \bibinfo{booktitle}{\emph{Proceedings of the 28th annual conference on Computer graphics and interactive techniques}}. \bibinfo{pages}{57--66}.
\newblock


\bibitem[Kobbelt et~al\mbox{.}(1999)]%
        {kobbelt1999shrink}
\bibfield{author}{\bibinfo{person}{Leif~P Kobbelt}, \bibinfo{person}{Jens Vorsatz}, \bibinfo{person}{Ulf Labsik}, {and} \bibinfo{person}{Hans-Peter Seidel}.} \bibinfo{year}{1999}\natexlab{}.
\newblock \showarticletitle{A shrink wrapping approach to remeshing polygonal surfaces}. In \bibinfo{booktitle}{\emph{Computer Graphics Forum}}, Vol.~\bibinfo{volume}{18}. Wiley Online Library, \bibinfo{pages}{119--130}.
\newblock


\bibitem[Koch et~al\mbox{.}(2019)]%
        {koch2019abc}
\bibfield{author}{\bibinfo{person}{Sebastian Koch}, \bibinfo{person}{Albert Matveev}, \bibinfo{person}{Zhongshi Jiang}, \bibinfo{person}{Francis Williams}, \bibinfo{person}{Alexey Artemov}, \bibinfo{person}{Evgeny Burnaev}, \bibinfo{person}{Marc Alexa}, \bibinfo{person}{Denis Zorin}, {and} \bibinfo{person}{Daniele Panozzo}.} \bibinfo{year}{2019}\natexlab{}.
\newblock \showarticletitle{Abc: {A} big cad model dataset for geometric deep learning}. In \bibinfo{booktitle}{\emph{CVPR}}.
\newblock


\bibitem[Koo et~al\mbox{.}(2023)]%
        {koo2023salad}
\bibfield{author}{\bibinfo{person}{Juil Koo}, \bibinfo{person}{Seungwoo Yoo}, \bibinfo{person}{Minh~Hieu Nguyen}, {and} \bibinfo{person}{Minhyuk Sung}.} \bibinfo{year}{2023}\natexlab{}.
\newblock \showarticletitle{Salad: Part-level latent diffusion for 3d shape generation and manipulation}. In \bibinfo{booktitle}{\emph{Proceedings of the IEEE/CVF International Conference on Computer Vision}}.
\newblock


\bibitem[Lamb et~al\mbox{.}(2022)]%
        {lamb2022deepmend}
\bibfield{author}{\bibinfo{person}{Nikolas Lamb}, \bibinfo{person}{Sean Banerjee}, {and} \bibinfo{person}{Natasha~Kholgade Banerjee}.} \bibinfo{year}{2022}\natexlab{}.
\newblock \showarticletitle{Deepmend: Learning occupancy functions to represent shape for repair}. In \bibinfo{booktitle}{\emph{European Conference on Computer Vision}}. Springer, \bibinfo{pages}{433--450}.
\newblock


\bibitem[Lempitsky(2010)]%
        {lempitsky2010surface}
\bibfield{author}{\bibinfo{person}{Victor Lempitsky}.} \bibinfo{year}{2010}\natexlab{}.
\newblock \showarticletitle{Surface extraction from binary volumes with higher-order smoothness}. In \bibinfo{booktitle}{\emph{2010 IEEE Computer Society Conference on Computer Vision and Pattern Recognition}}.
\newblock


\bibitem[Lewiner et~al\mbox{.}(2003)]%
        {lewiner2003efficient}
\bibfield{author}{\bibinfo{person}{Thomas Lewiner}, \bibinfo{person}{H{\'e}lio Lopes}, \bibinfo{person}{Ant{\^o}nio~Wilson Vieira}, {and} \bibinfo{person}{Geovan Tavares}.} \bibinfo{year}{2003}\natexlab{}.
\newblock \showarticletitle{Efficient implementation of marching cubes' cases with topological guarantees}.
\newblock \bibinfo{journal}{\emph{Journal of graphics tools}} (\bibinfo{year}{2003}).
\newblock


\bibitem[Lin et~al\mbox{.}(2022)]%
        {lin2022neuform}
\bibfield{author}{\bibinfo{person}{Connor Lin}, \bibinfo{person}{Niloy Mitra}, \bibinfo{person}{Gordon Wetzstein}, \bibinfo{person}{Leonidas~J Guibas}, {and} \bibinfo{person}{Paul Guerrero}.} \bibinfo{year}{2022}\natexlab{}.
\newblock \showarticletitle{Neuform: Adaptive overfitting for neural shape editing}.
\newblock \bibinfo{journal}{\emph{Advances in Neural Information Processing Systems}}  \bibinfo{volume}{35} (\bibinfo{year}{2022}), \bibinfo{pages}{15217--15229}.
\newblock


\bibitem[Lionar et~al\mbox{.}(2021)]%
        {lionar2021dynamic}
\bibfield{author}{\bibinfo{person}{Stefan Lionar}, \bibinfo{person}{Daniil Emtsev}, \bibinfo{person}{Dusan Svilarkovic}, {and} \bibinfo{person}{Songyou Peng}.} \bibinfo{year}{2021}\natexlab{}.
\newblock \showarticletitle{Dynamic plane convolutional occupancy networks}. In \bibinfo{booktitle}{\emph{Proceedings of the IEEE/CVF Winter Conference on Applications of Computer Vision}}. \bibinfo{pages}{1829--1838}.
\newblock


\bibitem[Liu et~al\mbox{.}(2023)]%
        {liu2023exim}
\bibfield{author}{\bibinfo{person}{Zhengzhe Liu}, \bibinfo{person}{Jingyu Hu}, \bibinfo{person}{Ka-Hei Hui}, \bibinfo{person}{Xiaojuan Qi}, \bibinfo{person}{Daniel Cohen-Or}, {and} \bibinfo{person}{Chi-Wing Fu}.} \bibinfo{year}{2023}\natexlab{}.
\newblock \showarticletitle{EXIM: A Hybrid Explicit-Implicit Representation for Text-Guided 3D Shape Generation}.
\newblock \bibinfo{journal}{\emph{ACM Transactions on Graphics (TOG)}} \bibinfo{volume}{42}, \bibinfo{number}{6} (\bibinfo{year}{2023}), \bibinfo{pages}{1--12}.
\newblock


\bibitem[Lopes and Brodlie(2003)]%
        {lopes2003improving}
\bibfield{author}{\bibinfo{person}{Adriano Lopes} {and} \bibinfo{person}{Ken Brodlie}.} \bibinfo{year}{2003}\natexlab{}.
\newblock \showarticletitle{Improving the robustness and accuracy of the marching cubes algorithm for isosurfacing}.
\newblock \bibinfo{journal}{\emph{ACM Transactions on Graphics}} (\bibinfo{year}{2003}).
\newblock


\bibitem[Lorensen and Cline(1987)]%
        {marchingcubes}
\bibfield{author}{\bibinfo{person}{William~E. Lorensen} {and} \bibinfo{person}{Harvey~E. Cline}.} \bibinfo{year}{1987}\natexlab{}.
\newblock \showarticletitle{Marching cubes: {A} high resolution {3D} surface construction algorithm}.
\newblock \bibinfo{journal}{\emph{SIGGRAPH}} (\bibinfo{year}{1987}).
\newblock


\bibitem[Manson and Schaefer(2010)]%
        {manson2010isosurfaces}
\bibfield{author}{\bibinfo{person}{Josiah Manson} {and} \bibinfo{person}{Scott Schaefer}.} \bibinfo{year}{2010}\natexlab{}.
\newblock \showarticletitle{Isosurfaces over simplicial partitions of multiresolution grids}. In \bibinfo{booktitle}{\emph{Computer Graphics Forum}}, Vol.~\bibinfo{volume}{29}. Wiley Online Library, \bibinfo{pages}{377--385}.
\newblock


\bibitem[Manson et~al\mbox{.}(2011)]%
        {manson2011contouring}
\bibfield{author}{\bibinfo{person}{Josiah Manson}, \bibinfo{person}{Jason Smith}, {and} \bibinfo{person}{Scott Schaefer}.} \bibinfo{year}{2011}\natexlab{}.
\newblock \showarticletitle{Contouring discrete indicator functions}. In \bibinfo{booktitle}{\emph{Computer Graphics Forum}}, Vol.~\bibinfo{volume}{30}. Wiley Online Library, \bibinfo{pages}{385--393}.
\newblock


\bibitem[Maruani et~al\mbox{.}(2023)]%
        {voromesh}
\bibfield{author}{\bibinfo{person}{Nissim Maruani}, \bibinfo{person}{Roman Klokov}, \bibinfo{person}{Maks Ovsjanikov}, \bibinfo{person}{Pierre Alliez}, {and} \bibinfo{person}{Mathieu Desbrun}.} \bibinfo{year}{2023}\natexlab{}.
\newblock \showarticletitle{{VoroMesh}: {Learning Watertight Surface Meshes with Voronoi Diagrams}}. In \bibinfo{booktitle}{\emph{ICCV}}.
\newblock


\bibitem[Maruani et~al\mbox{.}(2024)]%
        {maruani2024ponq}
\bibfield{author}{\bibinfo{person}{Nissim Maruani}, \bibinfo{person}{Maks Ovsjanikov}, \bibinfo{person}{Pierre Alliez}, {and} \bibinfo{person}{Mathieu Desbrun}.} \bibinfo{year}{2024}\natexlab{}.
\newblock \showarticletitle{PoNQ: a Neural QEM-based Mesh Representation}.
\newblock \bibinfo{journal}{\emph{arXiv preprint arXiv:2403.12870}} (\bibinfo{year}{2024}).
\newblock


\bibitem[McCormick and Fisher(2002)]%
        {mccormick2002edge}
\bibfield{author}{\bibinfo{person}{Neil~H McCormick} {and} \bibinfo{person}{Robert~B Fisher}.} \bibinfo{year}{2002}\natexlab{}.
\newblock \showarticletitle{Edge-constrained marching triangles}. In \bibinfo{booktitle}{\emph{Proceedings. First International Symposium on 3D Data Processing Visualization and Transmission}}.
\newblock


\bibitem[Mescheder et~al\mbox{.}(2019)]%
        {occupancenetworks}
\bibfield{author}{\bibinfo{person}{Lars Mescheder}, \bibinfo{person}{Michael Oechsle}, \bibinfo{person}{Michael Niemeyer}, \bibinfo{person}{Sebastian Nowozin}, {and} \bibinfo{person}{Andreas Geiger}.} \bibinfo{year}{2019}\natexlab{}.
\newblock \showarticletitle{Occupancy networks: {Learning} 3d reconstruction in function space}. In \bibinfo{booktitle}{\emph{CVPR}}.
\newblock


\bibitem[M{\"u}ller and Stark(1993)]%
        {muller1993adaptive}
\bibfield{author}{\bibinfo{person}{Heinrich M{\"u}ller} {and} \bibinfo{person}{Michael Stark}.} \bibinfo{year}{1993}\natexlab{}.
\newblock \showarticletitle{Adaptive generation of surfaces in volume data}.
\newblock \bibinfo{journal}{\emph{The Visual Computer}}  \bibinfo{volume}{9} (\bibinfo{year}{1993}), \bibinfo{pages}{182--199}.
\newblock


\bibitem[Muller and Wehle(1997)]%
        {muller1997visualization}
\bibfield{author}{\bibinfo{person}{Heinrich Muller} {and} \bibinfo{person}{Michael Wehle}.} \bibinfo{year}{1997}\natexlab{}.
\newblock \showarticletitle{Visualization of implicit surfaces using adaptive tetrahedrizations}. In \bibinfo{booktitle}{\emph{Scientific Visualization Conference}}. IEEE, \bibinfo{pages}{243--243}.
\newblock


\bibitem[Myles et~al\mbox{.}(2014)]%
        {myles2014robust}
\bibfield{author}{\bibinfo{person}{Ashish Myles}, \bibinfo{person}{Nico Pietroni}, {and} \bibinfo{person}{Denis Zorin}.} \bibinfo{year}{2014}\natexlab{}.
\newblock \showarticletitle{Robust field-aligned global parametrization.}
\newblock \bibinfo{journal}{\emph{ACM Transactions on Graphics}} (\bibinfo{year}{2014}).
\newblock


\bibitem[Natarajan(1994)]%
        {natarajan1994generating}
\bibfield{author}{\bibinfo{person}{Balas~K Natarajan}.} \bibinfo{year}{1994}\natexlab{}.
\newblock \showarticletitle{On generating topologically consistent isosurfaces from uniform samples}.
\newblock \bibinfo{journal}{\emph{The Visual Computer}}  \bibinfo{volume}{11} (\bibinfo{year}{1994}), \bibinfo{pages}{52--62}.
\newblock


\bibitem[Nielson(2003)]%
        {marchingcubes2003}
\bibfield{author}{\bibinfo{person}{Gregory~M. Nielson}.} \bibinfo{year}{2003}\natexlab{}.
\newblock \showarticletitle{On marching cubes}.
\newblock \bibinfo{journal}{\emph{IEEE Transactions on Visualization and Computer Graphics}} (\bibinfo{year}{2003}).
\newblock


\bibitem[Nielson(2004)]%
        {dualmarchingcubes}
\bibfield{author}{\bibinfo{person}{Gregory~M Nielson}.} \bibinfo{year}{2004}\natexlab{}.
\newblock \showarticletitle{Dual marching cubes}. In \bibinfo{booktitle}{\emph{IEEE Visualization}}.
\newblock


\bibitem[Nielson(2008)]%
        {nielson2008dual}
\bibfield{author}{\bibinfo{person}{Gregory~M Nielson}.} \bibinfo{year}{2008}\natexlab{}.
\newblock \showarticletitle{Dual marching tetrahedra: Contouring in the tetrahedronal environment}. In \bibinfo{booktitle}{\emph{International Symposium on Visual Computing}}. Springer, \bibinfo{pages}{183--194}.
\newblock


\bibitem[Nielson and Hamann(1991)]%
        {nielson1991asymptotic}
\bibfield{author}{\bibinfo{person}{Gregory~M Nielson} {and} \bibinfo{person}{Bernd Hamann}.} \bibinfo{year}{1991}\natexlab{}.
\newblock \showarticletitle{The asymptotic decider: resolving the ambiguity in marching cubes.}. In \bibinfo{booktitle}{\emph{IEEE visualization}}.
\newblock


\bibitem[Ohtake et~al\mbox{.}(2001)]%
        {ohtake2001mesh}
\bibfield{author}{\bibinfo{person}{Yutaka Ohtake}, \bibinfo{person}{Alexander Belyaev}, {and} \bibinfo{person}{Ilia Bogaevski}.} \bibinfo{year}{2001}\natexlab{}.
\newblock \showarticletitle{Mesh regularization and adaptive smoothing}.
\newblock \bibinfo{journal}{\emph{Computer-Aided Design}} \bibinfo{volume}{33}, \bibinfo{number}{11} (\bibinfo{year}{2001}), \bibinfo{pages}{789--800}.
\newblock


\bibitem[Park et~al\mbox{.}(2019)]%
        {deepsdf}
\bibfield{author}{\bibinfo{person}{Jeong~Joon Park}, \bibinfo{person}{Peter Florence}, \bibinfo{person}{Julian Straub}, \bibinfo{person}{Richard Newcombe}, {and} \bibinfo{person}{Steven Lovegrove}.} \bibinfo{year}{2019}\natexlab{}.
\newblock \showarticletitle{Deepsdf: {Learning} continuous signed distance functions for shape representation}. In \bibinfo{booktitle}{\emph{CVPR}}.
\newblock


\bibitem[Peng et~al\mbox{.}(2020)]%
        {peng2020convolutional}
\bibfield{author}{\bibinfo{person}{Songyou Peng}, \bibinfo{person}{Michael Niemeyer}, \bibinfo{person}{Lars Mescheder}, \bibinfo{person}{Marc Pollefeys}, {and} \bibinfo{person}{Andreas Geiger}.} \bibinfo{year}{2020}\natexlab{}.
\newblock \showarticletitle{Convolutional occupancy networks}. In \bibinfo{booktitle}{\emph{Computer Vision--ECCV 2020: 16th European Conference, Glasgow, UK, August 23--28, 2020, Proceedings, Part III 16}}. Springer, \bibinfo{pages}{523--540}.
\newblock


\bibitem[Poursaeed et~al\mbox{.}(2020)]%
        {poursaeed2020coupling}
\bibfield{author}{\bibinfo{person}{Omid Poursaeed}, \bibinfo{person}{Matthew Fisher}, \bibinfo{person}{Noam Aigerman}, {and} \bibinfo{person}{Vladimir~G Kim}.} \bibinfo{year}{2020}\natexlab{}.
\newblock \showarticletitle{Coupling explicit and implicit surface representations for generative 3d modeling}. In \bibinfo{booktitle}{\emph{Computer Vision--ECCV 2020: 16th European Conference, Glasgow, UK, August 23--28, 2020, Proceedings, Part X 16}}. Springer, \bibinfo{pages}{667--683}.
\newblock


\bibitem[Remelli et~al\mbox{.}(2020)]%
        {meshsdf}
\bibfield{author}{\bibinfo{person}{Edoardo Remelli}, \bibinfo{person}{Artem Lukoianov}, \bibinfo{person}{Stephan Richter}, \bibinfo{person}{Benoit Guillard}, \bibinfo{person}{Timur Bagautdinov}, \bibinfo{person}{Pierre Baque}, {and} \bibinfo{person}{Pascal Fua}.} \bibinfo{year}{2020}\natexlab{}.
\newblock \showarticletitle{Meshsdf: {Differentiable} iso-surface extraction}.
\newblock \bibinfo{journal}{\emph{NeurIPS}} (\bibinfo{year}{2020}).
\newblock


\bibitem[Salman et~al\mbox{.}(2010)]%
        {salman2010feature}
\bibfield{author}{\bibinfo{person}{Nader Salman}, \bibinfo{person}{Mariette Yvinec}, {and} \bibinfo{person}{Quentin M{\'e}rigot}.} \bibinfo{year}{2010}\natexlab{}.
\newblock \showarticletitle{Feature preserving mesh generation from 3d point clouds}. In \bibinfo{booktitle}{\emph{Computer graphics forum}}, Vol.~\bibinfo{volume}{29}. Wiley Online Library, \bibinfo{pages}{1623--1632}.
\newblock


\bibitem[Schaefer et~al\mbox{.}(2007)]%
        {manifolddualcontouring}
\bibfield{author}{\bibinfo{person}{Scott Schaefer}, \bibinfo{person}{Tao Ju}, {and} \bibinfo{person}{Joe Warren}.} \bibinfo{year}{2007}\natexlab{}.
\newblock \showarticletitle{Manifold dual contouring}.
\newblock \bibinfo{journal}{\emph{IEEE Transactions on Visualization and Computer Graphics}} (\bibinfo{year}{2007}).
\newblock


\bibitem[Sell{\'a}n et~al\mbox{.}(2023)]%
        {sellan2023reach}
\bibfield{author}{\bibinfo{person}{Silvia Sell{\'a}n}, \bibinfo{person}{Christopher Batty}, {and} \bibinfo{person}{Oded Stein}.} \bibinfo{year}{2023}\natexlab{}.
\newblock \showarticletitle{Reach For the Spheres: Tangency-aware surface reconstruction of SDFs}. In \bibinfo{booktitle}{\emph{SIGGRAPH Asia 2023 Conference Papers}}. \bibinfo{pages}{1--11}.
\newblock


\bibitem[Shekhar et~al\mbox{.}(1996)]%
        {shekhar1996octree}
\bibfield{author}{\bibinfo{person}{Raj Shekhar}, \bibinfo{person}{Elias Fayyad}, \bibinfo{person}{Roni Yagel}, {and} \bibinfo{person}{J~Fredrick Cornhill}.} \bibinfo{year}{1996}\natexlab{}.
\newblock \showarticletitle{Octree-based decimation of marching cubes surfaces}. In \bibinfo{booktitle}{\emph{Proceedings of Seventh Annual IEEE Visualization'96}}. IEEE, \bibinfo{pages}{335--342}.
\newblock


\bibitem[Shen et~al\mbox{.}(2021)]%
        {dmtet}
\bibfield{author}{\bibinfo{person}{Tianchang Shen}, \bibinfo{person}{Jun Gao}, \bibinfo{person}{Kangxue Yin}, \bibinfo{person}{Ming-Yu Liu}, {and} \bibinfo{person}{Sanja Fidler}.} \bibinfo{year}{2021}\natexlab{}.
\newblock \showarticletitle{Deep marching tetrahedra: a hybrid representation for high-resolution 3d shape synthesis}.
\newblock \bibinfo{journal}{\emph{NeurIPS}} (\bibinfo{year}{2021}).
\newblock


\bibitem[Shen et~al\mbox{.}(2023)]%
        {flexicubes}
\bibfield{author}{\bibinfo{person}{Tianchang Shen}, \bibinfo{person}{Jacob Munkberg}, \bibinfo{person}{Jon Hasselgren}, \bibinfo{person}{Kangxue Yin}, \bibinfo{person}{Zian Wang}, \bibinfo{person}{Wenzheng Chen}, \bibinfo{person}{Zan Gojcic}, \bibinfo{person}{Sanja Fidler}, \bibinfo{person}{Nicholas Sharp}, {and} \bibinfo{person}{Jun Gao}.} \bibinfo{year}{2023}\natexlab{}.
\newblock \showarticletitle{Flexible isosurface extraction for gradient-based mesh optimization}.
\newblock \bibinfo{journal}{\emph{ACM Transactions on Graphics}} (\bibinfo{year}{2023}).
\newblock


\bibitem[Shue et~al\mbox{.}(2023)]%
        {shue20233d}
\bibfield{author}{\bibinfo{person}{J~Ryan Shue}, \bibinfo{person}{Eric~Ryan Chan}, \bibinfo{person}{Ryan Po}, \bibinfo{person}{Zachary Ankner}, \bibinfo{person}{Jiajun Wu}, {and} \bibinfo{person}{Gordon Wetzstein}.} \bibinfo{year}{2023}\natexlab{}.
\newblock \showarticletitle{3d neural field generation using triplane diffusion}. In \bibinfo{booktitle}{\emph{Proceedings of the IEEE/CVF Conference on Computer Vision and Pattern Recognition}}. \bibinfo{pages}{20875--20886}.
\newblock


\bibitem[Stander and Hart(1995)]%
        {stander1995interactive}
\bibfield{author}{\bibinfo{person}{Barton~T Stander} {and} \bibinfo{person}{John~C Hart}.} \bibinfo{year}{1995}\natexlab{}.
\newblock \showarticletitle{Interactive re-polygonization of blobby implicit curves}. In \bibinfo{booktitle}{\emph{Proceedings of the Western Computer Graphics Symposium}}.
\newblock


\bibitem[Tang et~al\mbox{.}(2021)]%
        {tang2021sa}
\bibfield{author}{\bibinfo{person}{Jiapeng Tang}, \bibinfo{person}{Jiabao Lei}, \bibinfo{person}{Dan Xu}, \bibinfo{person}{Feiying Ma}, \bibinfo{person}{Kui Jia}, {and} \bibinfo{person}{Lei Zhang}.} \bibinfo{year}{2021}\natexlab{}.
\newblock \showarticletitle{Sa-convonet: Sign-agnostic optimization of convolutional occupancy networks}. In \bibinfo{booktitle}{\emph{Proceedings of the IEEE/CVF International Conference on Computer Vision}}. \bibinfo{pages}{6504--6513}.
\newblock


\bibitem[Taubin(1995)]%
        {taubin1995signal}
\bibfield{author}{\bibinfo{person}{Gabriel Taubin}.} \bibinfo{year}{1995}\natexlab{}.
\newblock \showarticletitle{A signal processing approach to fair surface design}. In \bibinfo{booktitle}{\emph{Proceedings of the 22nd annual conference on Computer graphics and interactive techniques}}. \bibinfo{pages}{351--358}.
\newblock


\bibitem[Tian et~al\mbox{.}(2024)]%
        {tian2024occ3d}
\bibfield{author}{\bibinfo{person}{Xiaoyu Tian}, \bibinfo{person}{Tao Jiang}, \bibinfo{person}{Longfei Yun}, \bibinfo{person}{Yucheng Mao}, \bibinfo{person}{Huitong Yang}, \bibinfo{person}{Yue Wang}, \bibinfo{person}{Yilun Wang}, {and} \bibinfo{person}{Hang Zhao}.} \bibinfo{year}{2024}\natexlab{}.
\newblock \showarticletitle{Occ3d: A large-scale 3d occupancy prediction benchmark for autonomous driving}.
\newblock \bibinfo{journal}{\emph{Advances in Neural Information Processing Systems}}  \bibinfo{volume}{36} (\bibinfo{year}{2024}).
\newblock


\bibitem[Van~Overveld and Wyvill(2004)]%
        {van2004shrinkwrap}
\bibfield{author}{\bibinfo{person}{Kees Van~Overveld} {and} \bibinfo{person}{Brian Wyvill}.} \bibinfo{year}{2004}\natexlab{}.
\newblock \showarticletitle{Shrinkwrap: {An} efficient adaptive algorithm for triangulating an iso-surface}.
\newblock \bibinfo{journal}{\emph{The Visual Computer}} (\bibinfo{year}{2004}).
\newblock


\bibitem[Wang(2009)]%
        {wang2009intersection}
\bibfield{author}{\bibinfo{person}{Charlie~CL Wang}.} \bibinfo{year}{2009}\natexlab{}.
\newblock \bibinfo{booktitle}{\emph{Intersection-free dual contouring on uniform grids: an approach based on convex/concave analysis}}.
\newblock \bibinfo{type}{{T}echnical {R}eport}. \bibinfo{institution}{Technical Report, 2009, http://www2. mae. cuhk. edu. hk/\~{} cwang/pubs~…}.
\newblock


\bibitem[Wang et~al\mbox{.}(2013)]%
        {wang2013feature}
\bibfield{author}{\bibinfo{person}{Jun Wang}, \bibinfo{person}{Z Yu}, \bibinfo{person}{W Zhu}, {and} \bibinfo{person}{J Cao}.} \bibinfo{year}{2013}\natexlab{}.
\newblock \showarticletitle{Feature-preserving surface reconstruction from unoriented, noisy point data}. In \bibinfo{booktitle}{\emph{Computer Graphics Forum}}, Vol.~\bibinfo{volume}{32}. Wiley Online Library, \bibinfo{pages}{164--176}.
\newblock


\bibitem[Wang et~al\mbox{.}(2024)]%
        {wang2024sparse}
\bibfield{author}{\bibinfo{person}{Tao Wang}, \bibinfo{person}{Jing Wu}, \bibinfo{person}{Ze Ji}, {and} \bibinfo{person}{Yu-Kun Lai}.} \bibinfo{year}{2024}\natexlab{}.
\newblock \showarticletitle{Sparse Convolutional Networks for Surface Reconstruction from Noisy Point Clouds}. In \bibinfo{booktitle}{\emph{Proceedings of the IEEE/CVF Winter Conference on Applications of Computer Vision}}. \bibinfo{pages}{3212--3221}.
\newblock


\bibitem[Wenger(2013)]%
        {wenger2013isosurfaces}
\bibfield{author}{\bibinfo{person}{Rephael Wenger}.} \bibinfo{year}{2013}\natexlab{}.
\newblock \bibinfo{booktitle}{\emph{Isosurfaces: geometry, topology, and algorithms}}.
\newblock \bibinfo{publisher}{CRC Press}.
\newblock


\bibitem[Westermann et~al\mbox{.}(1999)]%
        {westermann1999real}
\bibfield{author}{\bibinfo{person}{R{\"u}diger Westermann}, \bibinfo{person}{Leif Kobbelt}, {and} \bibinfo{person}{Thomas Ertl}.} \bibinfo{year}{1999}\natexlab{}.
\newblock \showarticletitle{Real-time exploration of regular volume data by adaptive reconstruction of isosurfaces}.
\newblock \bibinfo{journal}{\emph{The Visual Computer}} \bibinfo{volume}{15}, \bibinfo{number}{2} (\bibinfo{year}{1999}), \bibinfo{pages}{100--111}.
\newblock


\bibitem[Zhang et~al\mbox{.}(2022)]%
        {zhang20223dilg}
\bibfield{author}{\bibinfo{person}{Biao Zhang}, \bibinfo{person}{Matthias Nie{\ss}ner}, {and} \bibinfo{person}{Peter Wonka}.} \bibinfo{year}{2022}\natexlab{}.
\newblock \showarticletitle{3dilg: Irregular latent grids for 3d generative modeling}.
\newblock \bibinfo{journal}{\emph{Advances in Neural Information Processing Systems}}  \bibinfo{volume}{35} (\bibinfo{year}{2022}), \bibinfo{pages}{21871--21885}.
\newblock


\bibitem[Zhang et~al\mbox{.}(2023b)]%
        {zhang20233dshape2vecset}
\bibfield{author}{\bibinfo{person}{Biao Zhang}, \bibinfo{person}{Jiapeng Tang}, \bibinfo{person}{Matthias Niessner}, {and} \bibinfo{person}{Peter Wonka}.} \bibinfo{year}{2023}\natexlab{b}.
\newblock \showarticletitle{3dshape2vecset: A 3d shape representation for neural fields and generative diffusion models}.
\newblock \bibinfo{journal}{\emph{ACM Transactions on Graphics (TOG)}} \bibinfo{volume}{42}, \bibinfo{number}{4} (\bibinfo{year}{2023}), \bibinfo{pages}{1--16}.
\newblock


\bibitem[Zhang et~al\mbox{.}(2023a)]%
        {zhang2023surface}
\bibfield{author}{\bibinfo{person}{Congyi Zhang}, \bibinfo{person}{Guying Lin}, \bibinfo{person}{Lei Yang}, \bibinfo{person}{Xin Li}, \bibinfo{person}{Taku Komura}, \bibinfo{person}{Scott Schaefer}, \bibinfo{person}{John Keyser}, {and} \bibinfo{person}{Wenping Wang}.} \bibinfo{year}{2023}\natexlab{a}.
\newblock \showarticletitle{Surface Extraction from Neural Unsigned Distance Fields}. In \bibinfo{booktitle}{\emph{Proceedings of the IEEE/CVF International Conference on Computer Vision}}. \bibinfo{pages}{22531--22540}.
\newblock


\bibitem[Zhao et~al\mbox{.}(2024)]%
        {zhao2024michelangelo}
\bibfield{author}{\bibinfo{person}{Zibo Zhao}, \bibinfo{person}{Wen Liu}, \bibinfo{person}{Xin Chen}, \bibinfo{person}{Xianfang Zeng}, \bibinfo{person}{Rui Wang}, \bibinfo{person}{Pei Cheng}, \bibinfo{person}{Bin Fu}, \bibinfo{person}{Tao Chen}, \bibinfo{person}{Gang Yu}, {and} \bibinfo{person}{Shenghua Gao}.} \bibinfo{year}{2024}\natexlab{}.
\newblock \showarticletitle{Michelangelo: Conditional 3d shape generation based on shape-image-text aligned latent representation}.
\newblock \bibinfo{journal}{\emph{Advances in Neural Information Processing Systems}}  \bibinfo{volume}{36} (\bibinfo{year}{2024}).
\newblock


\bibitem[Zhou et~al\mbox{.}(2022)]%
        {zhou2022learning}
\bibfield{author}{\bibinfo{person}{Junsheng Zhou}, \bibinfo{person}{Baorui Ma}, \bibinfo{person}{Yu-Shen Liu}, \bibinfo{person}{Yi Fang}, {and} \bibinfo{person}{Zhizhong Han}.} \bibinfo{year}{2022}\natexlab{}.
\newblock \showarticletitle{Learning consistency-aware unsigned distance functions progressively from raw point clouds}.
\newblock \bibinfo{journal}{\emph{Advances in neural information processing systems}}  \bibinfo{volume}{35} (\bibinfo{year}{2022}), \bibinfo{pages}{16481--16494}.
\newblock


\bibitem[Zhou et~al\mbox{.}(1997)]%
        {zhou1997multiresolution}
\bibfield{author}{\bibinfo{person}{Yong Zhou}, \bibinfo{person}{Baoquan Chen}, {and} \bibinfo{person}{Arie Kaufman}.} \bibinfo{year}{1997}\natexlab{}.
\newblock \showarticletitle{Multiresolution tetrahedral framework for visualizing regular volume data}. In \bibinfo{booktitle}{\emph{Proceedings. Visualization'97}}. IEEE, \bibinfo{pages}{135--142}.
\newblock


\end{thebibliography}

\ifpaper
    \clearpage
    \newpage

    \setcounter{section}{0}
    \setcounter{table}{0}
    \setcounter{figure}{0}

    \renewcommand{\thesection}{A.\arabic{section}}
    \renewcommand{\thetable}{A\arabic{table}}
    \renewcommand{\thefigure}{A\arabic{figure}}

    {\Huge\sffamily Appendix}
    \vspace{0.3cm}

    We detail our method, ODC, in Section~\ref{supp_sec:detail_odc}, and outline the experimental settings in Section~\ref{supp_sec:exp_setting}.
\textcolor{color_4}{Mesh simplification of MISE~\cite{occupancenetworks}, results across various resolutions, and} additional results with IM-NET~\cite{imnet}, Michelangelo~\cite{zhao2024michelangelo} \textcolor{color_4}{and DeepSDF~\cite{deepsdf}} are discussed in Section~\ref{supp_sec:add_exp}. 
Section~\ref{supp_sec:more_results} offers more detailed qualitative and quantitative results of the experiments in the main paper.
\ifpaper
\else We use the consistent notation from the main paper.
\fi

\section{Details on Occupancy-Based Dual Contouring (ODC)}
\label{supp_sec:detail_odc}

\subsection{Configurations for ODC}

Following the uniform grid setup of Manifold Dual Contouring~\cite{manifolddualcontouring}, we adopted the same configuration as Dual Marching Cubes~\cite{dualmarchingcubes}. Refer to Figure~\ref{fig:22_config}, which illustrates 22 configurations out of a total of 256 possible configurations, accounting for rotation and mirroring. It demonstrates how the surface-crossing grid edges in a grid cell $c$ are partitioned into $\mathcal{E}^\mathcal{S}_{c,1}$, ... $\mathcal{E}^\mathcal{S}_{c,4}$, each colored differently. Since the polygons of this primal face form a cycle and 1D points and 2D points are alternative, each 1D point is adjacent to two 2D points, and each 2D point is also adjacent to two 1D points. Consequently, the normal of a local flat plane of a 1D point can be calculated along with the two adjacent 2D points, and 2D point search also can be done with the two adjacent 1D points. 

Although it generates manifold meshes for almost every cases, as mentioned on the Dual Marching Cubes paper, it may produce non-manifold edges when the ambiguity of C12 and C18 coincide. To address the this problem, we followed the solution described in~\cite{wenger2013isosurfaces}, and the output of Occupancy-based Dual Contouring (ODC) is guaranteed to be manifold.

\subsection{Proof of Lemma 1}
\label{supp_sec:proof_lemma_1}
\begin{lemma}
    \label{lem:2d_search_supp}
    Let $l_1$ and $l_2$ be lines that pass distinct 1D points $\mathbf{p}_{e_1}$ and $\mathbf{p}_{e_2}$, respectively. $l_1$ and $l_2$ are identical or meet at a unique point $\mathbf{u}$ which is neither $\mathbf{p}_{e_1}$ nor $\mathbf{p}_{e_2}$. Then, \textcolor{color_4}{the 2D point $\mathbf{p}_f$ is chosen as either} the middle point $\mathbf{m}$ on the former case or the intersection $\mathbf{u}$ on the latter case.
\end{lemma}

\begin{figure}[t]

  \centering
  \includegraphics[width=\linewidth]{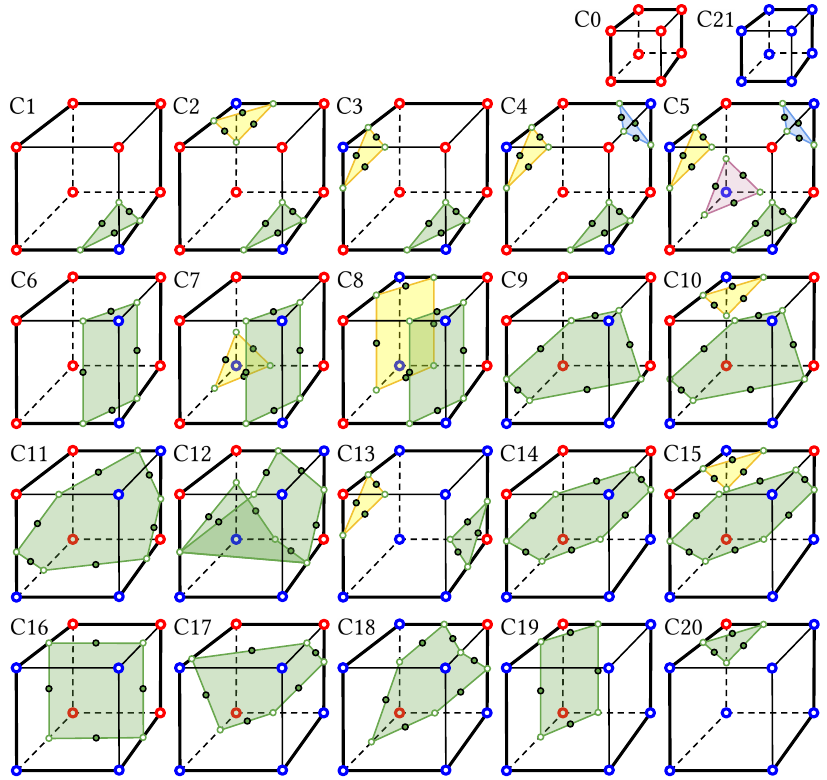}
  \vspace{-0.4cm}
  \caption{
  Configurations of primal faces from MC'03~\cite{marchingcubes2003}. Red and blue dots indicate grid vertices in $\mathcal{P}^\text{in}$ and $\mathcal{P}^\text{out}$, respectively. Dots outlined in green and black represent the 1D and 2D points, respectively. The polygons, distinguished by different colors, represent primal faces. Grid edges containing 1D points associated with the green, yellow, blue, and pink primal faces constitute $\mathcal{E}^\mathcal{S}_{c,1}$, ..., $\mathcal{E}^\mathcal{S}_{c,4}$, respectively. Note that there are two adjacent 2D points in the primal face for each 1D point, and there are two adjacent 1D points in the primal face for each 2D point. In Dual Marching Cubes~\cite{dualmarchingcubes} and Manifold Dual Contouring~\cite{manifolddualcontouring}, a 3D point is defined for each primal face, and the final mesh is created by connecting these 3D points.
  }
  \label{fig:22_config}
  \vspace{-0.3cm}
\end{figure}

\begin{proof}
When $l_1$ and $l_2$ are identical, the isocurve between $\mathbf{p}_{e_1}$ and $\mathbf{p}_{e_2}$ is a straight line segment connecting $\mathbf{p}_{e_1}$ and $\mathbf{p}_{e_2}$, and step (1) finds $\mathbf{q}$ as $\mathbf{m}$. Therefore, $\mathbf{p}_f$ is set to be $\mathbf{m}$.

Consider the case that $l_1$ and $l_2$ meet at the unique point $\mathbf{u}$. From step (1), we assume that $\mathbf{q}$ is on $l_1$ without loss of generality. ($\mathbf{q}$ and $\mathbf{m}$ are distinct.) If $\mathbf{u}$ is closer to $l$ than $\mathbf{q}$, then since $\mathbf{q}$ is outside the triangle $\mathbf{p}_{e_1}\mathbf{u}\mathbf{p}_{e_2}$, $\mathbf{r}$ must have detected a point $\mathbf{q}'$ on $l_1$ or $l_2$ before meeting $\mathbf{u}$, which is contradiction to the step (1). If $\mathbf{q}$ is $\mathbf{u}$, then trivially $\mathbf{q}_2$ is set to be $\mathbf{q}$ as it cannot be found on step (2), and $\mathbf{p}_f$ is set to be $\mathbf{u}$ on step (3). Else, if $\mathbf{q}$ is not $\mathbf{u}$, we have two distinct points $\mathbf{q}_1$ and $\mathbf{q}_2$ on step (2). Since both points are located in the same distance from $l$ with $\mathbf{q}$. Therefore, $\overleftrightarrow{\mathbf{p}_{e_2}\mathbf{q}_2}$ and $\overleftrightarrow{\mathbf{p}_{e_1}\mathbf{q}_1}$ should intersect at further location than $\mathbf{q}$, and which is satisfied by the directions specified on step (2) to detect $\mathbf{q}_1$ and $\mathbf{q}_2$. Hence, the line $\overleftrightarrow{\mathbf{p}_{e_1}\mathbf{q}_1}$ denotes $l_1$ and the line $\overleftrightarrow{\mathbf{p}_{e_2}\mathbf{q}_2}$ denotes $l_2$, and thus the intersection $\mathbf{p}_f$ is determined to be $\mathbf{u}$.
\end{proof}

\subsection{Line-Binary Search}
\label{supp_sec:line_binary_search}
We introduce line-binary search, an essential technique for 2D point search. Line-binary search identifies a point along a ray that differentiates between inside and outside regions. Unlike general binary search, it operates under the condition that it does not have two endpoints with different inside/outside labels. Given a ray $r$ with starting point $p \in \mathbb{R}^3$ and a direction $d \in \mathbb{R}^3$, the procedure begins with a sparse linear search to define the search space, followed by a binary search to ensure precision. (We assume that the occupancy of $\mathbf{p}$ is already known because practical 2D search always know it.)

It firstly calculates the occupancy of the points $\{\mathbf{p}_i = \mathbf{p} + \frac{Ri}{N}d\}_{i = 1, ..., N}$, where $R \in \mathbb{R}$ and $N \in \mathbb{N}_{>0}$ are parameters representing the maximum range and the number of samples, respectively. And then, it identifies the smallest $i$ for which $\mathbf{p}_i$ have a different occupancy than $\mathbf{p}$. ($\mathbf{p}_0 := \mathbf{p}$) The interval $[\mathbf{p}_{i-1}, \mathbf{p}_i)$ then becomes the search space for the binary search. However, sometimes such $i$ does not exist because a sharp feature is farther than $R$, the method positions the output as far from $R$ as possible. Therefore, $i$ is chosen to be $N$, giving the farthest search space.

During the binary search, the occupancy of the middle point is compared to the endpoint closer to $\mathbf{p}$, and the search space is adjusted accordingly. After the iterations, the output is selected as the endpoint closer to $\mathbf{p}$ that has the same occupancy as $\mathbf{p}$. Therefore, all line-binary searches during 2D point search find points with same occupancy as $\mathbf{m}$. This consistency is crucial to avoid divergence. If $\mathbf{q}$ has a different occupancy from $\mathbf{m}$ in step (1), then $\mathbf{p}_1$ or $\mathbf{p}_2$ is not detected correctly, resulting in an incorrect $\mathbf{p}_f$.

\subsection{Practical Implementation of 2D Point Search}
Assuming the boundary consists of two lines passing through distinct 1D points, $\mathbf{q}$ does not need to lie on the isocurve but must be inside the triangle $\mathbf{p}_{e_1}\mathbf{u}\mathbf{p}_{e_2}$. If step (2) correctly detects two points $\mathbf{q}_1$ and $\mathbf{q}_2$, then the lines $l_1$ and $l_2$ are accurately determined, setting $\mathbf{p}_f$ to $\mathbf{u}$. Therefore, the practical goal of step (1) is to place $\mathbf{q}$ inside the triangle $\mathbf{p}_{e_1}\mathbf{u}\mathbf{p}_{e_2}$ or confirm that $l_1$ and $l_2$ are identical. Nevertheless, by locating $\mathbf{q}$ as far from $l$ as possible, we enhance accuracy by avoiding floating-point operations with similar numbers. Thus, we set the maximum range of the line-binary search on step (1) to 0.8 (with the side length of a grid cell being 1), ensuring sufficient range for accurate the floating-points operations. Additionally, line-binary search in step (1) consists of 4 and 11 iterations for linear and binary search, respectively.

Also, under the assumption that the boundary is composed of two lines, the maximum width (parallel to $l$) of the triangle $\mathbf{p}_{e_1}\mathbf{u}\mathbf{p}_{e_2}$, is equal to the distance between $\mathbf{p}_{e_1}$ and $\mathbf{p}_{e_2}$. Considering that this distance can be up to $\sqrt{2}$, we set the maximal distance of the line-binary search on step (2) to $\frac{\sqrt{2}}{2}$. Additionally, line-binary search on step (2) consists of 3 and 12 iterations for linear and binary search, respectively.

\subsection{Polygonization}
\label{supp_sec:polygonization}
We employ the same quad splitting technique of Intersection-free Contouring~\cite{ju2006intersection}. Intersection-free Contouring~\cite{ju2006intersection} introduces a method for splitting a quad into two or four triangles. However, it does not explicitly address the necessity of splitting into four triangles on uniform grids, a requirement demonstrated by~\cite{wang2009intersection}. This study offers a robust and efficient implementation for determining quad splitting of~\cite{ju2006intersection}.

Let $\mathbf{p}_{c_1}\mathbf{p}_{c_2}\mathbf{p}_{c_3}\mathbf{p}_{c_4}$ be a quad composed of 3D points of adjacent cells to a grid edge $e = (v^\text{in}v^\text{out}) \in \mathcal{E}^{S}$ with 1D point $\mathbf{p}_{e}$. (The normal of the oriented triangle composed of center points of grid cell $c_1$, $c_2$, and $c_3$ in order is identical to the direction from $\mathbf{p}_{v^\text{in}}$ to $\mathbf{p}_{v^\text{out}}$.) Then, dual contouring can make the output triangles in two ways: ($\mathbf{p}_{c_1}\mathbf{p}_{c_2}\mathbf{p}_{c_3}$ and $\mathbf{p}_{c_1}\mathbf{p}_{c_3}\mathbf{p}_{c_4}$) (\emph{case 1}) or ($\mathbf{p}_{c_1}\mathbf{p}_{c_2}\mathbf{p}_{c_4}$ and $\mathbf{p}_{c_2}\mathbf{p}_{c_3}\mathbf{p}_{c_4}$) (\emph{case 2}). Here, ~\cite{ju2006intersection} has proven that the output mesh becomes intersection-free if the triangles are enclosed by an envelope $\mathbf{p}_{v^\text{in}}\mathbf{p}_{v^\text{out}}-\mathbf{p}_{c_1}\mathbf{p}_{c_2}\mathbf{p}_{c_3}\mathbf{p}_{c_4}$, a shape that merges four tetrahedrons $\mathbf{p}_{v^\text{in}}\mathbf{p}_{v^\text{out}}\mathbf{p}_{c_1}\mathbf{p}_{c_2}$, $\mathbf{p}_{v^\text{in}}\mathbf{p}_{v^\text{out}}\mathbf{p}_{c_2}\mathbf{p}_{c_3}$, $\mathbf{p}_{v^\text{in}}\mathbf{p}_{v^\text{out}}\mathbf{p}_{c_3}\mathbf{p}_{c_4}$, and $\mathbf{p}_{v^\text{in}}\mathbf{p}_{v^\text{out}}\mathbf{p}_{c_4}\mathbf{p}_{c_1}$. To satisfy this condition, for some cases, the quad should be divided into four triangles: $\mathbf{p}_{e}\mathbf{p}_{c_1}\mathbf{p}_{c_2}$, $\mathbf{p}_{e}\mathbf{p}_{c_2}\mathbf{p}_{c_3}$, $\mathbf{p}_{e}\mathbf{p}_{c_3}\mathbf{p}_{c_4}$, and $\mathbf{p}_{e}\mathbf{p}_{c_4}\mathbf{p}_{c_1}$ (\emph{case 3}).

On \emph{case 1} and \emph{case 2}, the edges $\mathbf{p}_{c_1}\mathbf{p}_{c_2}$, $\mathbf{p}_{c_2}\mathbf{p}_{c_3}$, $\mathbf{p}_{c_3}\mathbf{p}_{c_4}$, and $\mathbf{p}_{c_4}\mathbf{p}_{c_1}$ made by the two triangles are the parts of the envelope. Therefore, if a diagonal edge is contained by the envelope, then the two triangles are enclosed also. For example, on \emph{case 1}, the diagonal edge is $\mathbf{p}_{c_1}\mathbf{p}_{c_3}$ and this edge should intersect the quad $\mathbf{p}_{c_2}\mathbf{p}_{v^\text{in}}\mathbf{p}_{c_4}\mathbf{p}_{v^\text{out}}$ for being enclosed. \cite{wang2009intersection} presents a robust and efficient method, convex/concave analysis, to determine this intersection.

Let us define the term `concave'. The edge $\mathbf{p}_{c_2}\mathbf{p}_{v^\text{out}}$ is concave if and only if $\mathbf{p}_{c_2}$ is below the oriented triangle $\mathbf{p}_{c_1}\mathbf{p}_{c_3}\mathbf{p}_{v^\text{out}}$(\ref{eq:concave+}) and edge $\mathbf{p}_{c_2}\mathbf{p}_{v^\text{in}}$ is concave if and only if $\mathbf{p}_{c_2}$ is above the oriented triangle $\mathbf{p}_{c_1}\mathbf{p}_{c_3}\mathbf{p}_{v^\text{in}}$(\ref{eq:concave-}). 
\begin{equation}
    (\mathbf{p}_{c_2} - \mathbf{p}_{v^\text{out}})\cdot[(\mathbf{p}_{c_1} - \mathbf{p}_{v^\text{out}})\times(\mathbf{p}_{c_3} - \mathbf{p}_{v^\text{out}})] < 0.
    \label{eq:concave+}
\end{equation}
\begin{equation}
    (\mathbf{p}_{c_2} - \mathbf{p}_{v^\text{in}})\cdot[(\mathbf{p}_{c_1} - \mathbf{p}_{v^\text{in}})\times(\mathbf{p}_{c_3} - \mathbf{p}_{v^\text{in}})] > 0.
    \label{eq:concave-}
\end{equation}

Also, a vertex $\mathbf{p}_{c_2}$ is concave if and only if $\mathbf{p}_{c_2}\mathbf{p}_{v^\text{out}}$ or $\mathbf{p}_{c_2}\mathbf{p}_{v^\text{in}}$ is concave.

\cite{wang2009intersection} has demonstrated that \emph{case 1} is not enclosed by the envelope if either $\mathbf{p}_{c_2}$ or $\mathbf{p}_{c_4}$ is concave, while \emph{case 2} is not enclosed if either $\mathbf{p}_{c_1}$ or $\mathbf{p}_{c_3}$ is concave. If both are not enclosed, then \emph{case 3} is selected, as it is guaranteed to be enclosed by the envelope, albeit at the cost of increased triangle count compared to dual contouring.

The intersection-free property of this technique is proven for cases where maximally one 3D point exists per grid cell. However, Manifold Dual Contouring can locate multiple 3D points inside a grid cell, meaning this technique does not guarantee self-intersection-free results for our method. Nonetheless, quantitative experiment results demonstrate that self-intersections are significantly reduced compared to Manifold Dual Contouring.

\section{Experiment Setup}
\label{supp_sec:exp_setting}

\subsection{Baselines}
We evaluated topology-preserving Marching Cubes~\cite{lewiner2003efficient} using the scikit-image library\footnote{https://github.com/scikit-image/scikit-image}, while Lempitsky's smoothing method~\cite{lempitsky2010surface} was tested with PyMCubes library\footnote{https://github.com/pmneila/PyMCubes}. The code for Manifold Dual Contouring was obtained from FlexiCubes\footnote{https://github.com/nv-tlabs/FlexiCubes}~\cite{flexicubes}. Neural Dual Contouring~\cite{neuraldualcontouring} was tested using NDCx, provided by the authors, and Multiresolution IsoSurface Extraction (MISE)~\cite{occupancenetworks} was evaluated using the original implementation by the authors without changing the parameters.

\subsection{Binary Representation of Neural Implicit Functions}
Given an continuous-valued occupancy function $\phi: \mathbb{R}^3 \rightarrow \mathbb{R}$, we define $\phi': \mathbb{R}^3 \rightarrow \{0\ \text{(out)}, 1\ \text{(in)}\}$ by distinguishing from the isolevel. Occupancy-based models use $\phi'$ as input. For each grid cell, LMC and NDC assign a binary voxel value of $1$ if any of its vertices have a $\phi'$ value of $1$. CDIF calculates the DIF value as the average of $\phi'$ values at $6^3$ (=216) points inside each grid cell. Our model, ODC, utilizes $\phi'$ at arbitrary positions without discretizing the space.

\subsection{Runtime Measurement}
\textcolor{color_4}{
The runtime includes not only the mesh construction time but also the inference time of the occupancy function for query points. This is because mesh construction and occupancy function inference are not separable in our method. Also, since our method queries more points than Marching Cubes, the additional inference time should be considered in practical applications. Therefore, we report the runtime inclusive of inference time to ensure a fair comparison.
(In the SALAD~\cite{koo2023salad} experiment with a resolution of $128^3$, occupancy function inference accounts for 98\% of total runtime in the case of Marching Cubes.)
}

\begin{table}[!t]
    \setlength{\tabcolsep}{7.0pt}
    \small
    \centering
    \caption{\textcolor{color_4}{Results with various resolutions with SALAD~\cite{koo2023salad}. The table presents results for three resolutions: $128^3$, $256^3$, and $512^3$.}}
    \vspace{-\baselineskip}
    \label{tab:salad_scalability}
    \begin{tabular}{cccc|ccc}
        \toprule
        \textbf{Various} & \multicolumn{3}{c}{$|\phi - 0.5|$ $\downarrow$} & \multicolumn{3}{c}{Time (s)} \\
        \textbf{Resolutions}& $128^3$ & $256^3$ & $512^3$ & $128^3$ & $256^3$ & $512^3$ \\
        \midrule
        MC & 0.324 & 0.211 & 0.114 & 1.9 & 14.7 & 117.4 \\
        \rowcolor{Gray}
        ODC & \textbf{0.037} & \textbf{0.011} & \textbf{0.003} & 3.5 & 20.7 & 139.9 \\

        \bottomrule
    \end{tabular}
    \vspace{-\baselineskip}
\end{table}

\section{Additional Experiments}
\label{supp_sec:add_exp}
\textcolor{color_4}{This section presents five additional experiments designed to provide a comprehensive evaluation of our method. These experiments include:}

\begin{enumerate}[leftmargin=*,noitemsep,topsep=0em]
\item \textcolor{color_4}{MISE~\cite{occupancenetworks} without mesh simplification, ensuring its output to have similar number of vertices and triangles as ODC.}

\item \textcolor{color_4}{Various resolutions, assessing the scalability of our method in comparison to Marching Cubes.
}

\item Michelangelo~\cite{zhao2024michelangelo}: A conditional 3D generative model that accepts either text, images, or both as inputs. We use the official pretrained model along with the provided 32 images and 29 texts as inputs.

\item IM-NET~\cite{imnet} is an autoencoder that takes a 3D shape as voxels and decodes it using an MLP-based occupancy function. We use the official models trained for 13 categories of ShapeNet, sampling 100 random shapes for each category, resulting in 1,300 shapes in total.

\item \textcolor{color_4}{DeepSDF~\cite{deepsdf} is an autoencoder that encodes 3D shapes and decodes them using MLP-based signed distance function. We trained DeepSDF using 100 shapes from the ABC dataset~\cite{koch2019abc}.}

\end{enumerate}

\subsection{Results with Various Resolutions}
\label{supp_sec:salad_scalability}

\textcolor{color_4}{
We report the mesh reconstruction accuracy and runtime of MC and ODC for high resolutions of $128^3$, $256^3$, and $512^3$ in the SALAD~\cite{koo2023salad} experiments.
}

\textcolor{color_4}{
According to the quantitative results in Table~\ref{tab:salad_scalability}, even for the highest resolution of $512^3$, our ODC increases computation time by only 20$\%$ compared to MC, while producing a mesh with significantly better fidelity, achieving just 2.6$\%$ of the error compared to the error of MC.
}

\subsection{Evaluating MISE Without Mesh Simplification}
\label{supp_sec:MISE_mesh_simplification}

\textcolor{color_4}{
MISE incorporates a mesh simplification process. To ensure unbiased evaluation, we utilized the official MISE code provided by the authors of Occupancy Networks~\cite{occupancenetworks} without any modifications, including parameter settings. The simplification step reduces the mesh to 5,000 triangles, resulting in MISE generating fewer vertices and triangles compared to MC and ODC. To address the concern about fair comparison in terms of triangle count, we also present quantitative results in the SALAD~\cite{koo2023salad} experiment with MISE while excluding the mesh simplification step in Table~\ref{tab:MISE_mesh_simplification}.
}

\textcolor{color_4}{
Note that without the mesh simplification, MISE now produces meshes with a similar number of vertices and triangles as our ODC output meshes, while not significantly reducing the fitting error, which is still 5.7 times higher than that of ODC. This is because MISE is not designed to capture sharp features. Additionally, MISE produces many self-intersecting meshes, while our ODC prevents self-intersections almost perfectly.
}

\begin{table}[!t]
    \setlength{\tabcolsep}{2.3pt}
    \small
    \centering
    \caption{\textcolor{color_4}{Removal of mesh simplification step from MISE on the SALAD~\cite{koo2023salad} experiment. MISE w/o simpl. denotes MISE without the mesh simplification step.}}
    \vspace{-\baselineskip}
    \label{tab:MISE_mesh_simplification}
    \begin{tabular}{ccccccc}
        \toprule
        & $|\phi - 0.5|$ $\downarrow$ & SI (\%) $\downarrow$ & Man. (\%) $\uparrow$ & \#V & \#T & Time (s) \\
        \midrule
        MISE & 0.237 & 23.1 & 74.3 & 2500 & 5000 & 1.2 \\
        MISE w/o simpl. & 0.211 & 46.9 & 100. & 16261 & 32524 & 3.4 \\
        \rowcolor{Gray}
        ODC & \textbf{0.037} & 0.00 & 100. & 16310 & 32619 & 3.5 \\

        \bottomrule
    \end{tabular}
\end{table}

\subsection{Results with Michelangelo}
\label{supp_sec:michelangelo_results}
\begin{figure*}[!t]
    \centering
    \setlength{\tabcolsep}{0em}
    \def\arraystretch{0.0}

    \begin{tabular}{P{0.125\textwidth}P{0.125\textwidth}P{0.125\textwidth}P{0.125\textwidth}P{0.125\textwidth}P{0.125\textwidth}P{0.125\textwidth}P{0.125\textwidth}}
        \textbf{MC}~\shortcite{lewiner2003efficient} & \textbf{IC}~\shortcite{ju2006intersection} & \textbf{MDC}~\shortcite{manifolddualcontouring} & \textbf{LMC}~\shortcite{lempitsky2010surface} & \textbf{NDC}~\shortcite{neuraldualcontouring} & \textbf{CDIF}~\shortcite{manson2011contouring} & \textbf{MISE}~\shortcite{occupancenetworks} & \textbf{ODC} (ours) \\
        \midrule
        \multicolumn{8}{c}{\includegraphics[width=\textwidth]{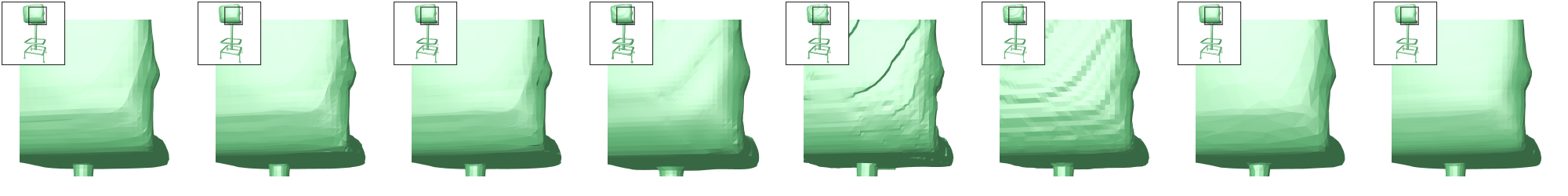}} \\
        \multicolumn{8}{c}{A 3D model of chair; Banqueta.} \\
        \midrule
        \multicolumn{8}{c}{\includegraphics[width=\textwidth]{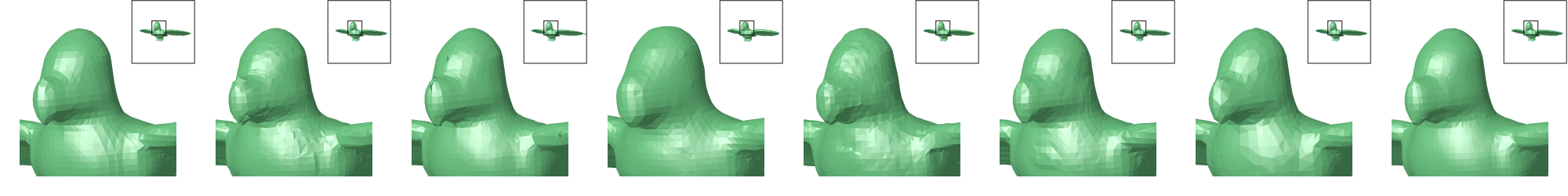}} \\
        \multicolumn{8}{c}{A 3D model of flying monster; Cute monster in the form of flying birds, with 2 feet.} \\
        \midrule
        \multicolumn{8}{c}{\includegraphics[width=\textwidth]{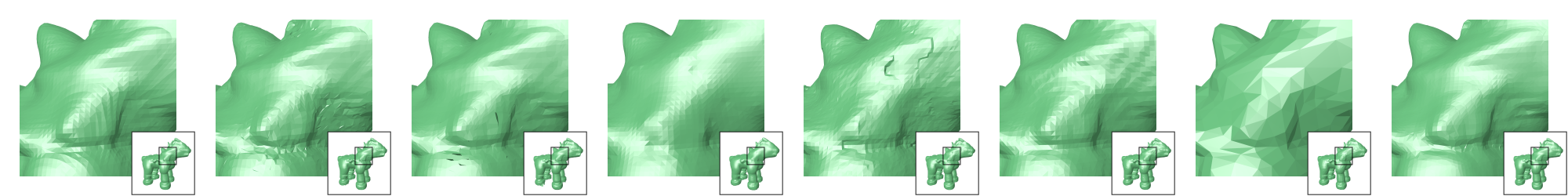}} \\
        \multicolumn{8}{c}{A 3D model of standing monster; Cute four-legged monster, with 4 feet.} \\
        \midrule
        \multicolumn{8}{c}{\includegraphics[width=\textwidth]{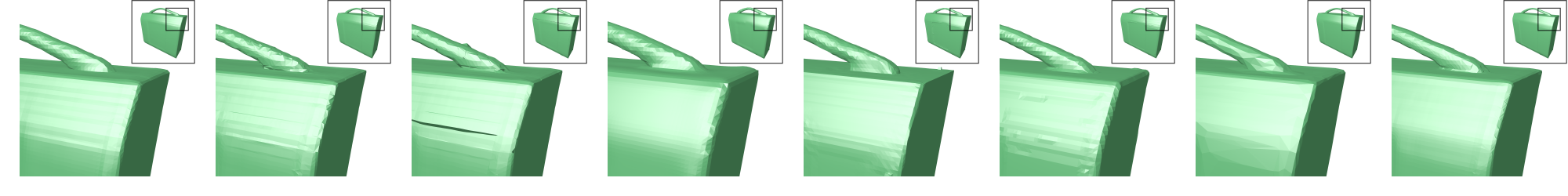}} \\
        \multicolumn{8}{c}{A 3D model of suitcase; Leather Suitcase.} \\
        \midrule
        \multicolumn{8}{c}{\includegraphics[width=\textwidth]{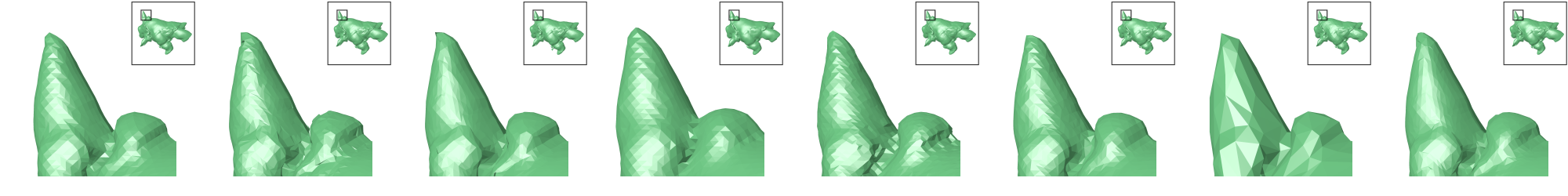}} \\
        \multicolumn{8}{c}{A 3D model of sandshrew monster; Cute monster in the form of mountain armor, with 2 feet and 2 hands.} \\
        \midrule
        \multicolumn{8}{c}{\includegraphics[width=\textwidth]{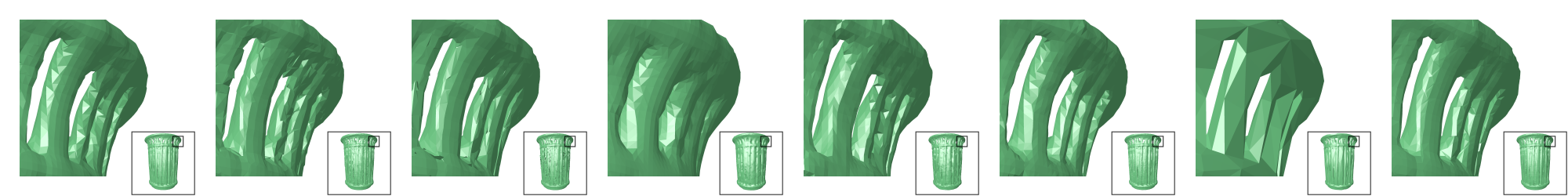}} \\
        \multicolumn{8}{c}{A 3D model of trash barrel; Thomas Steele Litter CRTR-32.} \\
        \midrule
        \multicolumn{8}{c}{\includegraphics[width=\textwidth]{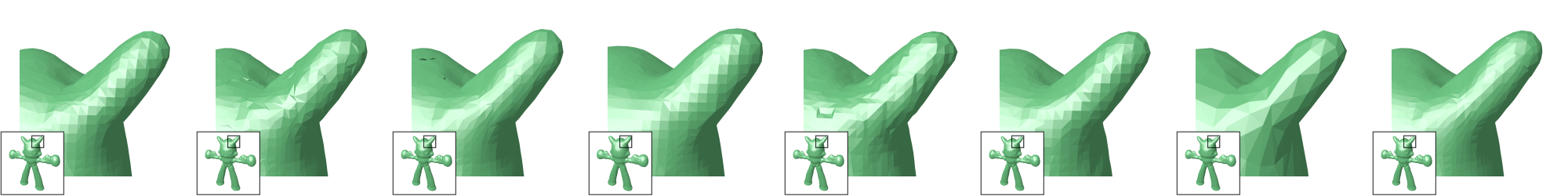}} \\
        \multicolumn{8}{c}{A 3D model of psi monster; The standing humanoid shape of the monster, with 2 feet and 2 hands.} \\
        \midrule
        \multicolumn{8}{c}{\includegraphics[width=\textwidth]{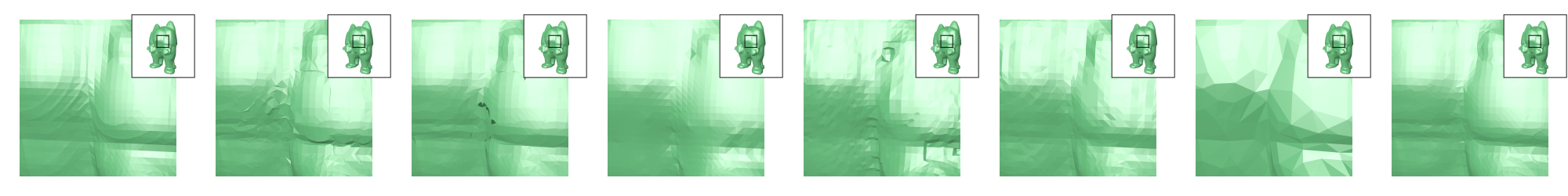}} \\
        \multicolumn{8}{c}{A 3d model of the monster; Cute monster in the form of a bear, with 2 feet and 2 hands.} \\
    \end{tabular}
    \vspace{-0.35cm}
    \caption{Qualitative results with text-conditioned generation of Michelangelo~\cite{zhao2024michelangelo}.}
    \label{fig:Michelangelo_text_result}
\end{figure*}

\begin{figure*}[!t]
    \centering
    \setlength{\tabcolsep}{0em}
    \def\arraystretch{0.0}

    \begin{tabular}{P{0.125\textwidth}P{0.125\textwidth}P{0.125\textwidth}P{0.125\textwidth}P{0.125\textwidth}P{0.125\textwidth}P{0.125\textwidth}P{0.125\textwidth}}
        \textbf{Input} & \textbf{MC}~\shortcite{lewiner2003efficient} & \textbf{MDC}~\shortcite{manifolddualcontouring} & \textbf{LMC}~\shortcite{lempitsky2010surface} & \textbf{NDC}~\shortcite{neuraldualcontouring} & \textbf{CDIF}~\shortcite{manson2011contouring} & \textbf{MISE}~\shortcite{occupancenetworks} & \textbf{ODC} (ours) \\
        \includegraphics[width=0.125\textwidth]{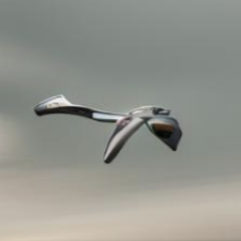} & \multicolumn{7}{c}{\includegraphics[width=0.875\textwidth]{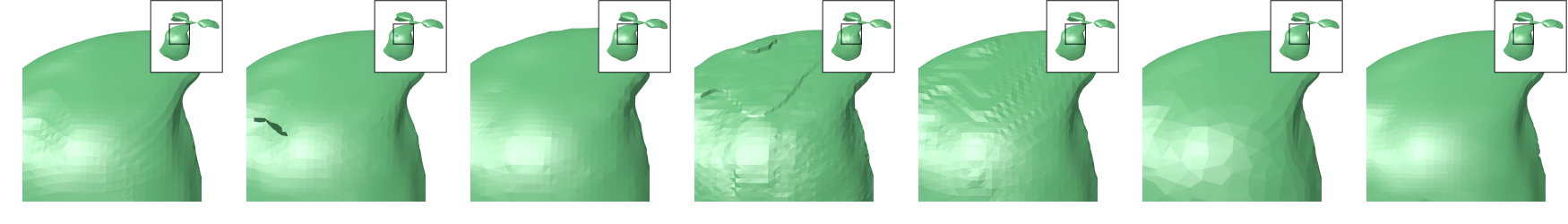}} \\
        \includegraphics[width=0.125\textwidth]{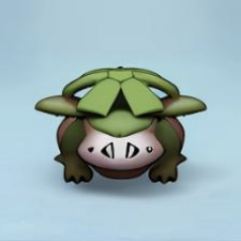} & \multicolumn{7}{c}{\includegraphics[width=0.875\textwidth]{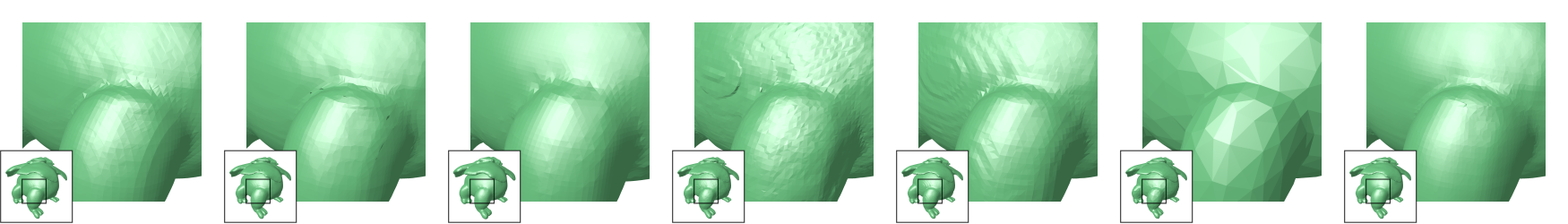}} \\
        \includegraphics[width=0.125\textwidth]{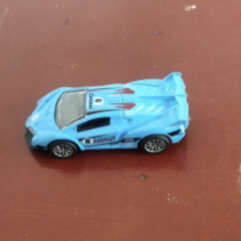} & \multicolumn{7}{c}{\includegraphics[width=0.875\textwidth]{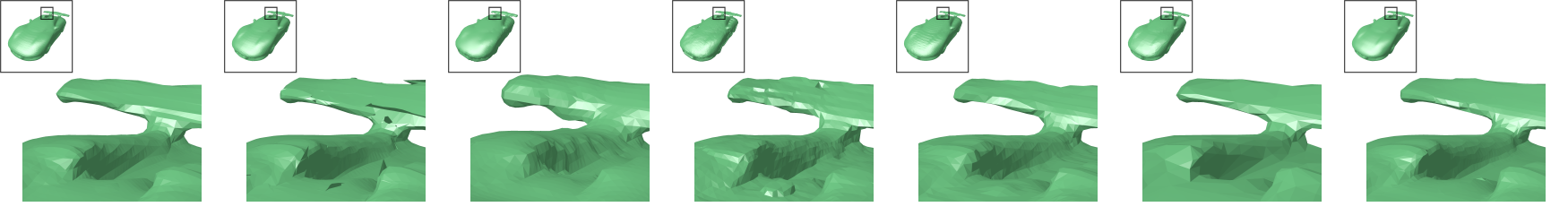}} \\
        \includegraphics[width=0.125\textwidth]{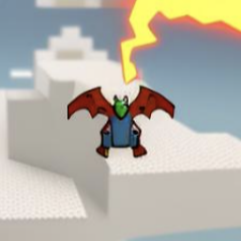} & \multicolumn{7}{c}{\includegraphics[width=0.875\textwidth]{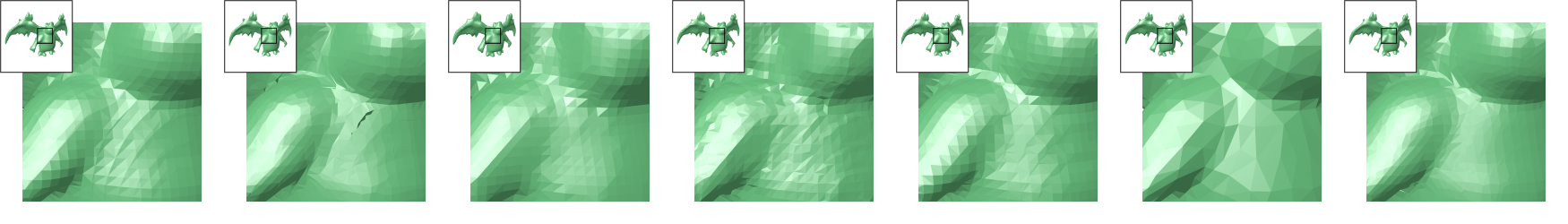}} \\
        \includegraphics[width=0.125\textwidth]{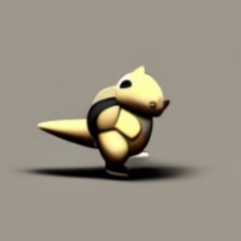} & \multicolumn{7}{c}{\includegraphics[width=0.875\textwidth]{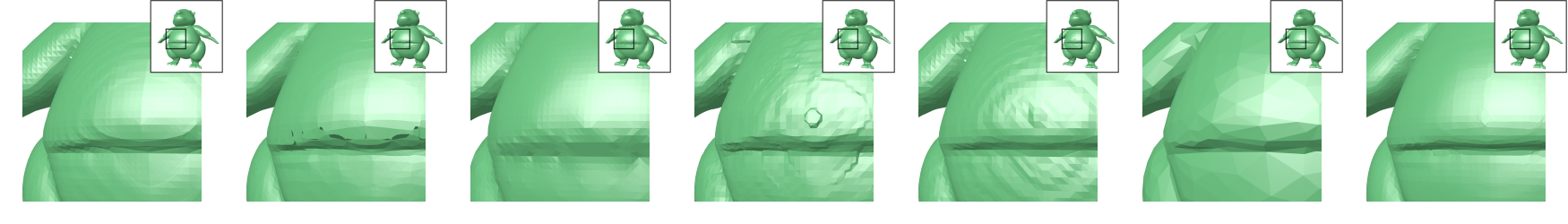}} \\
        \includegraphics[width=0.125\textwidth]{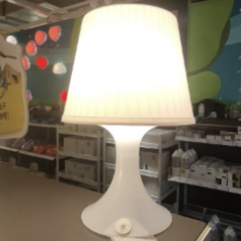} & \multicolumn{7}{c}{\includegraphics[width=0.875\textwidth]{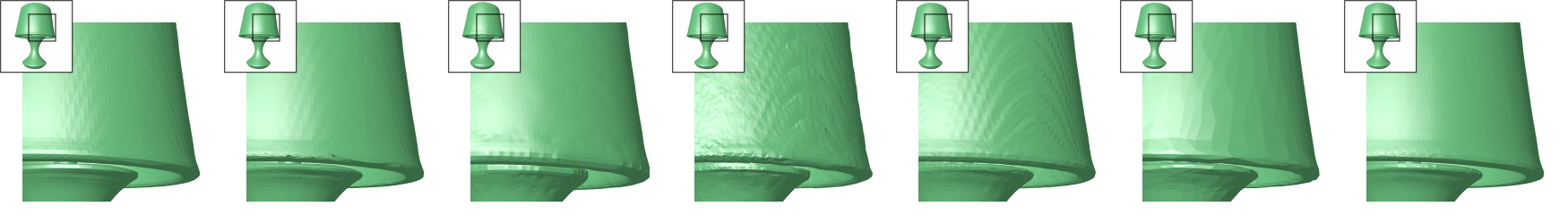}} \\
        \includegraphics[width=0.125\textwidth]{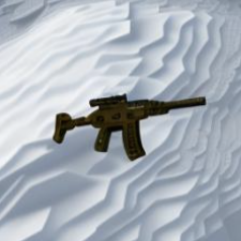} & \multicolumn{7}{c}{\includegraphics[width=0.875\textwidth]{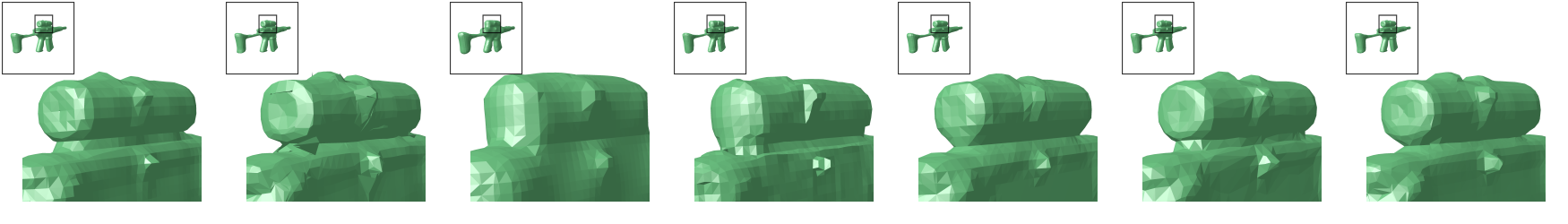}} \\
        \includegraphics[width=0.125\textwidth]{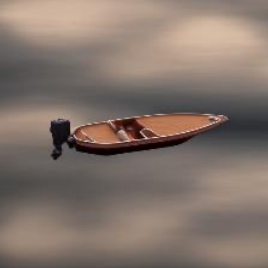} & \multicolumn{7}{c}{\includegraphics[width=0.875\textwidth]{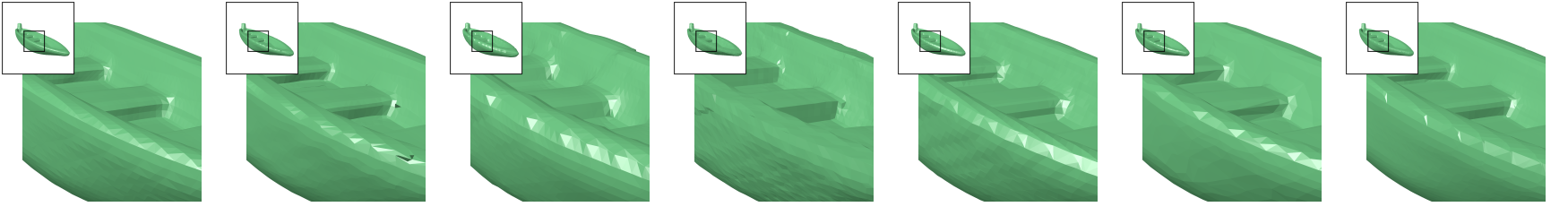}}
    \end{tabular}
    \caption{Qualitative results with image-conditioned generation of Michelangelo~\cite{zhao2024michelangelo}.}
    \label{fig:Michelangelo_image_result}
\end{figure*}
The qualitative results of the experiments on text-conditioned and image-conditioned mesh generation using Michelangelo~\cite{zhao2024michelangelo} are illustrated in Figure~\ref{fig:Michelangelo_text_result} and Figure~\ref{fig:Michelangelo_image_result}. The outputs from MC~\cite{lewiner2003efficient}, IC~\cite{ju2006intersection} (text-conditioned), and CDIF~\cite{manson2011contouring} reveal aliased artifacts, while MDC~\cite{manifolddualcontouring} shows self-intersections of mesh faces, indicated by darker triangles in the figure. LMC~\cite{lempitsky2010surface} produces overly smoothed and coarse shapes. Additionally, NDC~\cite{neuraldualcontouring} exhibits limited adaptability to ShapeNet-based models, as can be seen from the angular patterns on slightly inclined geometry. MISE~\cite{occupancenetworks} presents surfaces with cracked textures. In contrast, our ODC consistently creates meshes with significantly more realistic shapes.

The quantitative results can be found in Table~\ref{tab:michelangelo_result}. Our ODC method demonstrates over a two-fold reduction in fitting errors $|\phi - 0.5|$ on average, indicating higher fidelity to the implicit function compared to other methods. Out of the 61 ODC output meshes, all are manifold, with only one showing self-intersections. The number of triangles is similar across all baselines, except for MISE, which uses mesh simplification techniques to maintain a specific triangle count. The runtime for ODC is about 2.5 times slower than MC.

\begin{table}[!t]
    \setlength{\tabcolsep}{2.8pt}
    \small
    \centering
    \caption{Quantitative results with Michelangelo~\cite{zhao2024michelangelo} on text-conditioned generation and image-conditioned generation tasks. The text-conditioned generation is evaluated with 29 inputs, and image-conditioned generation is evaluated with 32 samples.}
    \vspace{-\baselineskip}
    \label{tab:michelangelo_result}
    \begin{tabular}{ccccccc}
        \toprule
        \textbf{Text-Cond.} & $|\phi - 0.5|$ $\downarrow$ & SI (\%) $\downarrow$ & Man. (\%) $\uparrow$ & \#V & \#T & Time (s) \\
        \midrule
        MC & 1.633 & 0.00 & 100. & 27913 & 55840 & 1.9 \\
        IC & 1.510 & 0.00 & 27.5 & 27946 & 56069 & 2.1 \\
        MDC & 1.525 & 100. & 100. & 27906 & 55802 & 2.2 \\
        LMC & 7.998 & 0.00 & 44.8 & 24760 & 49507 & 7.0 \\
        NDC & 2.523 & 44.8 & 31.0 & 27505 & 55174 & 0.8 \\
        CDIF & 1.499 & 0.00 & 100. & 27284 & 54575 & 6.2 \\
        MISE & 1.269 & 27.5 & 48.2 & 2497 & 5000 & 1.2 \\
        \rowcolor{Gray}
        ODC & \textbf{0.592} & 3.45 & 100. & 27951 & 55891 & 4.8 \\
        \bottomrule
        \toprule
        \textbf{Image-Cond.} & $|\phi - 0.5|$ $\downarrow$ & SI (\%) $\downarrow$ & Man. (\%) $\uparrow$ & \#V & \#T & Time (s) \\
        \midrule
        MC & 1.599 & 0.00 & 100. & 25277 & 50548 & 2.1 \\
        IC & 1.439 & 0.00 & 28.1 & 25329 & 50683 & 2.3 \\
        MDC & 1.404 & 100. & 100. & 25282 & 50553 & 2.5 \\
        LMC & 9.447 & 0.00 & 65.6 & 22659 & 45317 & 6.2 \\
        NDC & 3.074 & 34.3 & 34.3 & 25031 & 50088 & 0.8 \\
        CDIF & 1.602 & 0.00 & 100. & 24784 & 49559 & 6.3 \\
        MISE & 1.254 & 21.8 & 71.8 & 2499 & 5000 & 1.2 \\
        \rowcolor{Gray}
        ODC & \textbf{0.604} & 0.00 & 100. & 25300 & 50589 & 5.1 \\
        \bottomrule
    \end{tabular}
\end{table}

\subsection{Results with IM-NET}
\label{supp_sec:imnet_results}
\begin{figure*}[!t]
    \centering
    \setlength{\tabcolsep}{0em}
    \def\arraystretch{0.0}

    \begin{tabular}{P{0.125\textwidth}P{0.125\textwidth}P{0.125\textwidth}P{0.125\textwidth}P{0.125\textwidth}P{0.125\textwidth}P{0.125\textwidth}P{0.125\textwidth}}
        \textbf{MC}~\shortcite{lewiner2003efficient} & \textbf{IC}~\shortcite{ju2006intersection} & \textbf{MDC}~\shortcite{manifolddualcontouring} & \textbf{LMC}~\shortcite{lempitsky2010surface} & \textbf{NDC}~\shortcite{neuraldualcontouring} & \textbf{CDIF}~\shortcite{manson2011contouring} & \textbf{MISE}~\shortcite{occupancenetworks} & \textbf{ODC} (ours) \\
        \multicolumn{8}{c}{\includegraphics[width=\textwidth]{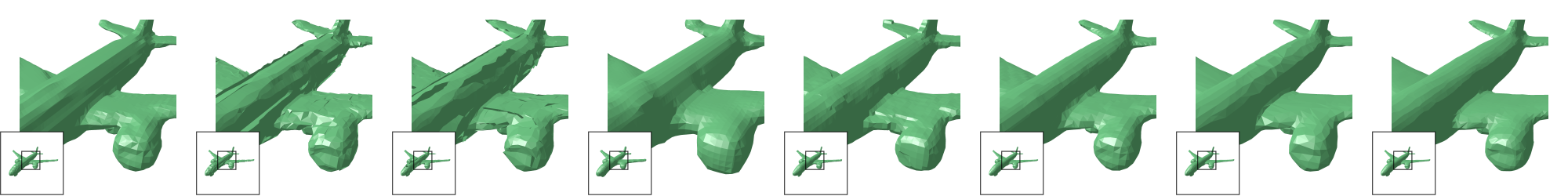}} \\
        \multicolumn{8}{c}{\includegraphics[width=\textwidth]{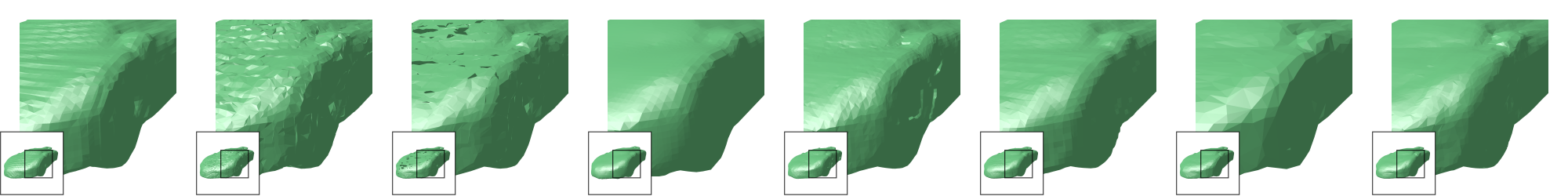}} \\
        \multicolumn{8}{c}{\includegraphics[width=\textwidth]{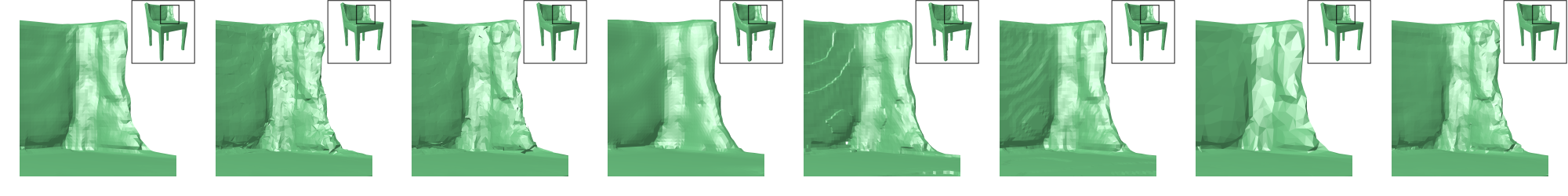}} \\
        \multicolumn{8}{c}{\includegraphics[width=\textwidth]{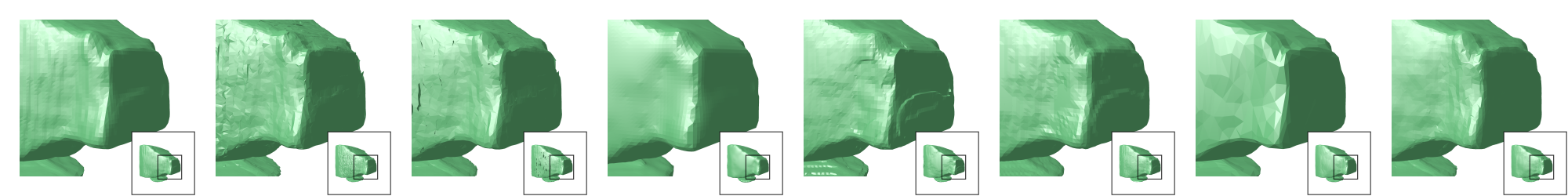}} \\
        \multicolumn{8}{c}{\includegraphics[width=\textwidth]{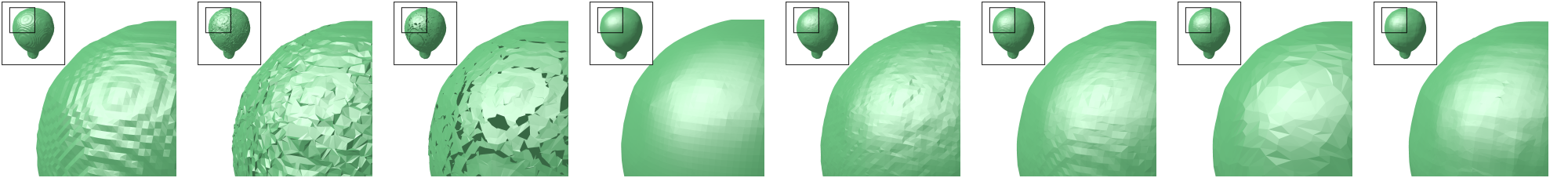}} \\
        \multicolumn{8}{c}{\includegraphics[width=\textwidth]{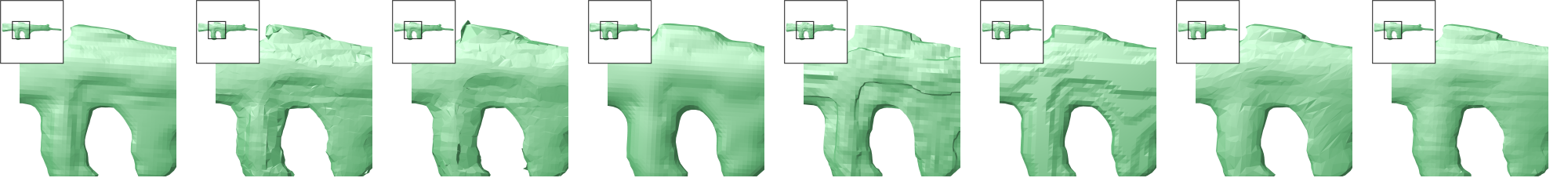}} \\
        \multicolumn{8}{c}{\includegraphics[width=\textwidth]{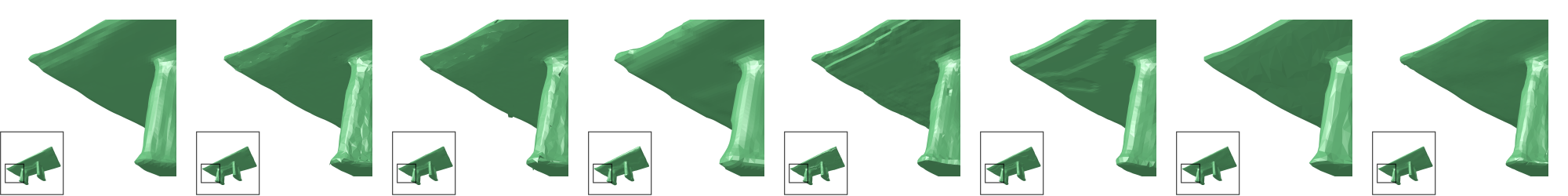}} \\
        \multicolumn{8}{c}{\includegraphics[width=\textwidth]{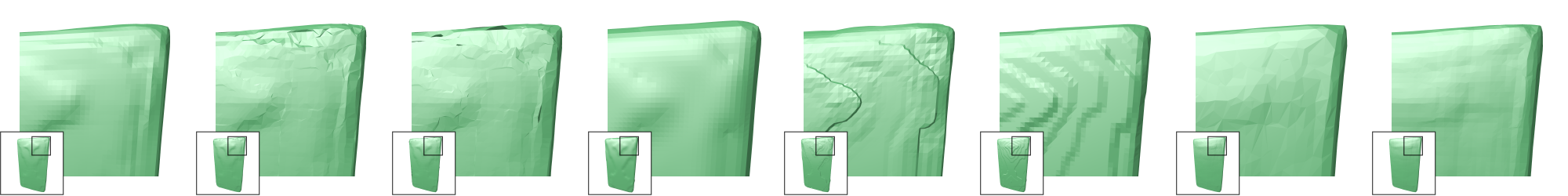}} \\
        \multicolumn{8}{c}{\includegraphics[width=\textwidth]{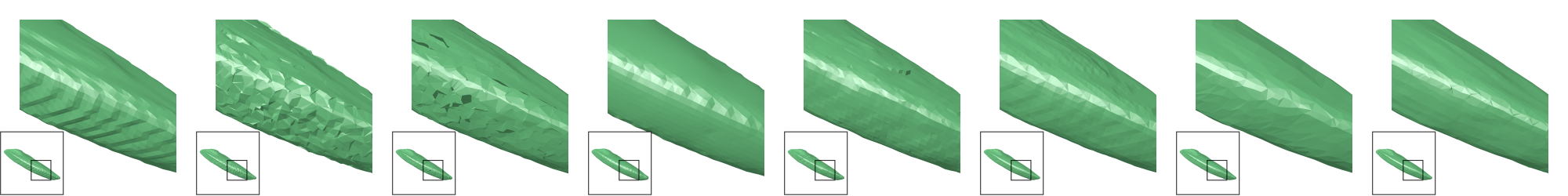}}
    \end{tabular}
    \caption{Qualitative results with IM-NET~\cite{imnet} pretrained on ShapeNet~\cite{chang2015shapenet}.}
    \label{fig:IMNET_result}
\end{figure*}


The qualitative results of the experiment using the conventional MLP-based occupancy function, IM-NET~\cite{imnet}, are depicted in Figure~\ref{fig:IMNET_result}. The outputs from MC~\cite{lewiner2003efficient}, IC~\cite{ju2006intersection}, and CDIF~\cite{manson2011contouring} exhibit staircase-like artifacts, while MDC~\cite{manifolddualcontouring} displays self-intersections of mesh faces, marked by darker triangles in the figure. LMC~\cite{lempitsky2010surface} produces overly-smoothed and coarse shapes. Additionally, NDC~\cite{neuraldualcontouring} shows limited adaptability to ShapeNet-based models, evident from angular patterns observed on slightly inclined geometry. MISE~\cite{occupancenetworks} presents surfaces with cracked textures. In contrast, our ODC consistently generates meshes with significantly more realistic shapes.

The quantitative results are also presented in Table~\ref{tab:imnet_iso}. In terms of fidelity to the given implicit function, our ODC exhibits more than a 10-fold reduction in fitting errors $|\phi - 0.5|$ on average compared to other methods, except for MISE, which demonstrates a 2.5-fold larger fitting error than ODC. Among the 1,300 ODC output meshes, all are manifold, with only 4 displaying self-intersections. The number of triangles remains comparable across all baselines, except for MISE, which employs mesh simplification techniques to achieve a specific triangle count. The runtime of ODC is approximately twice as slow as MC but still under 2 seconds on average.

\subsection{Results with DeepSDF}
\label{supp_sec:deepsdf_results}
\begin{figure*}[!]
    \centering
    \setlength{\tabcolsep}{0em}
    \def\arraystretch{0.0}

    \begin{tabular}{P{0.111\textwidth}P{0.111\textwidth}P{0.111\textwidth}P{0.111\textwidth}P{0.111\textwidth}P{0.111\textwidth}P{0.111\textwidth}P{0.111\textwidth}P{0.112\textwidth}}
        \textbf{MC}~\shortcite{lewiner2003efficient} & \textbf{IC}~\shortcite{ju2006intersection} & \textbf{MDC}~\shortcite{manifolddualcontouring} & \textbf{LMC}~\shortcite{lempitsky2010surface} & \textbf{NDC}~\shortcite{neuraldualcontouring} & \textbf{CDIF}~\shortcite{manson2011contouring} & \textbf{MISE}~\shortcite{occupancenetworks} & \textbf{PoNQ}~\shortcite{maruani2024ponq} & \textbf{ODC} (ours) \\
        \multicolumn{9}{c}{\includegraphics[width=\textwidth]{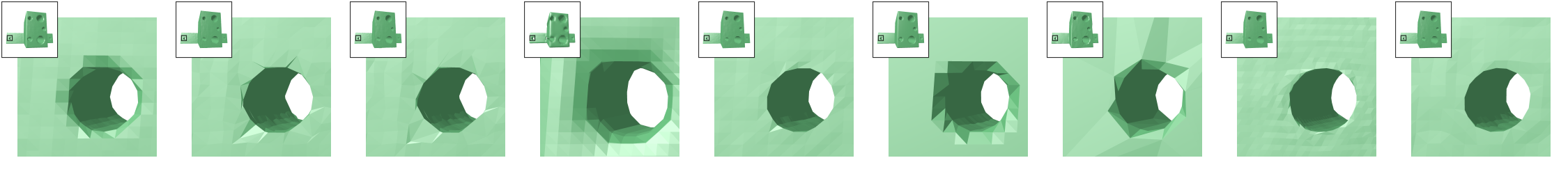}} \\
        \multicolumn{9}{c}{\includegraphics[width=\textwidth]{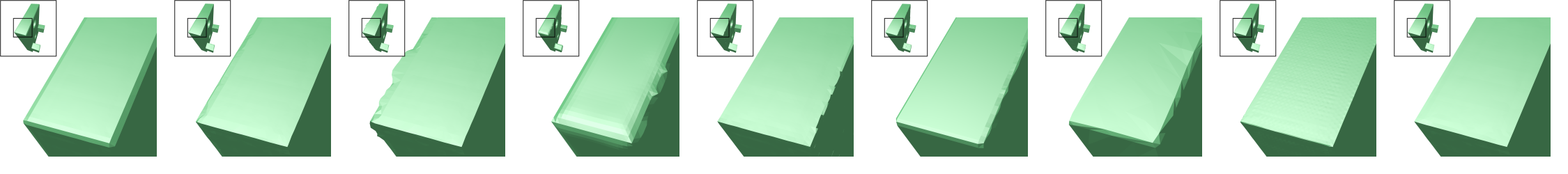}} \\
        \multicolumn{9}{c}{\includegraphics[width=\textwidth]{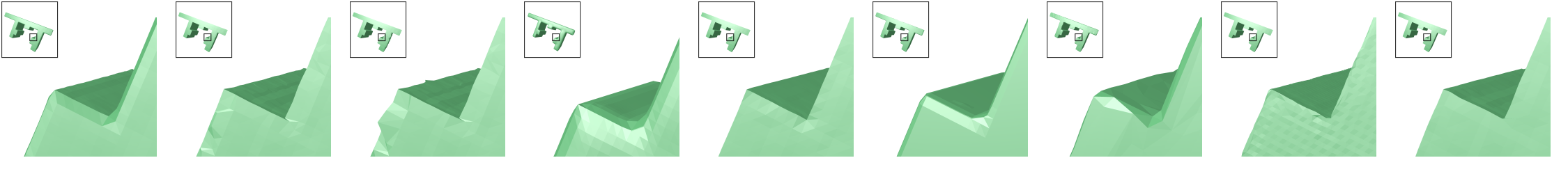}} \\
    \end{tabular}
    \vspace{-\baselineskip}
    \caption{\textcolor{color_4}{Qualitative results with DeepSDF~\cite{deepsdf} encoding shapes from ABC dataset~\cite{koch2019abc}.}}
    \label{fig:deepsdf_result}
    \vspace{0.6cm}
\end{figure*}

\textcolor{color_4}{
To demonstrate that our method also outperforms with neural signed distance functions, we present results with DeepSDF~\cite{deepsdf}. The occupancy is 1 (in) if and only if the signed distance is negative.
The qualitative results are shown in Figure ~\ref{fig:deepsdf_result}. The outputs from MC~\cite{lewiner2003efficient} fail to capture sharp features, and methods designed specifically for occupancy functions—such as LMC~\cite{lempitsky2010surface}, CDIF~\cite{manson2011contouring} and MISE~\cite{occupancenetworks}—also do not effectively capture sharp details. Although IC~\cite{ju2006intersection} and MDC~\cite{manifolddualcontouring} succeed in capturing edge sharpness, they still fail to represent the edges clearly. The supervised methods, NDC~\cite{neuraldualcontouring} and PoNQ~\cite{maruani2024ponq}, exhibit artifacts resulting from unclear edge representations and waviness in flat regions, respectively. In contrast, our method, ODC, effectively captures sharp edges and accurately reconstructs flat regions, demonstrating its adaptability to MLP-based signed distance functions.}

\textcolor{color_4}{The quantitative results are presented in Table~\ref{tab:deepsdf}. We report $|\psi|$ values since the input signed distance function $\psi$ encodes shapes along the 0-level, unlike occupancy functions that encode along the 0.5-level. Our method achieves more than a threefold reduction in $|\psi|$ values compared to any baseline, demonstrating the effectiveness of our 1D and 2D point searches. Additionally, our method produces fully manifold meshes with no self-intersections, highlighting the strength of the IC-based polygonization. The number of triangles is comparable to other baselines, except for MISE and PoNQ. Although the runtime for ODC is slower than MC, it still completes within 3 seconds.}

\begin{table}[!t]
    \setlength{\tabcolsep}{2.9pt}
    \small
    \centering
    \caption{\textcolor{color_4}{Quantitative results with DeepSDF~\cite{koo2023salad} using 100 shapes from the ABC dataset~\cite{koch2019abc}. $^\ast$LMC~\cite{lempitsky2010surface} failed to reconstruct for one shape.}}
    \label{tab:deepsdf}
    \vspace{-\baselineskip}
    \begin{tabular}{ccccccc}
        \toprule
        \textbf{DeepSDF} & $|\psi|$ ($\times 10^4$) $\downarrow$ & SI (\%) $\downarrow$ & Man. (\%) $\uparrow$ & \#V & \#T & Time (s) \\
        \midrule
        MC & 1.261 & 0.00 & 100. & 31610 & 63223 & 0.9 \\
        IC & 0.707 & 0.00 & 72.0 & 31621 & 63254 & 0.9 \\
        MDC & 0.892 & 100. & 100. & 31610 & 63221 & 0.9 \\
        $^\ast$LMC & 42.01 & 0.00 & 87.9 & 28194 & 56389 & 6.2 \\
        NDC & 0.700 & 83.0 & 72.0 & 31604 & 63221 & 0.8 \\
        CDIF & 7.017 & 0.00 & 100. & 31079 & 62159 & 3.2 \\
        MISE & 2.795 & 8.00 & 98.0 & 2499 & 5000 & 1.1 \\
        PoNQ & 1.349 & 0.00 & 40.0 & 119466 & 238945 & 50. \\
        \rowcolor{Gray}
        ODC & \textbf{0.211} & 0.00 & 100. & 31614 & 63230 & 2.2 \\
        \bottomrule
    \end{tabular}
\end{table}
\begin{table*}[!t]
    \setlength{\tabcolsep}{5.0pt}
    \small
    \centering
    \vspace{0.7cm}
    \caption{Quantitative results with IM-NET~\cite{imnet} pretrained on ShapeNet~\cite{chang2015shapenet}.}
    \label{tab:imnet_iso}
    \begin{tabular}{cccccccccccccc|c}
        \toprule
        \multirow{2}{*}{$|\phi - 0.5| \downarrow$} & Air- & \multirow{2}{*}{Bench} & Cabi- & \multirow{2}{*}{Car} & \multirow{2}{*}{Chair} & Dis- & \multirow{2}{*}{Lamp} & Loud- & \multirow{2}{*}{Rifle} & \multirow{2}{*}{Sofa} & \multirow{2}{*}{Table} & Tele- & Water- & \multirow{2}{*}{\textbf{Mean}}\\
        & plane & & net &  &  & play &  & speaker & & & & phone & craft & \\
        \midrule
        MC & 0.143 & 0.138 & 0.175 & 0.130 & 0.109 & 0.160 & 0.055 & 0.128 & 0.096 & 0.151 & 0.166 & 0.239 & 0.080 & 0.136 \\
        IC & 0.153 & 0.141 & 0.171 & 0.143 & 0.108 & 0.159 & 0.058 & 0.127 & 0.094 & 0.149 & 0.161 & 0.238 & 0.083 & 0.137 \\
        MDC & 0.130 & 0.127 & 0.167 & 0.118 & 0.098 & 0.151 & 0.049 & 0.120 & 0.086 & 0.142 & 0.156 & 0.229 & 0.072 & 0.126 \\
        LMC & 0.418 & 0.407 & 0.400 & 0.421 & 0.393 & 0.410 & 0.335 & 0.383 & 0.370 & 0.399 & 0.401 & 0.423 & 0.356 & 0.394 \\
        NDC & 0.165 & 0.194 & 0.257 & 0.164 & 0.169 & 0.231 & 0.095 & 0.196 & 0.146 & 0.224 & 0.238 & 0.300 & 0.119 & 0.192 \\
        CDIF & 0.101 & 0.105 & 0.130 & 0.087 & 0.090 & 0.122 & 0.052 & 0.100 & 0.076 & 0.116 & 0.134 & 0.194 & 0.062 & 0.105 \\
        MISE & 0.034 & 0.025 & 0.022 & 0.034 & 0.026 & 0.021 & 0.019 & 0.024 & 0.018 & 0.021 & 0.022 & 0.021 & 0.021 & 0.024 \\
        \rowcolor{Gray}
        ODC & \textbf{0.017} & \textbf{0.012} & \textbf{0.007} & \textbf{0.009} & \textbf{0.009} & \textbf{0.007} & \textbf{0.009} & \textbf{0.006} & \textbf{0.011} & \textbf{0.007} & \textbf{0.009} & \textbf{0.007} & \textbf{0.010} & \textbf{0.009} \\
        \bottomrule
        \toprule
        
        \multirow{2}{*}{SI (\%) $\downarrow$} & Air- & \multirow{2}{*}{Bench} & Cabi- & \multirow{2}{*}{Car} & \multirow{2}{*}{Chair} & Dis- & \multirow{2}{*}{Lamp} & Loud- & \multirow{2}{*}{Rifle} & \multirow{2}{*}{Sofa} & \multirow{2}{*}{Table} & Tele- & Water- & \multirow{2}{*}{\textbf{Mean}}\\
        & plane & & net &  &  & play &  & speaker & & & & phone & craft & \\
        \midrule
        MC & 0.00 & 0.00 & 0.00 & 0.00 & 0.00 & 0.00 & 0.00 & 0.00 & 0.00 & 0.00 & 0.00 & 0.00 & 0.00 & 0.00 \\
        IC & 0.00 & 0.00 & 0.00 & 0.00 & 0.00 & 0.00 & 0.00 & 0.00 & 0.00 & 0.00 & 0.00 & 0.00 & 0.00 & 0.00 \\
        MDC & 100. & 100. & 100. & 100. & 100. & 100. & 100. & 100. & 100. & 100. & 100. & 100. & 100. & 100. \\
        LMC & 0.00 & 0.00 & 0.00 & 0.00 & 0.00 & 0.00 & 0.00 & 0.00 & 0.00 & 0.00 & 0.00 & 0.00 & 0.00 & 0.00 \\
        NDC & 29.0 & 21.0 & 18.0 & 19.0 & 17.0 & 12.0 & 30.0 & 21.0 & 48.0 & 16.0 & 21.0 & 4.00 & 36.0 & 22.5 \\
        CDIF & 0.00 & 0.00 & 0.00 & 0.00 & 0.00 & 0.00 & 0.00 & 0.00 & 0.00 & 0.00 & 0.00 & 0.00 & 0.00 & 0.00 \\
        MISE & 14.0 & 26.0 & 20.0 & 14.0 & 18.0 & 18.0 & 29.0 & 19.0 & 16.0 & 10.0 & 30.0 & 10.0 & 30.0 & 19.5 \\
        \rowcolor{Gray}
        ODC & 0.00 & 0.00 & 0.00 & 1.00 & 0.00 & 0.00 & 0.00 & 1.00 & 0.00 & 0.00 & 0.00 & 0.00 & 2.00 & 0.30 \\
        \bottomrule
        \toprule
        
        \multirow{2}{*}{Man. (\%) $\uparrow$} & Air- & \multirow{2}{*}{Bench} & Cabi- & \multirow{2}{*}{Car} & \multirow{2}{*}{Chair} & Dis- & \multirow{2}{*}{Lamp} & Loud- & \multirow{2}{*}{Rifle} & \multirow{2}{*}{Sofa} & \multirow{2}{*}{Table} & Tele- & Water- & \multirow{2}{*}{\textbf{Mean}}\\
        & plane & & net &  &  & play &  & speaker & & & & phone & craft & \\
        \midrule
        MC & 100. & 100. & 100. & 100. & 100. & 100. & 100. & 100. & 100. & 100. & 100. & 100. & 100. & 100. \\
        IC & 16.0 & 32.0 & 43.0 & 19.0 & 22.0 & 44.0 & 20.0 & 12.0 & 59.0 & 40.0 & 34.0 & 60.0 & 24.0 & 32.6 \\
        MDC & 100. & 100. & 100. & 100. & 100. & 100. & 100. & 100. & 100. & 100. & 100. & 100. & 100. & 100. \\
        LMC & 57.0 & 66.0 & 69.0 & 72.0 & 73.0 & 73.0 & 59.0 & 30.0 & 85.0 & 76.0 & 75.0 & 91.0 & 54.0 & 67.6 \\
        NDC & 27.0 & 44.0 & 65.0 & 43.0 & 41.0 & 68.0 & 28.0 & 33.0 & 65.0 & 71.0 & 52.0 & 78.0 & 42.0 & 50.5 \\
        CDIF & 100. & 100. & 100. & 100. & 100. & 100. & 100. & 100. & 100. & 100. & 100. & 100. & 100. & 100. \\
        MISE & 81.0 & 69.0 & 72.0 & 78.0 & 69.0 & 83.0 & 67.0 & 57.0 & 91.0 & 86.0 & 74.0 & 95.0 & 70.0 & 76.3 \\
        \rowcolor{Gray}
        ODC & 100. & 100. & 100. & 100. & 100. & 100. & 100. & 100. & 100. & 100. & 100. & 100. & 100. & 100. \\
        \bottomrule
        \toprule
        
        \multirow{2}{*}{\#T} & Air- & \multirow{2}{*}{Bench} & Cabi- & \multirow{2}{*}{Car} & \multirow{2}{*}{Chair} & Dis- & \multirow{2}{*}{Lamp} & Loud- & \multirow{2}{*}{Rifle} & \multirow{2}{*}{Sofa} & \multirow{2}{*}{Table} & Tele- & Water- & \multirow{2}{*}{\textbf{Mean}}\\
        & plane & & net &  &  & play &  & speaker & & & & phone & craft & \\
        \midrule
        MC & 17302 & 37799 & 56874 & 42170 & 45559 & 44132 & 27463 & 64849 & 13665 & 47479 & 48413 & 35767 & 24658 & 38933 \\
        IC & 17520 & 37990 & 57021 & 42515 & 45805 & 44290 & 27721 & 65187 & 13749 & 47633 & 48597 & 35859 & 24895 & 39137 \\
        MDC & 17308 & 37790 & 56874 & 42173 & 45559 & 44136 & 27459 & 64856 & 13645 & 47483 & 48411 & 35769 & 24651 & 38932 \\
        LMC & 15900 & 34649 & 52965 & 38191 & 42327 & 41266 & 25000 & 58042 & 12777 & 44281 & 45418 & 33911 & 21832 & 35889 \\
        NDC & 17228 & 37583 & 56476 & 41867 & 45307 & 43894 & 27263 & 63858 & 13627 & 47219 & 48116 & 35641 & 24438 & 38655 \\
        CDIF & 16701 & 36921 & 56124 & 41702 & 44539 & 43514 & 26871 & 63853 & 13275 & 46717 & 47383 & 34925 & 23984 & 38193 \\
        MISE & 5000 & 5000 & 5000 & 5000 & 5000 & 5000 & 4985 & 5000 & 5000 & 5000 & 5000 & 5000 & 5000 & 4999 \\
        \rowcolor{Gray}
        ODC & 17365 & 37848 & 56919 & 41699 & 45626 & 44178 & 27529 & 64969 & 13671 & 47527 & 48471 & 35793 & 24725 & 38948 \\
        \bottomrule
        \toprule
        
        \multirow{2}{*}{Time (s)} & Air- & \multirow{2}{*}{Bench} & Cabi- & \multirow{2}{*}{Car} & \multirow{2}{*}{Chair} & Dis- & \multirow{2}{*}{Lamp} & Loud- & \multirow{2}{*}{Rifle} & \multirow{2}{*}{Sofa} & \multirow{2}{*}{Table} & Tele- & Water- & \multirow{2}{*}{\textbf{Mean}}\\
        & plane & & net &  &  & play &  & speaker & & & & phone & craft & \\
        \midrule
        MC & 0.933 & 0.916 & 0.906 & 0.889 & 0.913 & 0.862 & 0.915 & 0.943 & 0.900 & 0.885 & 0.909 & 0.893 & 0.890 & 0.904 \\
        IC & 1.120 & 1.109 & 1.154 & 1.179 & 1.148 & 1.129 & 1.074 & 1.180 & 1.052 & 1.148 & 1.214 & 1.126 & 1.103 & 1.133 \\
        MDC & 1.834 & 1.863 & 1.856 & 1.907 & 1.881 & 1.846 & 1.886 & 1.996 & 1.926 & 1.938 & 1.984 & 1.881 & 1.968 & 1.905 \\
        LMC & 2.959 & 5.193 & 6.913 & 6.026 & 6.464 & 6.401 & 4.019 & 7.773 & 2.753 & 6.643 & 6.266 & 5.422 & 3.790 & 5.432 \\
        NDC & 0.790 & 0.807 & 0.778 & 0.798 & 0.759 & 0.754 & 0.786 & 0.774 & 0.778 & 0.762 & 0.771 & 0.778 & 0.792 & 0.779 \\
        CDIF & 1.464 & 2.145 & 2.941 & 2.420 & 2.449 & 2.451 & 1.909 & 3.187 & 1.329 & 2.636 & 2.643 & 2.126 & 1.727 & 2.264 \\
        MISE & 0.932 & 0.982 & 1.084 & 1.028 & 1.014 & 0.998 & 0.951 & 1.154 & 0.861 & 1.017 & 1.056 & 0.944 & 0.926 & 0.996 \\
        \rowcolor{Gray}
        ODC & 1.339 & 1.741 & 2.113 & 1.838 & 1.918 & 1.853 & 1.531 & 2.363 & 1.187 & 1.997 & 1.990 & 1.717 & 1.390 & 1.767 \\
        \bottomrule
    \end{tabular}
\end{table*}



\section{Full Qualitative and Quantitative Results}
\label{supp_sec:more_results}
\ifpaper We provide more qualitative results and detailed quantitative results of the experiments of the main paper.
\else We provide more qualitative results and detailed quantitative results of the experiments in the main paper.
\fi

\subsection{Results with SALAD and 3DShape2VecSet}
\label{supp_sec:unconditional_results}
\begin{figure*}[!t]
    \centering
    \setlength{\tabcolsep}{0em}
    \def\arraystretch{0.0}

    \begin{tabular}{P{0.125\textwidth}P{0.125\textwidth}P{0.125\textwidth}P{0.125\textwidth}P{0.125\textwidth}P{0.125\textwidth}P{0.125\textwidth}P{0.125\textwidth}}
        \textbf{MC}~\shortcite{lewiner2003efficient} & \textbf{IC}~\shortcite{ju2006intersection} & \textbf{MDC}~\shortcite{manifolddualcontouring} & \textbf{LMC}~\shortcite{lempitsky2010surface} & \textbf{NDC}~\shortcite{neuraldualcontouring} & \textbf{CDIF}~\shortcite{manson2011contouring} & \textbf{MISE}~\shortcite{occupancenetworks} & \textbf{ODC} (ours) \\
        \multicolumn{8}{c}{\includegraphics[width=\textwidth]{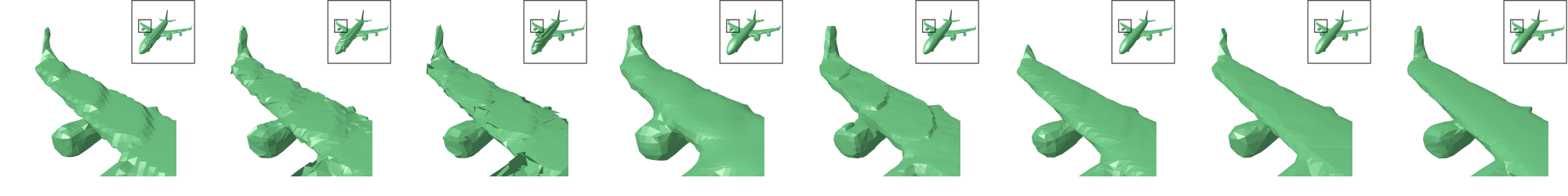}} \\
        \multicolumn{8}{c}{\includegraphics[width=\textwidth]{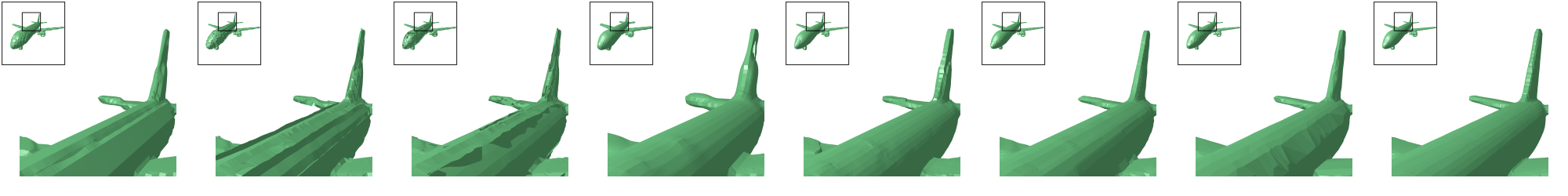}} \\
        \multicolumn{8}{c}{\includegraphics[width=\textwidth]{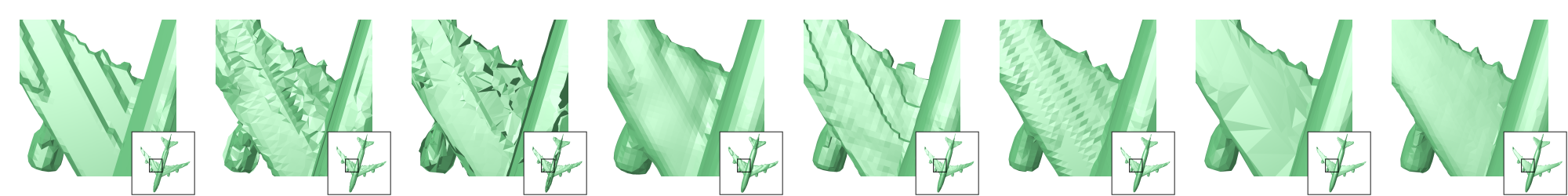}} \\
        \multicolumn{8}{c}{\includegraphics[width=\textwidth]{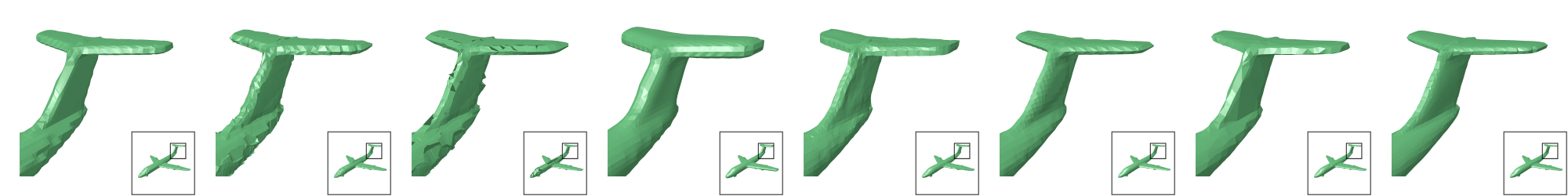}} \\
        \multicolumn{8}{c}{\includegraphics[width=\textwidth]{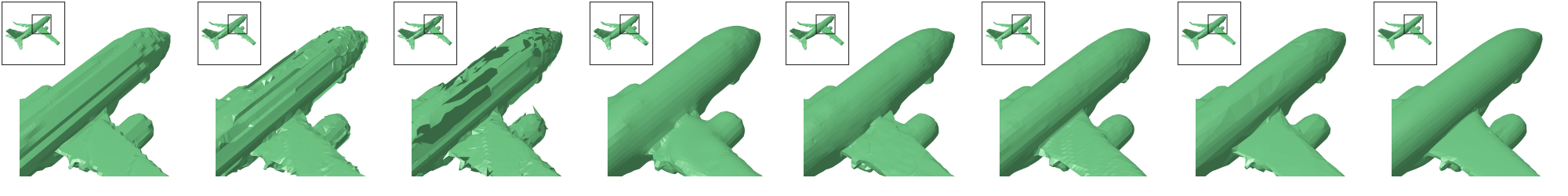}} \\
        \multicolumn{8}{c}{\includegraphics[width=\textwidth]{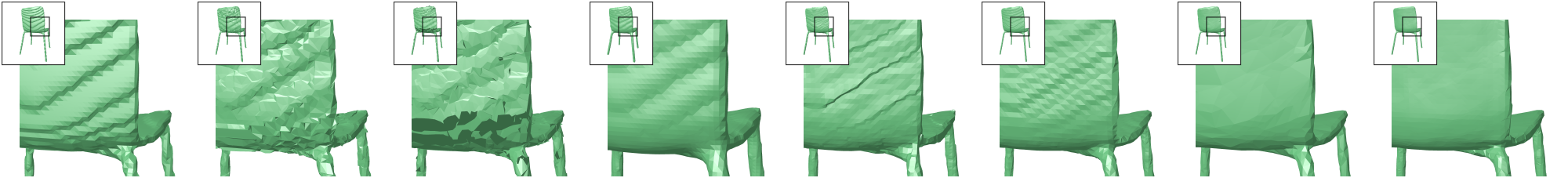}}
    \end{tabular}
    \vspace{-\baselineskip}
    \caption{Qualitative results with unconditional generation of SALAD~\cite{koo2023salad}.}
    \label{fig:SALAD_result}
    \vspace{-\baselineskip}
\end{figure*}

\begin{figure*}[!t]
    \centering
    \setlength{\tabcolsep}{0em}
    \def\arraystretch{0.0}

    \begin{tabular}{P{0.125\textwidth}P{0.125\textwidth}P{0.125\textwidth}P{0.125\textwidth}P{0.125\textwidth}P{0.125\textwidth}P{0.125\textwidth}P{0.125\textwidth}}
        \textbf{MC}~\shortcite{lewiner2003efficient} & \textbf{IC}~\shortcite{ju2006intersection} & \textbf{MDC}~\shortcite{manifolddualcontouring} & \textbf{LMC}~\shortcite{lempitsky2010surface} & \textbf{NDC}~\shortcite{neuraldualcontouring} & \textbf{CDIF}~\shortcite{manson2011contouring} & \textbf{MISE}~\shortcite{occupancenetworks} & \textbf{ODC} (ours) \\
        \multicolumn{8}{c}{\includegraphics[width=\textwidth]{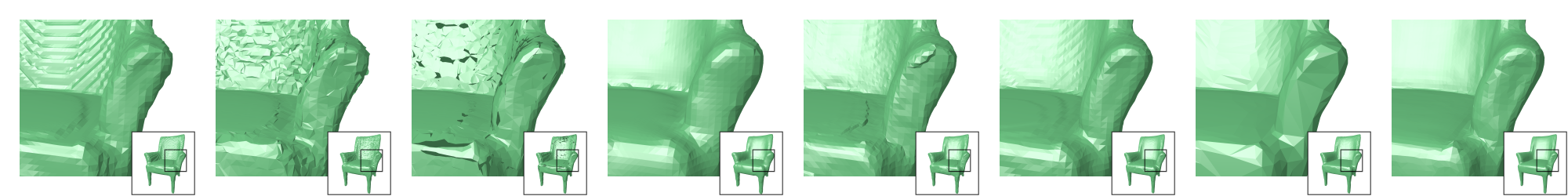}} \\
        \multicolumn{8}{c}{\includegraphics[width=\textwidth]{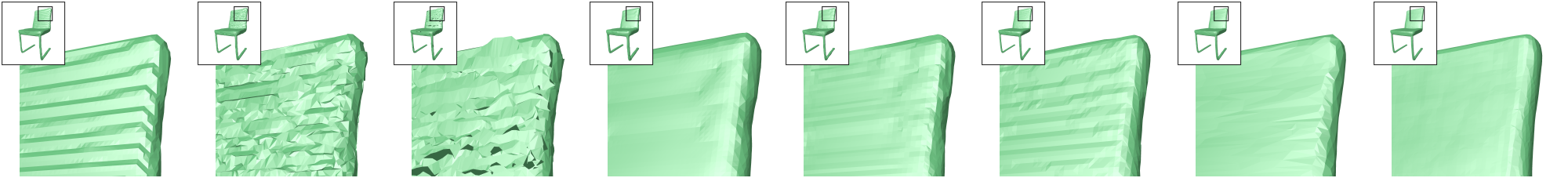}} \\
        \multicolumn{8}{c}{\includegraphics[width=\textwidth]{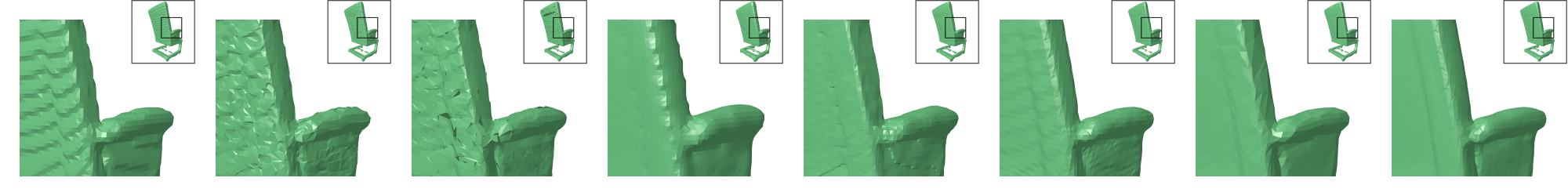}} \\
        \multicolumn{8}{c}{\includegraphics[width=\textwidth]{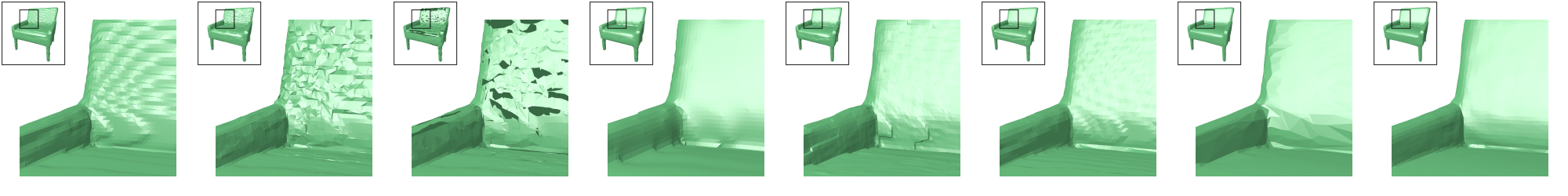}} \\
        \multicolumn{8}{c}{\includegraphics[width=\textwidth]{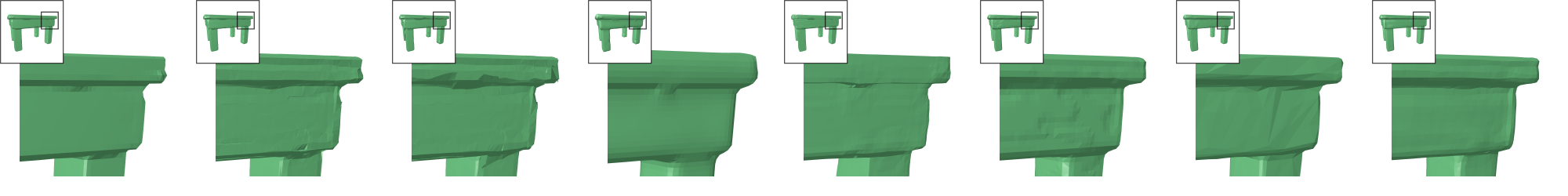}} \\
        \multicolumn{8}{c}{\includegraphics[width=\textwidth]{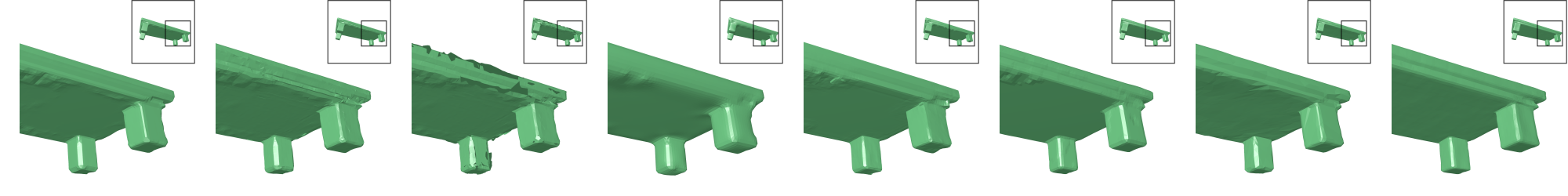}} \\
        \multicolumn{8}{c}{\includegraphics[width=\textwidth]{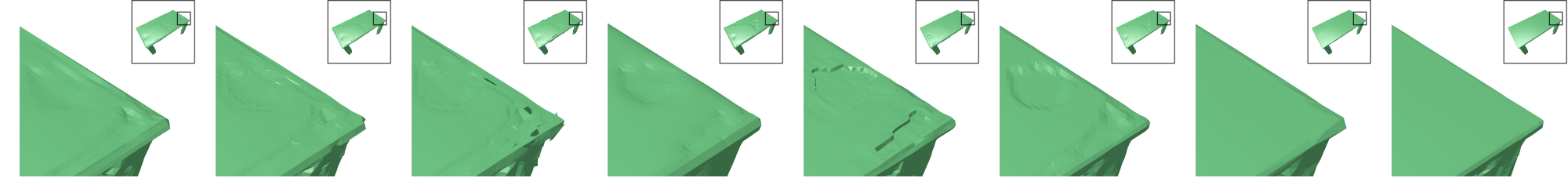}} \\
        \multicolumn{8}{c}{\includegraphics[width=\textwidth]{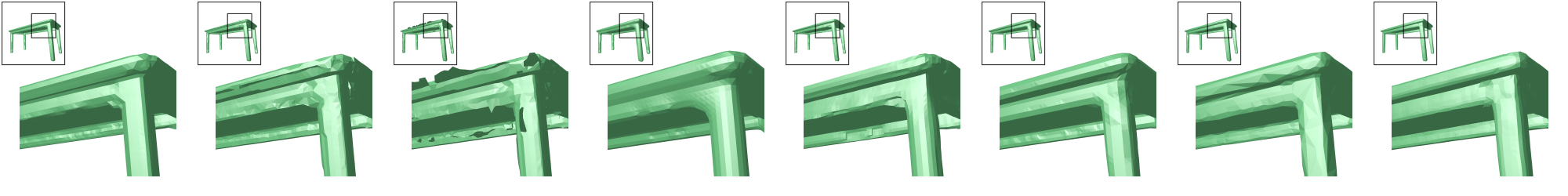}} \\
        \multicolumn{8}{c}{\includegraphics[width=\textwidth]{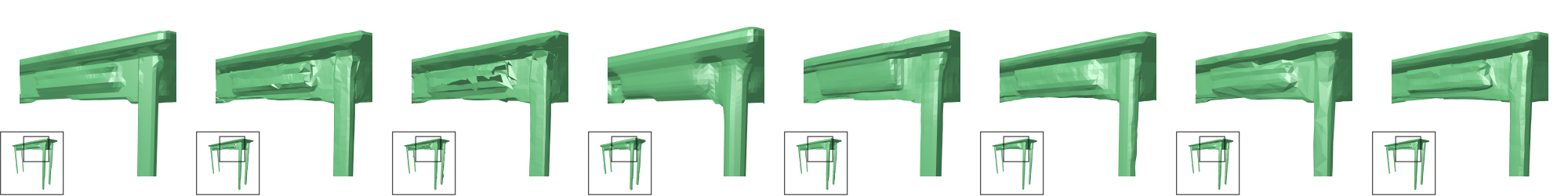}}
    \end{tabular}
    \caption{Qualitative results with unconditional generation of SALAD~\cite{koo2023salad}.}
    \label{fig:SALAD_result2}
\end{figure*}
\begin{figure*}[!t]
    \centering
    \setlength{\tabcolsep}{0em}
    \def\arraystretch{0.0}

    \begin{tabular}{P{0.125\textwidth}P{0.125\textwidth}P{0.125\textwidth}P{0.125\textwidth}P{0.125\textwidth}P{0.125\textwidth}P{0.125\textwidth}P{0.125\textwidth}}
        \textbf{MC}~\shortcite{lewiner2003efficient} & \textbf{IC}~\shortcite{ju2006intersection} & \textbf{MDC}~\shortcite{manifolddualcontouring} & \textbf{LMC}~\shortcite{lempitsky2010surface} & \textbf{NDC}~\shortcite{neuraldualcontouring} & \textbf{CDIF}~\shortcite{manson2011contouring} & \textbf{MISE}~\shortcite{occupancenetworks} & \textbf{ODC} (ours) \\
        \multicolumn{8}{c}{\includegraphics[width=\textwidth]{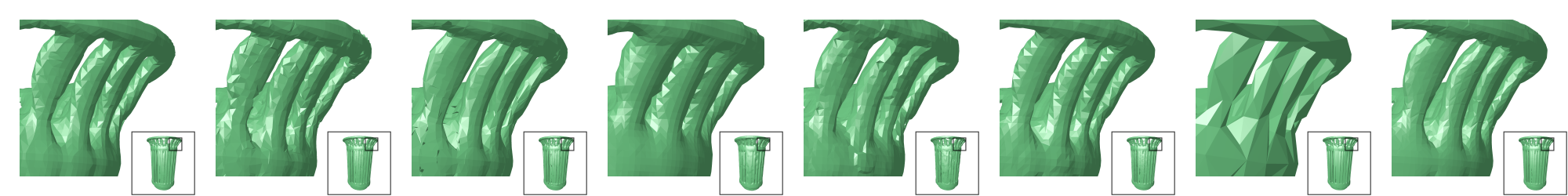}} \\
        \multicolumn{8}{c}{\includegraphics[width=\textwidth]{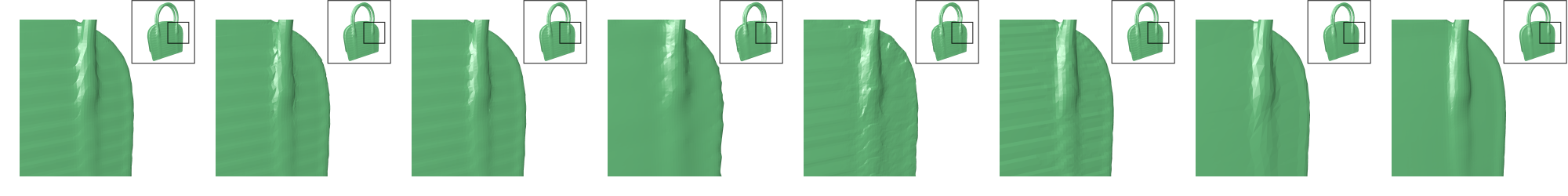}} \\
        \multicolumn{8}{c}{\includegraphics[width=\textwidth]{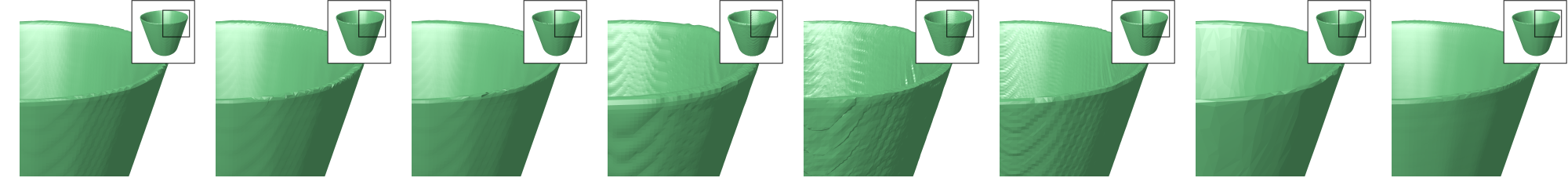}} \\
        \multicolumn{8}{c}{\includegraphics[width=\textwidth]{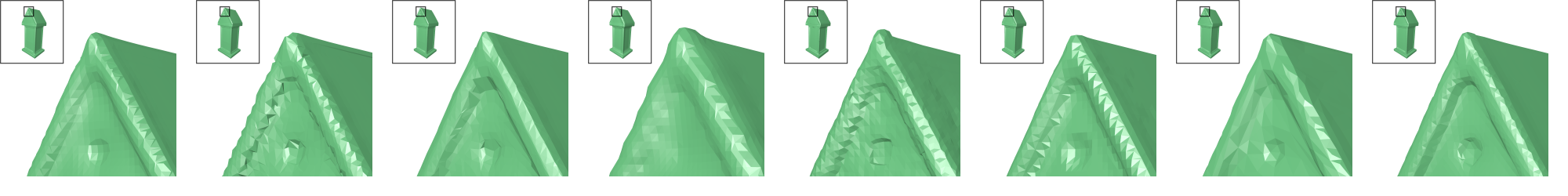}} \\
        \multicolumn{8}{c}{\includegraphics[width=\textwidth]{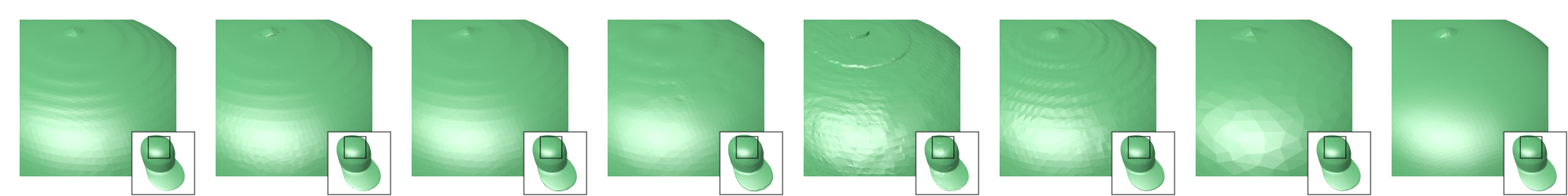}} \\
        \multicolumn{8}{c}{\includegraphics[width=\textwidth]{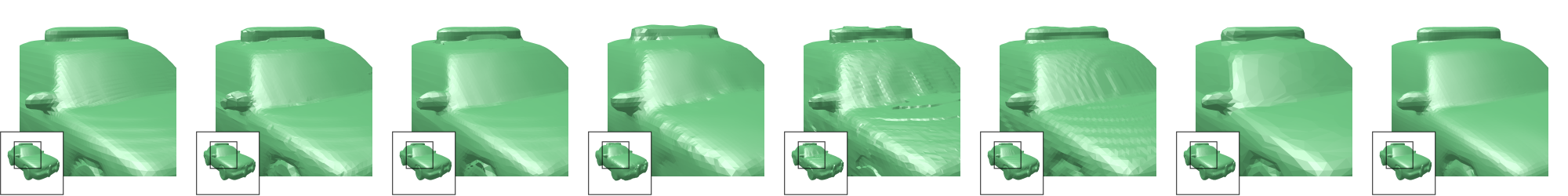}} \\
        \multicolumn{8}{c}{\includegraphics[width=\textwidth]{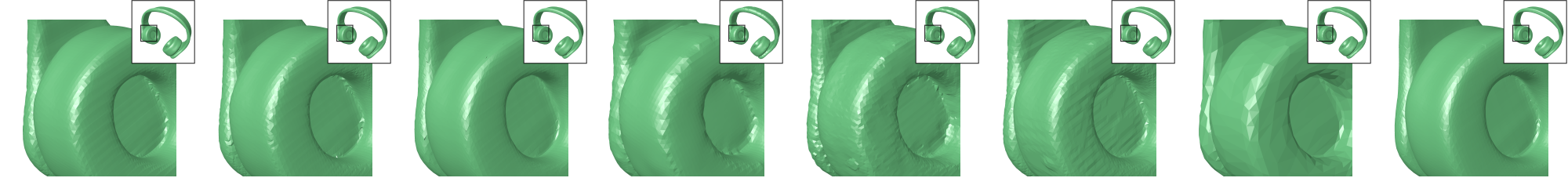}} \\
        \multicolumn{8}{c}{\includegraphics[width=\textwidth]{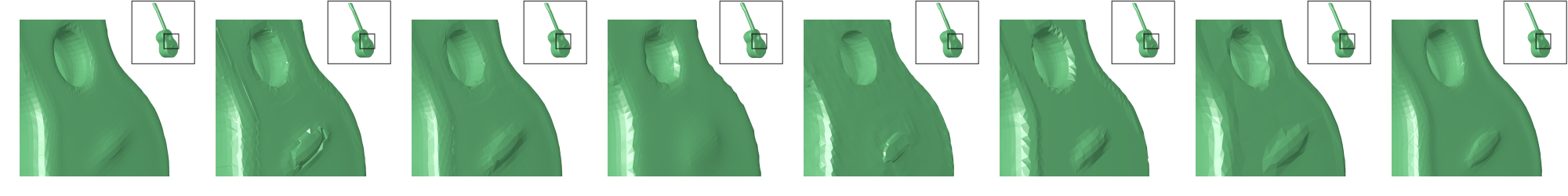}} \\
        \multicolumn{8}{c}{\includegraphics[width=\textwidth]{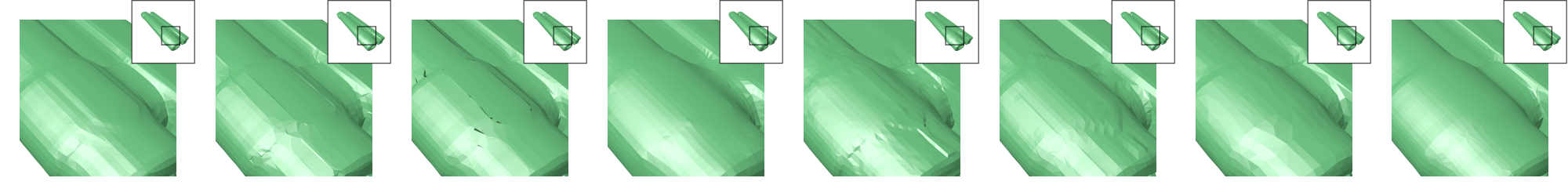}}
    \end{tabular}
    \caption{Qualitative results with generation of 3DShape2VecSet~\cite{zhang20233dshape2vecset}.}
    \label{fig:3DShape2VecSet_result}
\end{figure*}


\begin{table}[!t]
    \setlength{\tabcolsep}{2.4pt}
    \small
    \centering
    \caption{Quantitative results with SALAD~\cite{koo2023salad} and 3DShape2VecSet~\cite{zhang20233dshape2vecset}.}
    \vspace{-\baselineskip}
    \label{tab:salad_3dshape2vecset_supp}
    \begin{tabular}{ccccc|cccc}
        \toprule
        \multirow{2}{*}{$|\phi - 0.5|$ $\downarrow$} & \multicolumn{4}{c}{\textbf{SALAD}} & \multicolumn{4}{c}{\textbf{3DShape2VecSet}} \\
        & airplane & chair & table & mean & airplane & chair & table & mean \\
        \midrule
        MC & 0.315 & 0.268 & 0.390 & 0.324 & 1.515 & 1.422 & 3.368 & 2.330 \\
        IC & 0.322 & 0.275 & 0.388 & 0.329 & 1.597 & 1.316 & 3.244 & 2.292 \\
        MDC & 0.294 & 0.248 & 0.374 & 0.305 & 1.401 & 1.314 & 3.235 & 2.245 \\
        LMC & 0.478 & 0.485 & 0.490 & 0.484 & 9.256 & 7.361 & 9.932 & 8.824 \\
        NDC & 0.299 & 0.275 & 0.398 & 0.324 & 2.872 & 2.384 & 5.367 & 3.552 \\
        CDIF & 0.214 & 0.175 & 0.315 & 0.234 & 1.639 & 1.115 & 3.017 & 1.961 \\
        MISE & 0.207 & 0.170 & 0.335 & 0.237 & 0.873 & 1.013 & 2.287 & 1.733 \\
        \rowcolor{Gray}
        ODC & \textbf{0.047} & \textbf{0.024} & \textbf{0.041} & \textbf{0.037} & \textbf{0.237} & \textbf{0.102} & \textbf{0.135} & \textbf{0.112} \\
        \bottomrule
        \toprule
        
        \multirow{2}{*}{SI (\%) $\downarrow$} & \multicolumn{4}{c}{\textbf{SALAD}} & \multicolumn{4}{c}{\textbf{3DShape2VecSet}} \\
        & airplane & chair & table & mean & airplane & chair & table & mean \\
        \midrule

        MC & 0.00 & 0.00 & 0.00 & 0.00 & 0.00 & 0.00 & 0.00 & 0.00 \\
        IC & 0.00 & 0.00 & 0.00 & 0.00 & 0.00 & 0.00 & 0.00 & 0.00 \\
        MDC & 100. & 100. & 100. & 100. & 100. & 100. & 100. & 100. \\
        LMC & 0.00 & 0.00 & 0.00 & 0.00 & 0.00 & 0.00 & 0.00 & 0.00 \\
        NDC & 49.6 & 25.8 & 26.2 & 33.9 & 26.7 & 20.0 & 30.0 & 21.2 \\
        CDIF & 0.00 & 0.00 & 0.00 & 0.00 & 0.00 & 0.00 & 0.00 & 0.00 \\
        MISE & 30.0 & 13.0 & 26.4 & 23.1 & 13.3 & 20.0 & 20.0 & 22.3 \\
        \rowcolor{Gray}
        ODC & 0.00 & 0.00 & 0.00 & 0.00 & 0.00 & 0.00 & 0.00 & 0.06 \\
        \bottomrule
        \toprule
        
        \multirow{2}{*}{Man. (\%) $\uparrow$} & \multicolumn{4}{c}{\textbf{SALAD}} & \multicolumn{4}{c}{\textbf{3DShape2VecSet}} \\
        & airplane & chair & table & mean & airplane & chair & table & mean \\
        \midrule

        MC & 100. & 100. & 100. & 100. & 100. & 100. & 100. & 100. \\
        IC & 6.00 & 31.0 & 28.4 & 21.8 & 33.3 & 33.3 & 60.0 & 53.2 \\
        MDC & 100. & 100. & 100. & 100. & 100. & 100. & 100. & 100. \\
        LMC & 30.6 & 73.6 & 66.4 & 56.9 & 56.7 & 70.0 & 90.0 & 72.8 \\
        NDC & 10.8 & 50.6 & 50.0 & 37.1 & 40.0 & 46.7 & 60.0 & 53.3 \\
        CDIF & 100. & 100. & 100. & 100. & 100. & 100. & 100. & 100. \\
        MISE & 77.2 & 79.0 & 66.6 & 74.3 & 83.3 & 66.7 & 86.7 & 82.0 \\
        \rowcolor{Gray}
        ODC & 100. & 100. & 100. & 100. & 100. & 100. & 100. & 100. \\
        \bottomrule
        \toprule
        
        \multirow{2}{*}{\#T} & \multicolumn{4}{c}{\textbf{SALAD}} & \multicolumn{4}{c}{\textbf{3DShape2VecSet}} \\
        & airplane & chair & table & mean & airplane & chair & table & mean \\
        \midrule

        MC & 18760 & 37833 & 41118 & 32570 & 20626 & 67374 & 65767 & 61445 \\
        IC & 19073 & 38112 & 41360 & 32848 & 20725 & 67484 & 65845 & 61549 \\
        MDC & 18767 & 37828 & 41116 & 32570 & 20639 & 67362 & 65763 & 61442 \\
        LMC & 16983 & 35177 & 38007 & 30056 & 18944 & 61606 & 61701 & 56043 \\
        NDC & 18547 & 37589 & 40404 & 32180 & 20590 & 66698 & 65487 & 60993 \\
        CDIF & 17908 & 36924 & 40016 & 31616 & 19848 & 66005 & 64692 & 60661 \\
        MISE & 5000 & 5000 & 5000 & 5000 & 5000 & 5000 & 5000 & 5000 \\
        \rowcolor{Gray}
        ODC & 18825 & 37860 & 41173 & 32619 & 20668 & 67389 & 65782 & 61467 \\
        \bottomrule
        \toprule
        
        \multirow{2}{*}{Time (s)} & \multicolumn{4}{c}{\textbf{SALAD}} & \multicolumn{4}{c}{\textbf{3DShape2VecSet}} \\
        & airplane & chair & table & mean & airplane & chair & table & mean \\
        \midrule

        MC & 1.917 & 1.854 & 1.899 & 1.890 & 0.443 & 0.441 & 0.437 & 0.442 \\
        IC & 1.885 & 1.920 & 1.886 & 1.897 & 0.473 & 0.481 & 0.468 & 0.474 \\
        MDC & 1.964 & 1.961 & 2.008 & 1.978 & 0.522 & 0.527 & 0.525 & 0.529 \\
        LMC & 3.115 & 5.118 & 4.976 & 4.403 & 4.529 & 11.633 & 9.033 & 9.750 \\
        NDC & 0.629 & 0.630 & 0.639 & 0.620 & 0.989 & 0.988 & 0.991 & 0.986 \\
        CDIF & 3.351 & 4.778 & 5.101 & 4.410 & 0.761 & 1.521 & 1.487 & 1.408 \\
        MISE & 1.157 & 1.229 & 1.280 & 1.222 & 1.835 & 1.849 & 1.562 & 1.814 \\
        \rowcolor{Gray}
        ODC & 3.004 & 3.658 & 3.898 & 3.520 & 1.411 & 1.778 & 1.823 & 1.700 \\

        \bottomrule
    \end{tabular}
    \vspace{0.5cm}
\end{table}
More qualitative results of SALAD~\cite{koo2023salad} are presented in Figure~\ref{fig:SALAD_result} and~\ref{fig:SALAD_result2}. Also, those of 3DShape2VecSet~\cite{zhang20233dshape2vecset} are presented in Figure~\ref{fig:3DShape2VecSet_result}.

The quantitative results of the three categories and mean across all categories are presented in Table~\ref{tab:salad_3dshape2vecset_supp}.

\subsection{Results with Myles' Dataset}
\label{supp_sec:myles_occ_results}
\begin{figure*}[!t]
    \centering
    \setlength{\tabcolsep}{0em}
    \def\arraystretch{0.0}

    \begin{tabular}{P{0.111\textwidth}P{0.111\textwidth}P{0.111\textwidth}P{0.111\textwidth}P{0.111\textwidth}P{0.111\textwidth}P{0.111\textwidth}P{0.111\textwidth}P{0.112\textwidth}}
        \textbf{MC}~\shortcite{lewiner2003efficient} & \textbf{$^\dagger$IC}~\shortcite{ju2006intersection} & \textbf{$^\dagger$MDC}~\shortcite{manifolddualcontouring} & \textbf{LMC}~\shortcite{lempitsky2010surface} & \textbf{NDC}~\shortcite{neuraldualcontouring} & \textbf{CDIF}~\shortcite{manson2011contouring} & \textbf{$^\dagger$MISE}~\shortcite{occupancenetworks} & \textbf{ODC} (ours) & \textbf{GT} \\
        \multicolumn{9}{c}{\includegraphics[width=\textwidth]{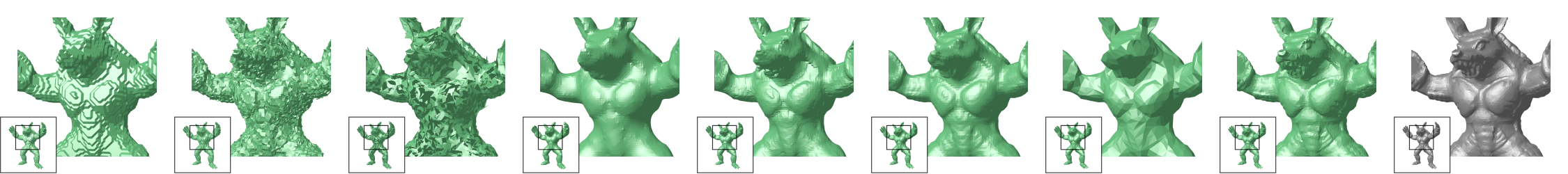}} \\
        \multicolumn{9}{c}{\includegraphics[width=\textwidth]{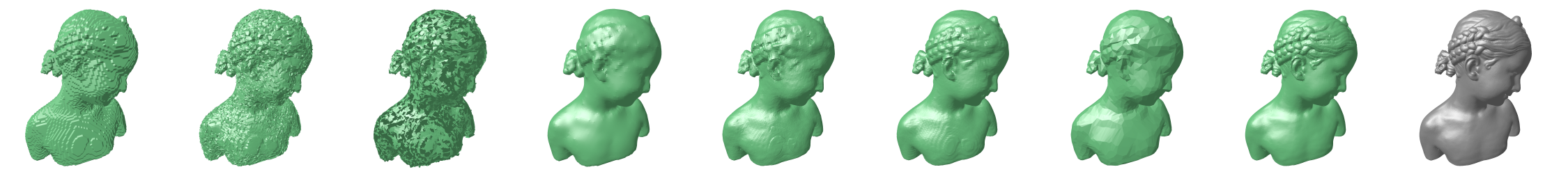}} \\
        \multicolumn{9}{c}{\includegraphics[width=\textwidth]{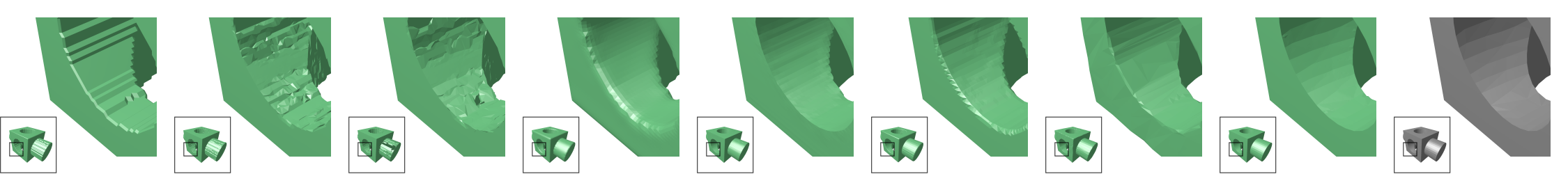}} \\
        \multicolumn{9}{c}{\includegraphics[width=\textwidth]{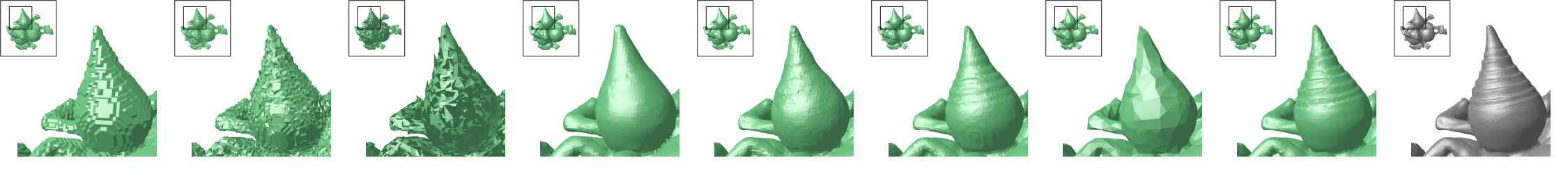}} \\
        \multicolumn{9}{c}{\includegraphics[width=\textwidth]{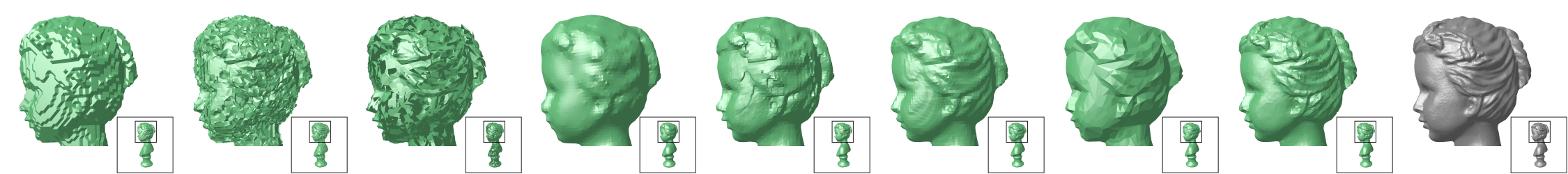}} \\
        \multicolumn{9}{c}{\includegraphics[width=\textwidth]{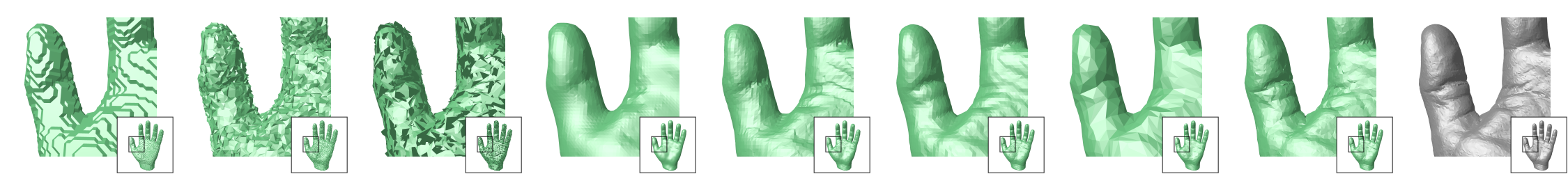}} \\
        \multicolumn{9}{c}{\includegraphics[width=\textwidth]{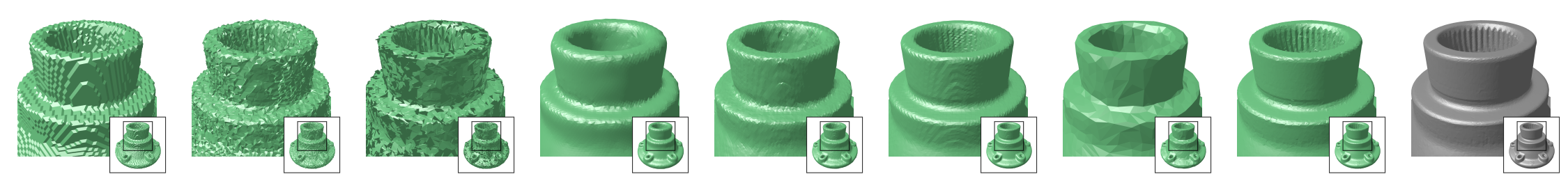}} \\
        \multicolumn{9}{c}{\includegraphics[width=\textwidth]{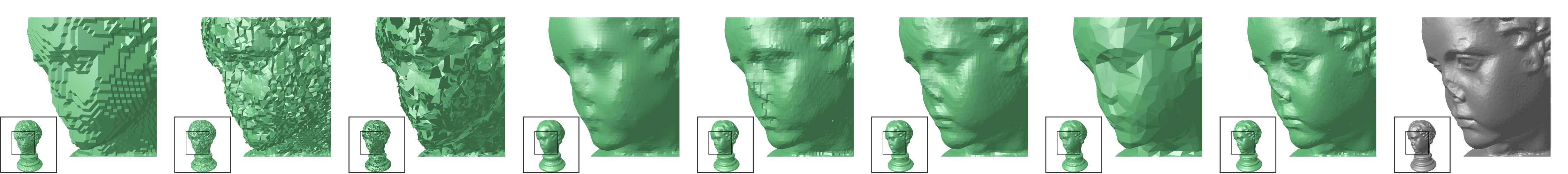}} \\
        \multicolumn{9}{c}{\includegraphics[width=\textwidth]{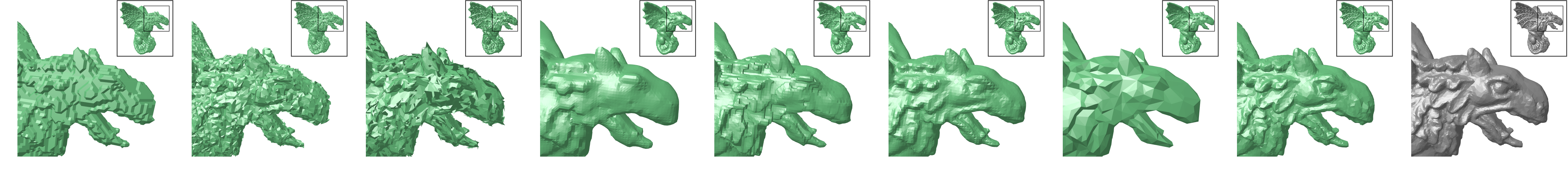}} \\
        \multicolumn{9}{c}{\includegraphics[width=\textwidth]{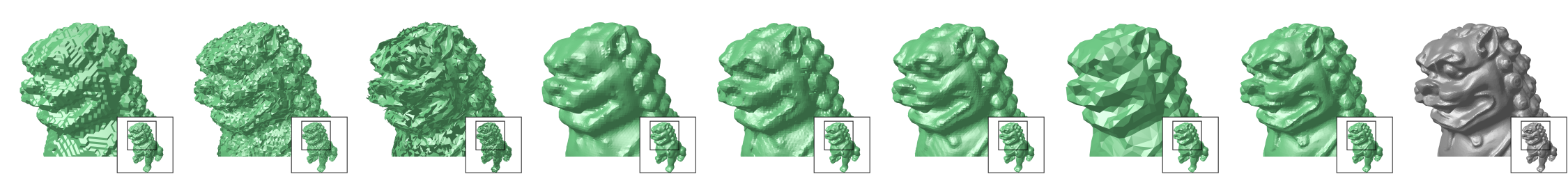}}
    \end{tabular}
    \caption{Qualitative results with Myles' dataset~\cite{myles2014robust}. The input is provided by an occupancy function. $^\dagger$The method also requires the gradient of the SDF due to its specifications.}
    \label{fig:Myles_occ_result_supp}
\end{figure*}

More qualitative results of occupancy input from Myles' dataset~\cite{myles2014robust} are presented in Figure~\ref{fig:Myles_occ_result_supp}.

\subsection{Ablation Study with SALAD and Myles' Dataset}
\label{supp_sec:salad_myles_ablation}

More qualitative results of ablation study with SALAD~\cite{koo2023salad} and Myles' dataset~\cite{myles2014robust} are presented in Figure~\ref{fig:SALAD_ablation_supp} and~\ref{fig:Myles_ablation}, respectively.

More quantitative results of SALAD ablation study with SALAD and Myles' dataset are presented in Table~\ref{tab:salad_ablation} and ~\ref{tab:myles_ablation}, respectively.

\newpage

\begin{figure*}[!t]
    \centering
    \setlength{\tabcolsep}{0em}
    \def\arraystretch{0.0}

    \begin{tabular}{P{0.143\textwidth}P{0.143\textwidth}P{0.143\textwidth}P{0.143\textwidth}P{0.143\textwidth}P{0.143\textwidth}P{0.142\textwidth}}
        \textbf{MC}~\shortcite{lewiner2003efficient} & \textbf{+ 1D Search} & \textbf{IC}~\shortcite{ju2006intersection} & \textbf{MDC}~\shortcite{manifolddualcontouring} & \textbf{+ 1D Search} & \textbf{+ 2D Search} & \textbf{+ IC (=ODC)} \\
        \multicolumn{7}{c}{\includegraphics[width=\textwidth]{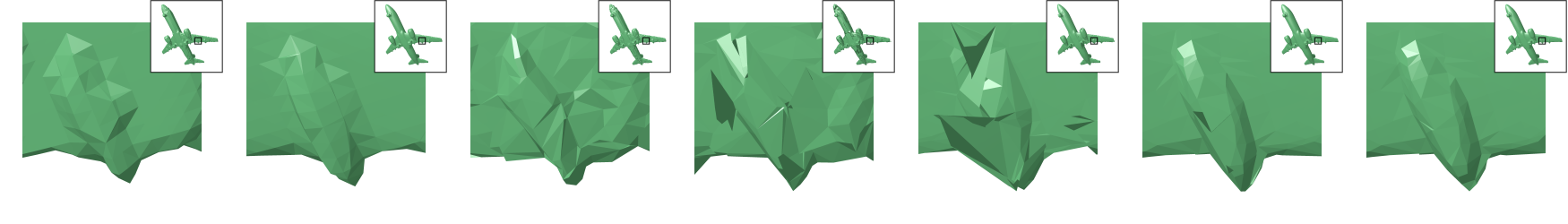}} \\
        \multicolumn{7}{c}{\includegraphics[width=\textwidth]{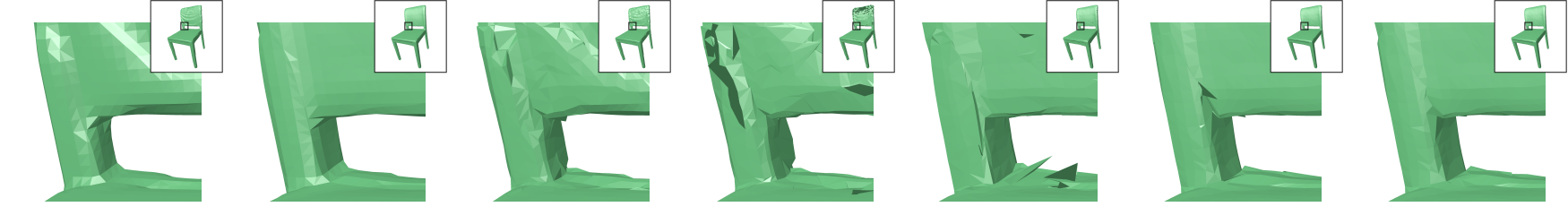}} \\
        \multicolumn{7}{c}{\includegraphics[width=\textwidth]{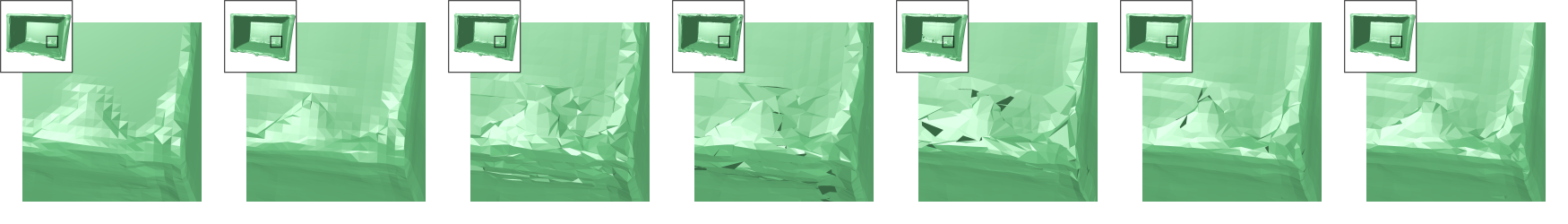}}
    \end{tabular}
    \vspace{-0.55cm}
    \caption{\textcolor{color_4}{Ablation studies with unconditional generation of SALAD~\cite{koo2023salad}. The second column from the left represents MC with 1D point search, while the rightmost three columns correspond to MDC with 1D point search and gradient as a normal, MDC with 1D and 2D point search, and ODC, respectively.}}
    \label{fig:SALAD_ablation_supp}
\end{figure*}

\begin{figure*}[!]
    \centering
    \setlength{\tabcolsep}{0em}
    \def\arraystretch{0.0}
    \vspace{0.1cm}
    \begin{tabular}{P{0.125\textwidth}P{0.125\textwidth}P{0.125\textwidth}P{0.125\textwidth}P{0.125\textwidth}P{0.125\textwidth}P{0.125\textwidth}P{0.125\textwidth}}
        \textbf{MC}~\shortcite{lewiner2003efficient} & \textbf{+ 1D Search} & \textbf{$^\dagger$IC}~\shortcite{ju2006intersection} & \textbf{$^\dagger$MDC}~\shortcite{manifolddualcontouring} & \textbf{$^\dagger$+ 1D Search} & \textbf{+ 2D Search} & \textbf{+ IC (=ODC)} & \textbf{GT} \\
        \multicolumn{8}{c}{\includegraphics[width=\textwidth]{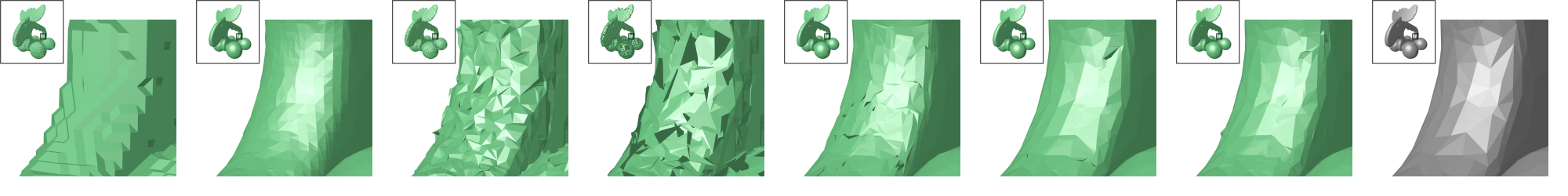}} \\
        \multicolumn{8}{c}{\includegraphics[width=\textwidth]{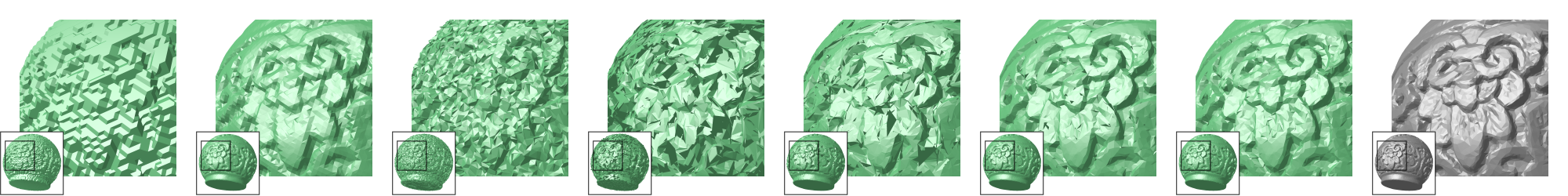}} \\
        \multicolumn{8}{c}{\includegraphics[width=\textwidth]{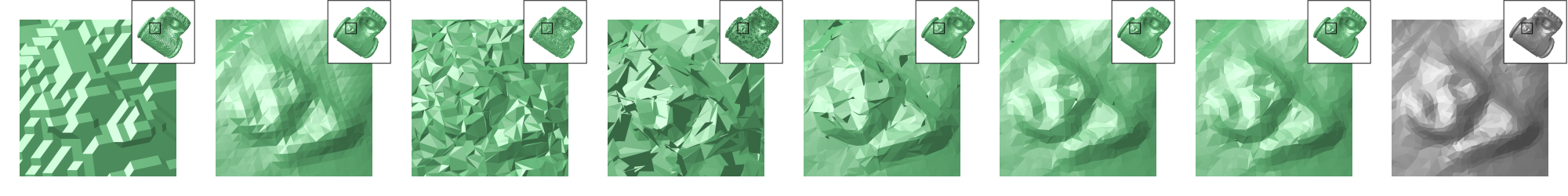}} \\
        \multicolumn{8}{c}{\includegraphics[width=\textwidth]{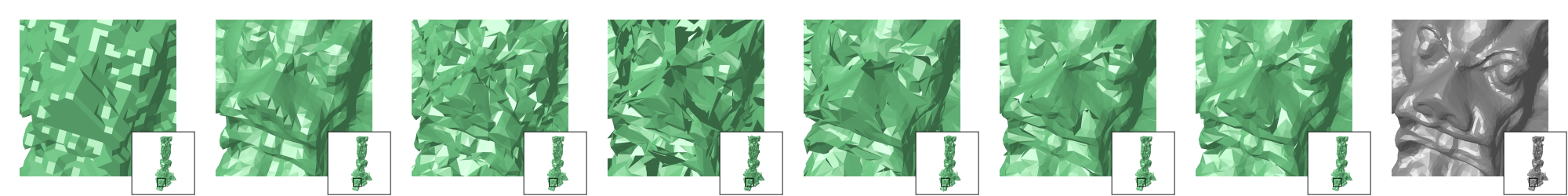}} \\
        \multicolumn{8}{c}{\includegraphics[width=\textwidth]{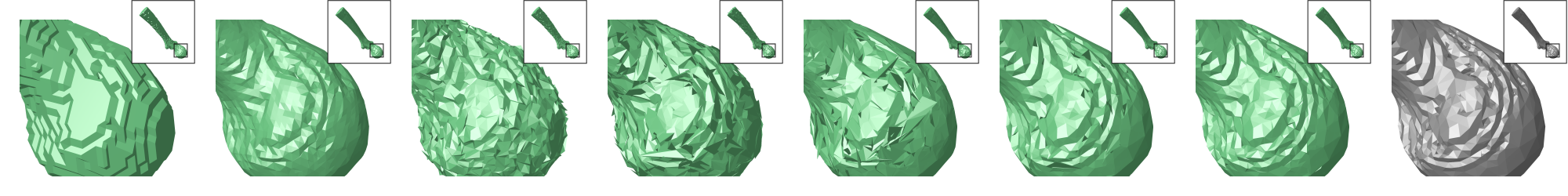}}
    \end{tabular}
    \vspace{-0.5cm}
    \caption{\textcolor{color_4}{Ablation studies with Myles'~\cite{myles2014robust} dataset. The input is provided by an occupancy function. $^\dagger$The method also requires the gradient of the SDF due to its specifications. The second column from the left represents MC with 1D point search, while the rightmost four columns correspond to MDC with 1D point search and gradient as local surface normal, MDC with 1D and 2D point search, ODC, and the ground truth shape, respectively.}}
    \label{fig:Myles_ablation}
    \vspace{-\baselineskip}
\end{figure*}

\begin{table}[!t]
    \setlength{\tabcolsep}{2.1pt}
    \small
    \centering
    \caption{\textcolor{color_4}{Ablation study with SALAD~\cite{koo2023salad}.}}
    \vspace{-\baselineskip}
    \label{tab:salad_ablation}
    \begin{tabular}{ccccc|cccc}
        \toprule
        \textbf{SALAD} & \multicolumn{4}{c}{$|\phi - 0.5|$ $\downarrow$} & \multicolumn{4}{c}{SI (\%) $\downarrow$} \\
        \textbf{Ablation}& airplane & chair & table & mean & airplane & chair & table & mean \\
        \midrule
        MC & 0.315 & 0.268 & 0.390 & 0.324 & 0.00 & 0.00 & 0.00 & 0.00 \\
        + 1D search & 0.075 & 0.039 & 0.059 & 0.058 & 0.00 & 0.00 & 0.00 & 0.00 \\
        \midrule
        IC & 0.322 & 0.275 & 0.388 & 0.329 & 0.00 & 0.00 & 0.00 & 0.00 \\
        \midrule
        MDC & 0.294 & 0.248 & 0.374 & 0.305 & 100. & 100. & 100. & 100. \\
        + 1D search & 0.146 & 0.075 & 0.106 & 0.109 & 100. & 100. & 100. & 100. \\
        + 2D search & 0.048 & \textbf{0.024} & 0.042 & 0.038 & 100. & 100. & 100. & 100. \\
        \rowcolor{Gray}
        + IC (=ODC) & \textbf{0.047} & \textbf{0.024} & \textbf{0.041} & \textbf{0.037} & 0.00 & 0.00 & 0.00 & 0.00 \\
        \bottomrule
        \toprule
        \textbf{SALAD} & \multicolumn{4}{c}{Man. (\%) $\uparrow$} & \multicolumn{4}{c}{Time (s)} \\
        \textbf{Ablation}& airplane & chair & table & mean & airplane & chair & table & mean \\
        \midrule

        MC & 100. & 100. & 100. & 100. & 1.9 & 1.9 & 1.9 & 1.9 \\
        + 1D search & 100. & 100. & 100. & 100. & 1.9 & 2.0 & 2.0 & 2.0 \\
        \midrule
        IC & 6.00 & 31.0 & 28.4 & 21.8 & 1.9 & 1.9 & 1.9 & 1.9 \\
        \midrule
        MDC & 100. & 100. & 100. & 100. & 2.0 & 2.0 & 2.0 & 2.0 \\
        + 1D search & 100. & 100. & 100. & 100. & 2.1 & 2.3 & 2.3 & 2.2 \\
        + 2D search & 100. & 100. & 100. & 100. & 2.8 & 3.8 & 3.8 & 3.5 \\
        \rowcolor{Gray}
        + IC (=ODC) & 100. & 100. & 100. & 100. & 3.0 & 3.7 & 3.9 & 3.5 \\

        \bottomrule
        \toprule
        \textbf{SALAD} & \multicolumn{4}{c}{\#V} & \multicolumn{4}{c}{\#T} \\
        \textbf{Ablation}& airplane & chair & table & mean & airplane & chair & table & mean \\
        \midrule

        MC & 9384 & 18915 & 20558 & 16286 & 18760 & 37833 & 41118 & 32570 \\
        + 1D search & 9384 & 18914 & 20558 & 16285 & 18758 & 37833 & 41118 & 32570 \\
        \midrule
        IC & 9531 & 19049 & 20674 & 16418 & 19073 & 38112 & 41360 & 32848 \\
        \midrule
        MDC & 9388 & 18913 & 20557 & 16286 & 18767 & 37828 & 41116 & 32570 \\
        + 1D search & 9388 & 18913 & 20557 & 16286 & 18767 & 37829 & 41115 & 32570 \\
        + 2D search & 9388 & 18913 & 20557 & 16286 & 18767 & 37829 & 41115 & 32570 \\
        \rowcolor{Gray}
        + IC (=ODC) & 9417 & 18929 & 20586 & 16310 & 18825 & 37860 & 41173 & 32619 \\
        
        \bottomrule
    \end{tabular}
    \vspace{1.5cm}
\end{table}
\begin{table}[!t]
    \setlength{\tabcolsep}{6.0pt}
    \small
    \centering
    \caption{\textcolor{color_4}{Ablation study with Myles's dataset~\cite{myles2014robust}. The input is given by an occupancy function. $^\dagger$The method also requires the gradient of the SDF due to its specifications.}}
    \vspace{-\baselineskip}
    \label{tab:myles_ablation}
    \begin{tabular}{cccccccc}
        \toprule
        \textbf{Myles'} & MD2 $\downarrow$ & \multirow{2}{*}{NIC $\downarrow$} & HDD $\downarrow$ & \multirow{2}{*}{\#T} & SI $\downarrow$ & Man. $\uparrow$ \\
        \textbf{Ablation} & ($\times 10^6$) &  & ($\times 10^2$) & & (\%) & (\%) \\
        \midrule
        MC & 2.261 & 0.366 & 0.894 & 77944 & 0.00 & 100. \\
        + 1D search & 0.208 & 0.093 & 0.792 & 77934 & 0.00 & 100. \\
        \midrule
        IC & 2.808 & 0.583 & 0.959 & 80267 & 0.00 & 11.4 \\
        \midrule
        $^\dagger$MDC & 1.857 & 1.042 & 2.309 & 77926 & 100. & 100. \\
        $^\dagger$+ 1D search & 0.545 & 0.307 & 2.166 & 77926 & 100. & 100. \\
        + 2D search & 0.114 & 0.098 & 1.079 & 77926 & 98.7 & 100. \\
        \rowcolor{Gray}
        + IC (=ODC) & \textbf{0.113} & \textbf{0.072} & \textbf{0.632} & 78142 & 0.00 & 100. \\
        \bottomrule
    \end{tabular}
    \ifpaper \vspace{2.0cm}
    \else \vspace{-\baselineskip}
    \fi
\end{table}

\else

\fi

\end{document}